\IfFileExists{robbyant.cls}{
    \documentclass[onecolumn]{robbyant}
}{
    \documentclass[onecolumn]{./robbyant}
}

\makeatletter
\def\input@path{{./}{./}}
\makeatother
\graphicspath{{./}{./}}

\metadata[Website]{\url{https://technology.robbyant.com/lingbot-video}}
\metadata[Github]{\url{https://github.com/robbyant/lingbot-video}}
\metadata[Checkpoints]{\url{https://huggingface.co/robbyant/lingbot-video}}
\newcommand{\methodname}{LingBot-Video}
\newcommand{\method}{\texttt{\methodname}\xspace}
\newcommand{\acwm}{\texttt{\methodname-A2V}\xspace}

\newcommand{\x}{{\bm x}}
\newcommand{\E}{\mathbb{E}}

\newcommand{\Loss}{\mathcal{L}}
\usepackage[table]{xcolor}

\usepackage{listings}
\tcbuselibrary{listings}
\newtcblisting{captionexample}[1]{
  listing only,
  enhanced jigsaw,
  breakable,
  colback=gray!10,
  colframe=gray!80,
  width=\textwidth,
  title={#1},
  fonttitle=\bfseries\small,
  listing options={
    basicstyle=\ttfamily\scriptsize,
    breaklines=true,
    columns=fullflexible,
    keepspaces=true,
    postbreak=\mbox{\textcolor{gray}{$\hookrightarrow$}\space},
  },
}

\title{Scaling Mixture-of-Experts Video Pretraining \\[4pt] for Embodied Intelligence}

\author{
\begin{center}
    Shuailei Ma$^*$,
    Jiaqi Liao$^*$,
    Xinyang Wang$^*$,
    Jingjing Wang$^*$,
    Chaoran Feng,
    Zijing Hu,
    Chong Bao,
    \\[2pt]
    Zichen Xi,
    Yuqi Gan, 
    Weisen Wang, 
    Yanhong Zeng, 
    Qin Zhao, 
    Zifan Shi, 
    Wei Wu, 
    Hao Ouyang, 
    \\[2pt]
    Qiuyu Wang,
    Shangzhan Zhang,
    Jiahao Shao, 
    Yipengjing Sun, 
    Liangxiao Hu,
    Lunke Pan,
    Nan Xue,
    \\[2pt]
    Kecheng Zheng, 
    Yinghao Xu, 
    Xing Zhu, 
    Yujun Shen, 
    Ka Leong Cheng$^\dagger$
    \\[12pt]
    $^*$Equal Contribution \qquad
    $^\dagger$Project Lead
\end{center}
}

\begin{document}

% !TEX root = ../main.tex
\abstract{
Despite the recent promise in robot control, video generative models suffer from a domain mismatch due to their primary focus on content creation.
For example, their design inherently prioritizes visual fidelity and creativity over computational efficiency and physical realism.
In this work, we present \method, a DiT-based video pretraining paradigm specifically tailored for embodied intelligence.
From the \textbf{\textit{architecture}} perspective, we adopt the Mixture-of-Experts (MoE), instead of dense, framework to achieve a better trade-off between modeling capacity and inference efficiency, and manage to scale it up from scratch.
From the \textbf{\textit{data}} perspective, we construct a data profiling engine that augments standard internet videos with extensive robot-oriented footage, encompassing manipulation, navigation, and egocentric perspectives, to equip the base model with an intrinsic understanding of actions and world dynamics.
From the \textbf{\textit{training}} perspective, we develop a multi-dimensional reward system to enforce the alignment regarding physical rationality and task completion, going beyond standard criteria such as aesthetics, prompt-following, and motion consistency.
Comprehensive evaluations validate its performance and efficiency as a video foundation model.
We contribute \method as the inaugural large-scale, open-source MoE video foundation model to the community, in a pioneering effort to bridge digital creativity and physical actuation.

}

\maketitle
\begin{figure}[ht]
\centering
\includegraphics[width=0.96\linewidth]{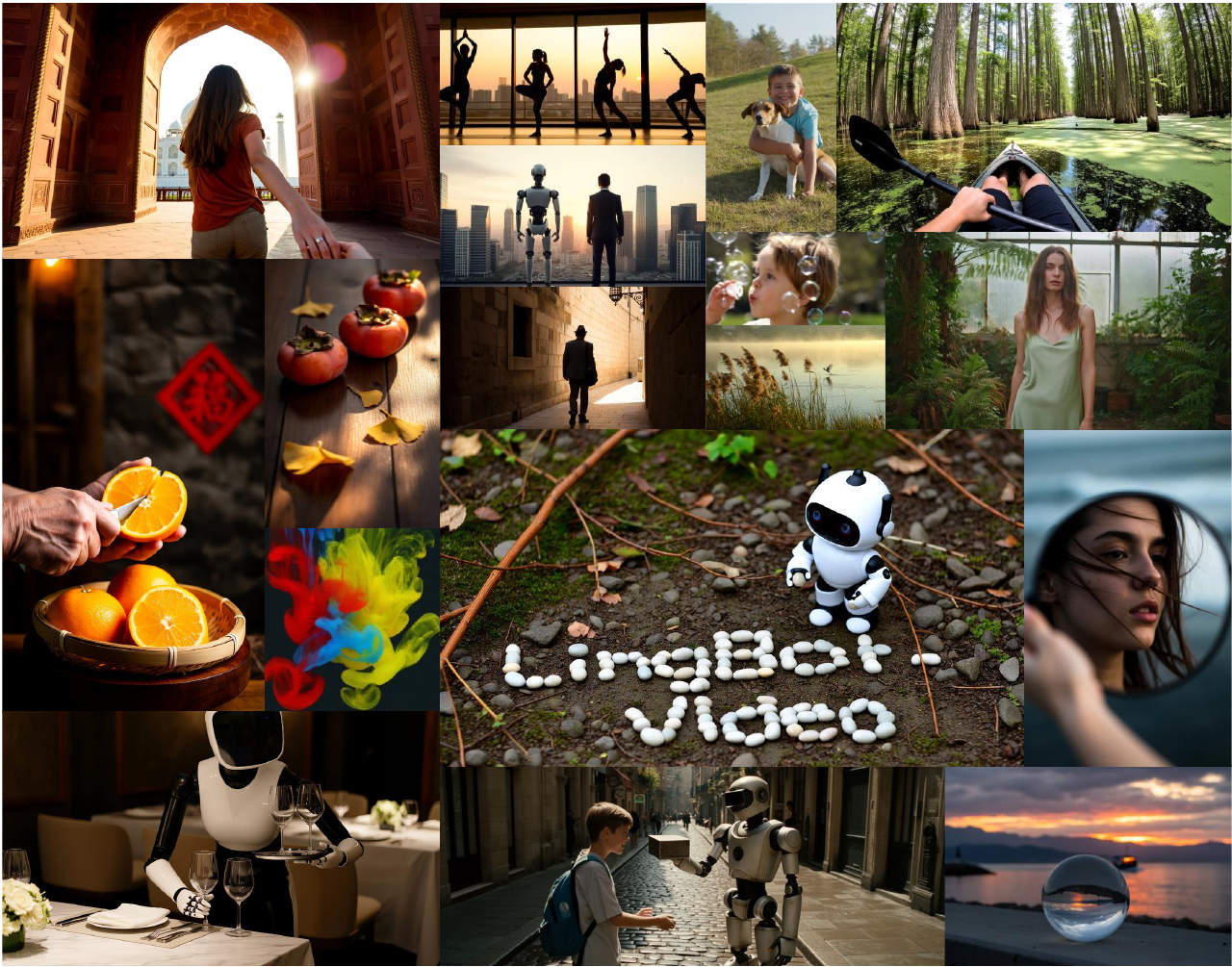}
\caption{Samples of \textit{Text-to-Image} and \textit{Text-to-Video} tasks generated by \method. 
\method can produce images and videos with high visual fidelity, rich details, and strong text-prompt alignment across diverse scenes and subjects.}
\label{fig:teaser}
\end{figure}

\justifying
% !TEX root = ../main.tex
\section{Introduction} \label{sec:intro}

Beyond their success in content creation, diffusion-based \cite{ho2022video,ho2022imagen,singer2022make,bar2024lumiere,blattmann2023stable,wan2025wan,seedance2026seedance} and autoregressive \cite{yan2021videogpt,villegas2022phenaki,gupta2024photorealistic} video models have demonstrated remarkable ability to synthesize temporally coherent and photorealistic sequences conditioned on text, images, and other control signals \cite{esser2023structure,guo2023animatediff,bruce2024genie,team2026advancing,sun2025worldplay}. This capability has motivated a growing body of work that interprets video models as implicit simulators of the physical world, enabling their use in robotics \cite{agarwal2026cosmos,liao2025genie,li2026causal}, autonomous driving \cite{russell2025gaia,ren2025cosmos}, and interactive environments \cite{bruce2024genie}. In this paradigm, video models serve not only as generative systems but also as predictive world models \cite{assran2025v} that support planning, policy learning, and imagination-based control. However, translating these models from passive video generation to active embodied reasoning and intelligence remains an open challenge.

Despite their promise, a fundamental gap persists between video generation models and embodied intelligence requirements. Most video foundation models are optimized for perceptual quality—such as realism, aesthetics, and text alignment—rather than physical correctness or controllability. While these objectives yield visually compelling results, they do not explicitly enforce consistency with physical interaction constraints, such as contact stability, rigid-body dynamics, or long-horizon state consistency under intervention. This highlights a key tension: while internet-scale video provides rich visual diversity, it does not guarantee fidelity to the constraints of embodied interaction.

Existing efforts to bridge video generation and embodied intelligence typically fall short across three tightly coupled dimensions: architecture, data, and training objectives. Architecturally, most diffusion-based video transformers rely on dense computation, where all parameters are activated uniformly across tokens and timesteps, resulting in prohibitive inference costs and limited scalability. While recent sparse Mixture-of-Experts (MoE) formulations~\cite{deepseek2024v3,lepikhin2020gshard} demonstrate promising efficiency gains in large language models (LLM) \cite{achiam2023gpt,touvron2023llama,team2023gemini,bai2023qwen,bi2024deepseek,zeng2022glm}, their adoption in video generation remains limited. 
Data-wise, training corpora are dominated by internet videos lacking robot embodiment priors or precise interaction dynamics, leading to weak grounding in physical simulation.
From a training perspective, current alignment strategies primarily optimize aesthetic quality and text-video correspondence, without incorporating explicit physical feasibility, task completion, or long-horizon reward signals. Consequently, existing systems struggle to simultaneously achieve scalability, physical consistency, and embodiment grounding.

In this work, we propose \method, a DiT-based video pretraining paradigm specifically designed for embodied intelligence. Our approach addresses the aforementioned limitations through three integrated components. First, we introduce a \textbf{Mixture-of-Experts (MoE) video frameword}, which enables sparse conditional computation, improving inference efficiency while scaling model capacity for complex spatiotemporal dynamics. Second, we construct \textbf{a robot-augmented pretraining corpus} that unifies internet-scale videos with robot manipulation, navigation, and egocentric datasets, thereby injecting explicit embodiment and interaction priors into the model. Third, we develop \textbf{a multi-dimensional reward system} that extends beyond aesthetic objectives to incorporate physical rationality and task-oriented success signals, encouraging the model to learn dynamics and interactions consistent with embodied environments. Together, these components enable a more physically grounded and computationally efficient video foundation model.

Overall, we present \method, to the best of our knowledge the \textbf{first} large-scale open-source Mixture-of-Experts video foundation model for embodied intelligence, bridging the gap between digital video generation and physical actuation. Our contributions are threefold:
\begin{itemize}
    \item We introduce a sparse Mixture-of-Experts (MoE) video diffusion framework with a scalable training paradigm, enabling scalability–efficiency trade-off for spatiotemporal modeling.
    \item We develop a dedicated data profiling engine that systematically analyzes, filters, and rebalances heterogeneous video sources. This enables effective integration of large-scale internet videos with embodied datasets, resulting in improved grounding in physical interactions, action semantics, and embodiment-specific dynamics.
    \item We propose a multi-dimensional reward system that incorporates physical plausibility and task-level success signals, extending beyond conventional perceptual and text-alignment objectives.
\end{itemize}

% !TEX root = ../main.tex
\section{Method} \label{sec:method}

\subsection{Task-Unified Single-Stream Diffusion Transformer}
Our architecture adopts a cascaded design consisting of a task-unified base generator and a refiner.
The base generator employs a Single-Stream Diffusion Transformer to process compact visual latents and multimodal conditions.
We use Qwen3-VL-4B~\cite{Qwen3-VL} to extract condition from multimodal instructions.
Wan2.1-VAE~\cite{wan2025wan} is employed for efficient visual latent compression.
This section is organized as follows: first, we introduce the \textbf{Unified Input Formulation} and \textbf{Single-Stream Diffusion Transformer};
then, we detail our scaling strategy via \textbf{Sparse Mixture-of-Experts}; finally, we present the \textbf{Cascaded Refiner}.

\noindent\textbf{Unified Input Formulation.}
We represent each training sample as a single token sequence consisting of visual latent patch and condition tokens. Specifically, after projecting visual patches and condition features into the same hidden dimension, we concatenate them along the sequence dimension to form a unified input. Under this formulation, we handle Text-to-Image (T2I), Text-to-Video (T2V), and Image-to-Video (TI2V) tasks within a single framework, representing image targets as a special single-frame ($T=1$) video generation case. To resolve the structural discrepancy between condition and visual tokens, we employ a 3D MM-RoPE~\cite{qin2025lumina,wu2025omnigen2,cai2025z} mechanism to place them in non-overlapping temporal coordinate ranges, eliminating the need for task-specific architectures or encoders.

\noindent\textbf{Single-Stream Diffusion Transformer.}
Efficiency and scalability guide the design of \method. Inspired by the scalability of LLMs' decoder-only architecture~\cite{achiam2023gpt,touvron2023llama,team2023gemini,bai2023qwen} and the parameter efficiency of single-stream designs, we adopt a streamlined single-stream diffusion transformer backbone. After lightweight modality-specific input projections, all visual and condition tokens share the same transformer blocks. This single-stream architecture maximizes parameter reuse and facilitates dense cross-modal interactions at every layer. Compared to dual-stream architectures~\cite{esser2024scaling} that process modalities via separate pathways, our unified backbone handles all tokens as a single sequence. This design groups multi-modal features into larger, unified GEMM computations, aiming to improve Model FLOPs Utilization (MFU)~\cite{chowdhery2022palmscalinglanguagemodeling} and reduce kernel launch overhead. Furthermore, while dual-stream counterparts require frequent concatenation and splitting of conditioning and latent tokens before and after attention within every block, our model processes the unified sequence continuously. This reduces memory-bandwidth-bound tensor reorganization and layout conversion overhead~\cite{dao2022flashattention}, which is particularly beneficial under sequence-parallel distributed layouts~\cite{korthikanti2022reducingactivationrecomputationlarge,jacobs2023deepspeed}. Consequently, in distributed scaling scenarios, this design simplifies communication patterns and helps avoid extra load imbalance and synchronization overhead from dual-stream partitioning.

\noindent\textbf{Multi-Modal 3D RoPE.}
To separate condition and visual tokens while preserving video geometry within a single stream, we map all tokens to a Multi-Modal 3D RoPE~\cite{qin2025lumina,wu2025omnigen2,cai2025z} coordinate system. Specifically, given $L$ condition tokens and an $F \times H \times W$ visual latent grid, conditioning tokens use temporal-only coordinates $(i,0,0)$ for $i=1,\ldots,L$, while visual patch tokens use $(L+1+f,h,w)$.
Query and key head dimensions are split across the temporal, vertical, and horizontal axes, applying the corresponding rotary frequencies independently.
This maintains spatial locality and temporal order while keeping attention fully single-stream.

\noindent\textbf{QK-Norm.}
To stabilize attention in deep transformer backbones, we normalize queries and keys with per-head RMSNorm before computing attention, following prior large-scale vision and diffusion transformer practice~\cite{dehghani2023vit22b,esser2024scaling}.
This stabilization mechanism helps control attention-logit growth and feature scales during high-resolution training, particularly under mixed precision.
Additionally, each block utilizes RMSNorm around the attention and feed-forward residual branches to keep activations bounded without introducing modality-specific pathways.

\noindent\textbf{AdaLN-Single Modulation.}
To reduce modulation overhead, we adopt the adaLN-single design~\cite{chen2023pixart,wan2025wan,wan2025wan2} by computing a shared timestep modulation once before the transformer stack. Each block adds a layer-specific trainable modulation table to this shared signal and uses the resulting shift, scale, and gate parameters to modulate its attention and feed-forward branches. The shared projection is zero-initialized following DiT-style adaptive normalization~\cite{peebles2023scalablediffusionmodelstransformers}, while the per-layer modulation tables are initialized with small random values, providing stable residual learning without per-block timestep MLPs.

\begin{figure}[t]
    \centering
    \includegraphics[width=0.8\linewidth]{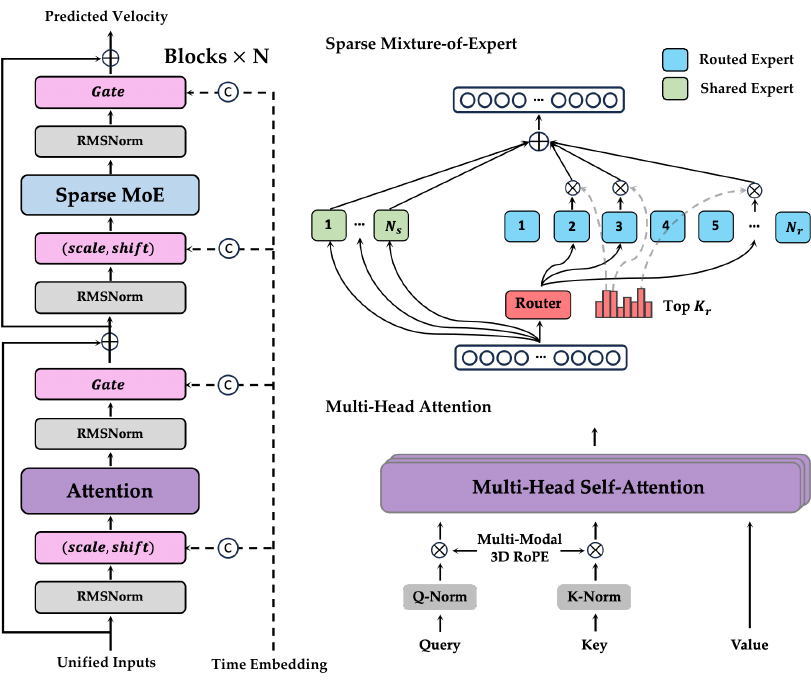}
    \caption{Overview of the task-unified single-stream diffusion transformer. Unified inputs are processed by stacked transformer blocks, where timestep modulation controls the attention and Sparse MoE branches; the attention branch applies QK-Norm and multi-modal 3D RoPE, while the MoE branch combines always-on shared experts with top-$K_{\mathrm{r}}$ routed experts before predicting velocity.}
    \label{fig:pure_single_stream_dit}
\end{figure}

\subsection{Scaling with Sparse Mixture-of-Experts}
The Mixture-of-Experts (MoE) paradigm is fundamentally motivated by the need to model highly diverse and complex data distributions, utilizing specialized subnetworks (experts) to capture distinct patterns or domains~\cite{jacobs1991adaptive,jordan1994hierarchical}. Video generation, which aims to simulate the continuous physical world, demands the model to capture the visual representations of complex physical processes (such as fluid motion and three-dimensional spatial consistency) along with varying motion trajectories and rich material textures~\cite{ho2022video,blattmann2023stable,wan2025wan}. Consequently, MoE is a natural candidate for scaling up video diffusion models to capture the vast complexity of physical laws~\cite{park2024switch,sha2026sparse}. In a conventional dense model, the Feed-Forward Network (FFN) forces all tokens to share the same parameter pathway. In unified video pretraining, this leads to severe subtask interference~\cite{jacobs1991adaptive,nowlan1990evaluation,jordan1994hierarchical} since a single set of parameters must simultaneously model asymmetric domains, such as spatial textures versus temporal motion, and accommodate diverse task formats spanning T2I, T2V, and TI2V conditions. Particularly for diffusion models, the training dynamics inherently exhibit a shift between distinct noise regimes, requiring the model to resolve the discrepancy between low-noise detail reconstruction and high-noise global layout formulation.

Sparse Mixture-of-Experts (MoE) has been robustly proven in large language models as a powerful paradigm to scale parameter capacity under a constant computational budget~\cite{shazeer2017moe,lepikhin2020gshard,switchtransformer,dai2024deepseekmoe,deepseek2024v3}. Driven by this success, we incorporate the sparse MoE framework into our base generator to achieve capacity-compute decoupling. This allows us to scale the total parameter capacity---creating a vast repository for diverse physical priors---while strictly controlling per-token active FLOPs. This decoupling is especially critical for high-resolution video generation, which easily scales to million-token (1M+) spatio-temporal sequences where dense scaling would incur prohibitive computational overhead. Diffusion-based video generators repeatedly process the full visual latent sequence across denoising steps~\cite{ho2022video,peebles2023scalablediffusionmodelstransformers}, making per-token transformer computation a dominant scaling pressure for long videos. Sparse routing reduces the active feed-forward computation at each denoising step while preserving total parameter capacity, making long-video capacity scaling more practical.

\noindent\textbf{Sparse MoE Architecture.}
In each transformer block, we preserve the single-stream FFN residual branch structure and replace only its dense feed-forward computation with a token-choice sparse MoE layer~\cite{shazeer2017moe,lepikhin2020gshard,switchtransformer}, as shown in~\cref{fig:pure_single_stream_dit}. Our MoE design incorporates key architectural principles of DeepSeekMoE~\cite{dai2024deepseekmoe,deepseek2024v3}, specifically fine-grained expert segmentation and shared expert isolation, to encourage expert specialization while maintaining shared common priors. Given the modulated FFN input $\mathbf{u}_t$ of token $t$, the Sparse MoE layer computes a branch output from shared experts and routed experts:
\begin{equation}
m(\mathbf{u}_t) = \sum_{i=1}^{N_{\mathrm{s}}} E_i^{(\mathrm{s})}(\mathbf{u}_t) + \sum_{j \in \mathcal{R}_{\mathrm{b}}(\mathbf{u}_t)} g_{t,j} E_j^{(\mathrm{r})}(\mathbf{u}_t),
\end{equation}
where $\mathcal{R}_{\mathrm{b}}(\mathbf{u}_t)$ is the selected routed-expert set, and $E_i^{(\mathrm{s})}(\cdot)$ and $E_j^{(\mathrm{r})}(\cdot)$ denote the $i$-th shared expert function and $j$-th routed expert function, respectively. Each expert is implemented as a SwiGLU MLP~\cite{shazeer2020gluvariantsimprovetransformer}. We denote $N_{\mathrm{s}}$ as the number of shared experts and $N_{\mathrm{r}}$ as the total number of routed experts. The gate $g_{t,j}$ is defined by the routing equations below. The resulting MoE branch output $m(\mathbf{u}_t)$ is injected back into the transformer block through the gated residual branch of the single-stream block. To maximize expert specialization and reduce knowledge hybridity~\cite{dai2024deepseekmoe}, we adopt a fine-grained expert segmentation strategy, setting the intermediate dimension of both the shared and routed experts to a smaller width than a standard dense FFN counterpart. Under this design, the shared experts provide a common pathway to capture general physical principles and spatial consistency across all tokens, while the routed experts capture specialized features.

For token $t$, we compute routed-expert affinities with a sigmoid router~\cite{deepseek2024v3}:
\begin{equation}
\alpha_{t,j} = \operatorname{Sigmoid}\left(\mathbf{u}_t^\top \mathbf{r}_j\right),
\end{equation}
Here, $\alpha_{t,j}$ is the router affinity between token $t$ and routed expert $j$, and $\mathbf{r}_j \in \mathbb{R}^d$ is the learnable router embedding of the $j$-th routed expert. To control communication cost in distributed training, we adopt a DeepSeek-style group-limited routing strategy~\cite{deepseek2024v3}. We divide the $N_{\mathrm{r}}$ routed experts into $N_{\mathrm{g}}$ groups. After adding the online correction bias, $\tilde{\alpha}_{t,j} = \alpha_{t,j} + b_j$, we select the top $K_{\mathrm{g}}$ groups, where each group is scored by the sum of its top-2 bias-corrected affinity scores. Let $\mathcal{G}_{\mathrm{b}}(\mathbf{u}_t)$ denote the set of experts belonging to these selected groups. Within $\mathcal{G}_{\mathrm{b}}(\mathbf{u}_t)$, we select the top $K_{\mathrm{r}}$ experts as $\mathcal{R}_{\mathrm{b}}(\mathbf{u}_t)$. The overall MoE architecture is depicted in~\cref{fig:pure_single_stream_dit}.

\noindent\textbf{Online Bias Correction for Load Balancing.}
To maintain load balance while preserving representation capacity, we use an auxiliary-loss-free load-balancing strategy~\cite{wang2024auxlossfree,deepseek2024v3}. We introduce a dynamic correction bias $b_j$ for each expert, which is adjusted during training and used only for selecting the top experts:
\begin{equation}
g'_{t,j} = \begin{cases} \alpha_{t,j}, & \text{if } \tilde{\alpha}_{t,j} \in \operatorname{TopK}_{K_{\mathrm{r}}}(\{\tilde{\alpha}_{t,k} \mid k \in \mathcal{G}_{\mathrm{b}}(\mathbf{u}_t)\}), \\ 0, & \text{otherwise}. \end{cases}
\end{equation}
Here, $g'_{t,j}$ is the sparse unnormalized gate: it keeps the original affinity $\alpha_{t,j}$ for selected experts and sets all unselected experts to zero. The final gating values are obtained by normalizing the selected original affinity scores and applying a route scaling factor $\gamma$:
\begin{equation}
g_{t,j} = \gamma \frac{g'_{t,j}}{\sum_{k=1}^{N_{\mathrm{r}}} g'_{t,k}}.
\end{equation}
Here, $g_{t,j}$ is the final routing weight applied to expert $j$ for token $t$, $\gamma$ is the route scaling factor, and $k$ indexes routed experts in the normalization term.
During training, the correction bias for expert $j$ is updated online at each optimizer step using the sign of its load deviation:
\begin{equation}
b_j \leftarrow b_j - \eta\,\operatorname{sign}(n_j - \bar{n}),
\end{equation}
where $b_j$ is the correction bias of expert $j$, $n_j$ is the number of valid token assignments selecting expert $j$ (accumulated globally across ranks), $\bar{n}$ is the average load per expert, and $\eta$ is the learning rate for bias updates. When bias centering is enabled, we mean-center the bias after each update:
\begin{equation}
b_j \leftarrow b_j - \frac{1}{N_{\mathrm{r}}}\sum_{k=1}^{N_{\mathrm{r}}} b_k.
\end{equation}
This centering preserves relative load-balancing signals while keeping the router input-dependent and stable.

\noindent\textbf{Sequence-Wise Auxiliary Loss.}
Video generation naturally involves long spatio-temporal token sequences, where batch-level expert-balance statistics can hide routing imbalance within individual videos.
We therefore adopt the sequence-wise auxiliary balance loss from DeepSeek-V3~\cite{deepseek2024v3}, which encourages balanced expert usage within each packed video sequence rather than only at the global batch level.
For a packed batch containing $S$ sequences, the sequence-wise balance loss is defined as:
\begin{equation}
\mathcal{L}_{\mathrm{seq}} = \frac{1}{S}\sum_{s=1}^{S}\sum_{j=1}^{N_{\mathrm{r}}} f_{j}^{(s)} P_{j}^{(s)},
\end{equation}
Here, $S$ is the number of packed sequences, $s$ indexes a sequence, and $j$ indexes a routed expert. We compute the average normalized routing probability for expert $j$ in sequence $s$ as:
\begin{equation}
P_{j}^{(s)} = \frac{1}{T_s}\sum_{t=1}^{T_s} p_{t,j},\qquad p_{t,j} = \frac{\alpha_{t,j}}{\sum_{k=1}^{N_{\mathrm{r}}} \alpha_{t,k}}.
\end{equation}
Here, $T_s$ is the token length of the $s$-th packed unified sequence, $p_{t,j}$ is the normalized routing probability of token $t$ choosing expert $j$, and $P_j^{(s)}$ is its average over the sequence. We compute the normalized assignment frequency as:
\begin{equation}
f_{j}^{(s)} = \frac{N_{\mathrm{r}}}{K_{\mathrm{r}}T_s}c_{j}^{(s)},\qquad c_{j}^{(s)} = \sum_{t=1}^{T_s}\mathbf{1}\left[\alpha_{t,j} \in \operatorname{TopK}_{K_{\mathrm{r}}}(\{\alpha_{t,k} \mid 1 \le k \le N_{\mathrm{r}}\})\right].
\end{equation}
Here, $c_j^{(s)}$ is the number of tokens in sequence $s$ for which expert $j$ belongs to the unbiased top-$K_{\mathrm{r}}$ set computed from the raw affinity $\alpha_{t,j}$ before online bias correction and group-limited routing. The indicator $\mathbf{1}[\cdot]$ returns one when this selection condition is true and zero otherwise. The normalized frequency $f_j^{(s)}$ is detached from the computation graph so that gradients only flow through $P_j^{(s)}$. The auxiliary balance loss is applied to all transformer blocks and added to the diffusion loss:
\begin{equation}
\mathcal{L}_{\mathrm{aux}} = \lambda_{\mathrm{aux}}\mathcal{L}_{\mathrm{seq}},
\end{equation}
\begin{equation}
\mathcal{L} = \mathcal{L}_{\mathrm{diff}} + \mathcal{L}_{\mathrm{aux}},
\end{equation}
where $\lambda_{\mathrm{aux}}$ is the auxiliary loss weight, $\mathcal{L}_{\mathrm{aux}}$ is the weighted sequence-wise balance loss, $\mathcal{L}_{\mathrm{diff}}$ is the diffusion training loss, and $\mathcal{L}$ is the final training objective. By routing each token to $K_{\mathrm{r}}$ active routed experts, we keep the active computational cost per token comparable to a standard dense counterpart, while the total parameter capacity scales with the size of the expert pool $N_{\mathrm{r}}$. This decoupling ensures that the generator scales its representation capacity to capture complex physical priors without incurring a linear increase in per-token FLOPs.

\subsection{Sparse Mixture-of-Experts Recipe Exploration} \label{subsec:moe_recipe}
To identify the optimal sparse routing configuration for unified video diffusion pre-training, we systematically explore the MoE design space. We structure our exploration along two primary dimensions: first, the scaling of the expert pool (total parameter capacity) under a fixed active compute budget; second, the granularity of routing (fine-grained routing with many small experts versus coarse routing with fewer large experts) under a fixed total parameter budget. All configurations are evaluated on a unified pre-training mixture.

\begin{figure}[htbp]
    \centering
    \begin{subfigure}[htbp]{0.49\linewidth}
        \centering
        \includegraphics[width=\linewidth]{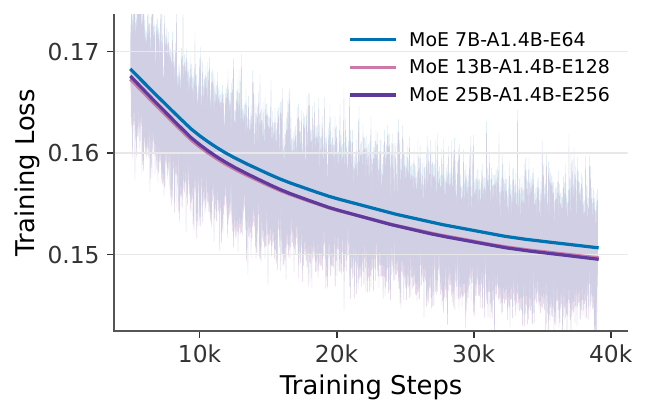}
        \caption{Training loss.}
        \label{fig:recipe_expert_scale_training}
    \end{subfigure}
    \hfill
    \begin{subfigure}[htbp]{0.49\linewidth}
        \centering
        \includegraphics[width=\linewidth]{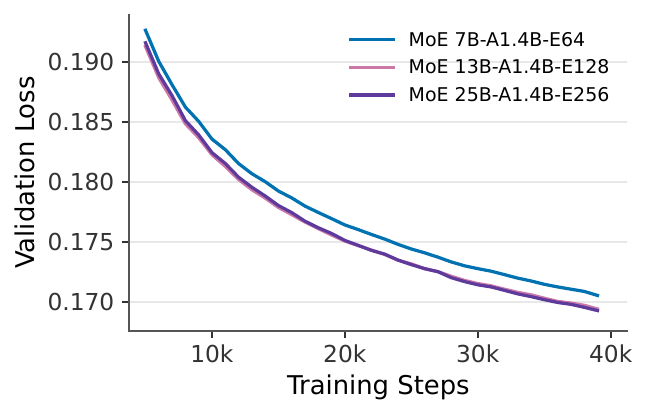}
        \caption{Validation loss.}
        \label{fig:recipe_expert_scale_validation}
    \end{subfigure}
    \caption{Expert-count recipe comparison using training and validation loss. Training curves show raw logged losses in the background and smoothed curves for visualization; validation curves are unsmoothed and aligned to the training-loss step range.}
    \label{fig:recipe_expert_scale_training_validation}
\end{figure}

\noindent\textbf{Expert Scale \& Capacity-Compute Decoupling.}
We first evaluate the impact of expanding the total parameter capacity by scaling the number of available routed experts $E \in \{64, 128, 256\}$ while holding the active parameter scale strictly constant at $1.4\,\mathrm{B}$ per token. As shown in~\cref{fig:recipe_expert_scale_training_validation}, scaling the expert count yields consistent improvements in both training and validation losses, demonstrating the efficacy of capacity-compute decoupling. However, the performance gain from $E=128$ to $E=256$ is marginal compared to the significant improvement from $E=64$ to $E=128$. Considering this trade-off among performance, latency, and memory footprint, we choose $E=128$ as the default expert scale, which captures most of the observed capacity benefit while avoiding the additional communication and storage overhead of a larger expert pool.

\begin{figure}[htbp]
    \centering
    \begin{subfigure}[htbp]{0.49\linewidth}
        \centering
        \includegraphics[width=\linewidth]{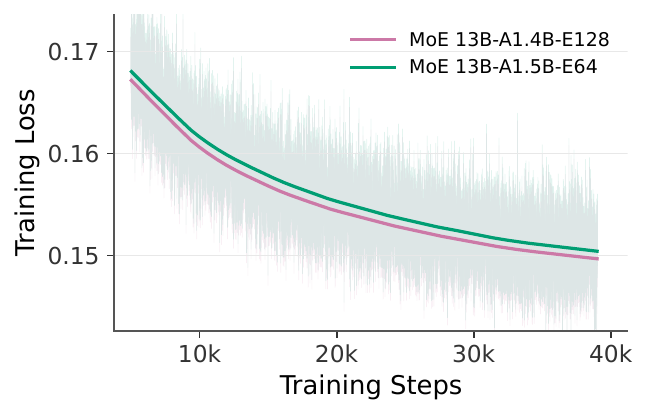}
        \caption{Training loss.}
        \label{fig:recipe_active_capacity_training}
    \end{subfigure}
    \hfill
    \begin{subfigure}[htbp]{0.49\linewidth}
        \centering
        \includegraphics[width=\linewidth]{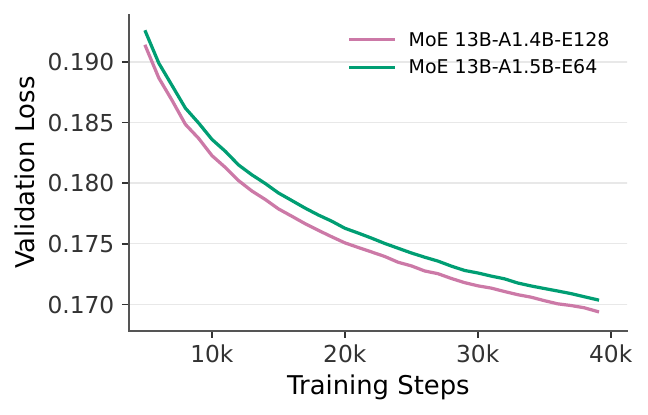}
        \caption{Validation loss.}
        \label{fig:recipe_active_capacity_validation}
    \end{subfigure}
    \caption{Active-capacity recipe comparison using training and validation loss. Training curves show raw logged losses in the background and smoothed curves for visualization; validation curves are unsmoothed and aligned to the training-loss step range.}
    \label{fig:recipe_active_capacity_training_validation}
\end{figure}

\noindent\textbf{Fine-Grained Specialization vs. Coarse Routing.}
Next, we ablate the partitioning of FFN parameters under a fixed total parameter budget of $13\,\mathrm{B}$. We compare a fine-grained routing recipe (\textit{MoE 13B-A1.4B-E128}, which routes each token to $K_{\mathrm{r}} = 8$ out of 128 experts) against a coarse routing recipe (\textit{MoE 13B-A1.5B-E64}, which routes each token to $K_{\mathrm{r}} = 4$ out of 64 experts). As illustrated in~\cref{fig:recipe_active_capacity_training_validation}, despite the coarse routing model having a higher active parameter count (and thus a larger FLOP footprint per token), it performs consistently worse than the fine-grained counterpart throughout training. This performance gap highlights the advantages of fine-grained expert specialization~\cite{dai2024deepseekmoe,deepseek2024v3}. By dividing the FFN parameters into more, smaller experts, the model gains access to a dramatically larger combinatorial routing space ($\binom{128}{8}$ vs $\binom{64}{4}$), following the sparse top-$K$ routing principle of MoE models~\cite{shazeer2017moe,lepikhin2020gshard,switchtransformer}. This enables visual tokens representing heterogeneous modalities and noise levels to form more customized execution pathways, whereas routing to fewer, larger experts leads to parameter-level gradient conflicts~\cite{jacobs1991adaptive,nowlan1990evaluation,jordan1994hierarchical} and knowledge hybridity~\cite{dai2024deepseekmoe}.

\subsection{Scaling Experiments} \label{subsec:scaling_experiments}
To verify the scalability and efficiency of our sparse single-stream diffusion transformer, we conduct scaling experiments across models of varying sizes, following the established practice of analyzing model scaling behavior and sparse MoE efficiency~\cite{kaplan2020scalinglaws,hoffmann2022trainingcomputeoptimal,shazeer2017moe,lepikhin2020gshard,switchtransformer}. We first benchmark the model under a compute-comparable regime, evaluating the benefits of sparse parameters against standard dense baselines. We then investigate the scaling trajectories of our architecture up to $120\,\mathrm{B}$ total parameters. Finally, we analyze the inference efficiency of our sparse architecture under varying sequence lengths.

\begin{figure}[htbp]
    \centering
    \includegraphics[width=0.96\linewidth]{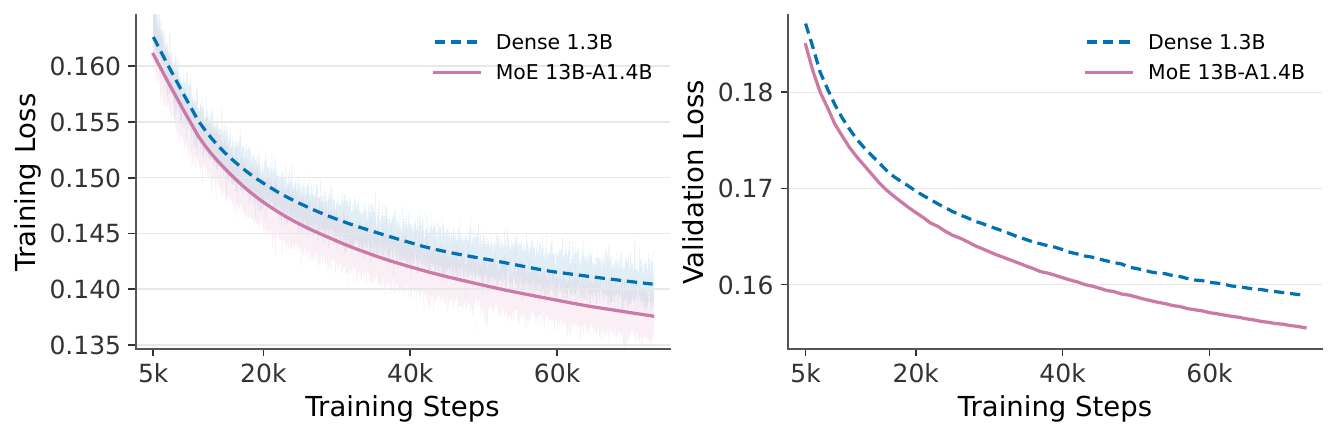}
    \caption{Comparable active-parameter scaling comparison between Dense 1.3B and MoE 13B-A1.4B using training and validation loss. Training curves show raw logged losses in the background and smoothed curves for visualization; validation curves are unsmoothed and aligned to the training-loss step range.}
    \label{fig:comparable_active_parameter_scaling}
\end{figure}

\noindent\textbf{Active-Parameter Comparable Scaling.}
As illustrated in~\cref{fig:comparable_active_parameter_scaling}, we compare \textit{MoE 13B-A1.4B} (13B total parameters, 1.4B active parameters) against a standard \textit{Dense 1.3B} baseline under a similar active compute budget. The sparse model exhibits a substantial performance advantage in both training and validation losses throughout the training process. In video pre-training, modeling complex spatio-temporal features and continuous physical interactions requires a high representation capacity~\cite{ho2022video,peebles2023scalablediffusionmodelstransformers}. Under equivalent computational constraints, the tenfold increase in total parameter capacity of the sparse MoE model provides a significantly larger repository for physical-world priors, consistent with the capacity-compute decoupling enabled by sparse expert routing~\cite{shazeer2017moe,lepikhin2020gshard,switchtransformer}. This effectively resolves the feature-capacity bottleneck of the dense baseline, enabling the sparse model to achieve superior performance without inflating the compute budget.

\begin{figure}[htbp]
    \centering
    \begin{subfigure}[htbp]{0.49\linewidth}
        \centering
        \includegraphics[width=\linewidth]{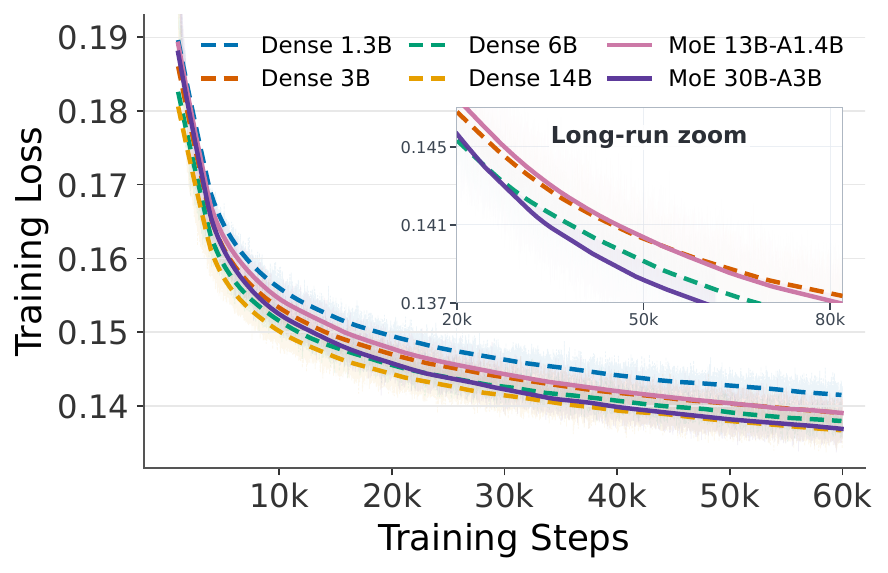}
        \caption{Dense and sparse scaling.}
        \label{fig:compute_comparable_dense_moe}
    \end{subfigure}
    \hfill
    \begin{subfigure}[htbp]{0.49\linewidth}
        \centering
        \includegraphics[width=\linewidth]{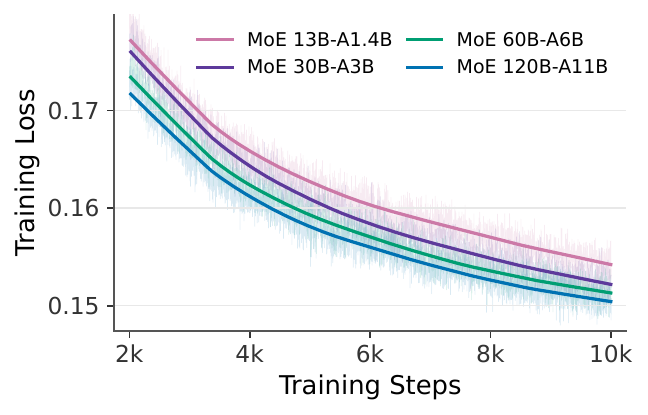}
        \caption{MoE scaling at aligned steps.}
        \label{fig:compute_comparable_moe_scaling}
    \end{subfigure}
    \caption{Training-loss comparison for compute-comparable scaling experiments. The faint background traces show the raw logged training losses; the bold curves are smoothed only for visualization and are not used to alter the underlying measurements.}
    \label{fig:compute_comparable_scaling}
\end{figure}

\noindent\textbf{Cross-Compute Dominance.}
Furthermore, our sparse architecture demonstrates striking cross-compute dominance over dense baselines. As shown in~\cref{fig:compute_comparable_dense_moe}, both \textit{MoE 13B-A1.4B} and \textit{MoE 30B-A3B} consistently outperform dense models with roughly twice their active parameter scale. Specifically, \textit{MoE 30B-A3B} closely approaches the performance of Dense 14B. This proves that capacity-compute decoupling fundamentally accelerates the scaling efficiency of video diffusion models, unlocking superior representation power.

\noindent\textbf{Predictable Scaling Laws up to 120B.}
To probe the scaling potential of our architecture under practical resource constraints, we conduct an early-stage scaling study by increasing the total parameter count from $13\,\mathrm{B}$ to $120\,\mathrm{B}$, with active parameters scaled proportionally from $1.4\,\mathrm{B}$ to $11\,\mathrm{B}$ (\textit{MoE 13B-A1.4B}, \textit{MoE 30B-A3B}, \textit{MoE 60B-A6B}, and \textit{MoE 120B-A11B}). As shown in~\cref{fig:compute_comparable_moe_scaling}, when the models are compared at aligned training steps, larger sparse models consistently achieve lower training loss, exhibiting a predictable scale-dependent improvement consistent with neural scaling-law observations~\cite{kaplan2020scalinglaws,hoffmann2022trainingcomputeoptimal}. Although these runs are not trained to full convergence due to resource constraints, the early training trajectories provide a practical initial validation of the scalability of our sparse single-stream design with auxiliary-loss-free load balancing and sequence-wise auxiliary loss.

\noindent\textbf{Inference Efficiency.} To verify that sparse capacity does not translate into prohibitive inference costs, we benchmark depth-matched dense and MoE DiT variants across sequence lengths ranging from 16K to 1M (1,048,576) tokens, following the sparse MoE literature's emphasis on conditional computation and routing overhead~\cite{shazeer2017moe,lepikhin2020gshard,switchtransformer,deepseek2024v3}.

\begin{wrapfigure}{r}{0.50\linewidth}
    \vspace{-0.8em}
    \centering
    \includegraphics[width=\linewidth]{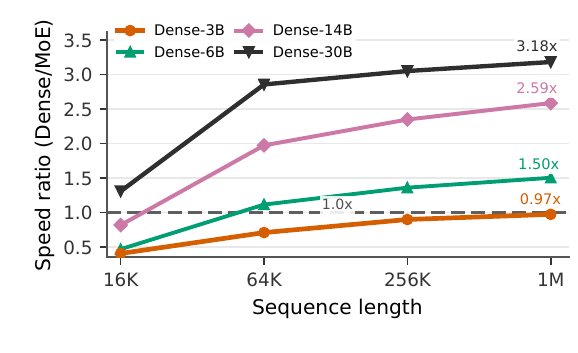}
    \caption{MoE-to-dense speed ratio, computed as dense latency divided by \textit{MoE 30B-A3B} latency.}
    \label{fig:moe_dense_inference_latency}
    \vspace{-1.0em}
\end{wrapfigure}

As illustrated in~\cref{fig:moe_dense_inference_latency}, we analyze the inference speed ratio, defined as $r=T_{\mathrm{dense}}/T_{\mathrm{MoE}}$, where $r > 1$ indicates that \textit{MoE 30B-A3B} is faster than the corresponding dense baseline.
We first compare the MoE model against its active-parameter equivalent, \textit{Dense 3B}, to assess routing overhead.
At a sequence length of 1M tokens, where attention and FFN computations dominate execution~\cite{williams2009roofline}, the sparse model achieves near-parity latency with a ratio of $0.97\times$.
Next, we evaluate the efficiency gains of sparsity against larger dense models.
As the parameter scale increases, the MoE architecture demonstrates substantial latency advantages: at 1M tokens, the speed ratios against \textit{Dense 6B}, \textit{Dense 14B}, and \textit{Dense 30B} reach $1.50\times$, $2.59\times$, and $3.18\times$, respectively.
These results demonstrate that our sparse framework successfully scales model capacity while preserving the inference efficiency of a 3B-scale model, offering a highly practical architecture for long-context video generation.

\subsection{Cascaded Refiner}
Directly generating high-resolution videos is computationally expensive due to the massive number of spatio-temporal tokens.
To balance the trade-off between computational complexity and generation quality, we adopt a cascaded design.
This follows the common practice of decomposing high-resolution diffusion generation into a base stage and one or more refinement or super-resolution stages~\cite{saharia2021sr3,ho2022imagen,blattmann2023stable}.
A high-capacity base generator first models the overall motion and scene layout at a lower resolution.
Subsequently, a dedicated second-stage refiner increases the resolution.
In practice, we upsample the video from 480p to 1080p.
During training, we simulate the base generator's outputs by downsampling the target video.
We then apply synthetic degradations, including Gaussian blur and compression, to construct a degraded low-resolution input.
Such synthetic degradation pipelines are widely used in blind restoration and real-world super-resolution to improve robustness to imperfect low-resolution inputs~\cite{zhang2021bsrgan,wang2021realesrgan}.
This degraded video is spatially upsampled to the target resolution in pixel space.
We then encode it using the VAE encoder to obtain the condition latent $\x_{\mathrm{lr}}$.
Processing the upsampled conditioning directly in the latent space avoids intermediate pixel-space decoding.
This reduces computational overhead while preserving visual quality~\cite{flashvideo,hunyuanvideo15,longcatvideo,waver}.

Instead of denoising from pure Gaussian noise, the refiner learns a conditional rectified flow starting from the degraded condition $\x_{\mathrm{lr}}$ toward the clean target latent $\x_0$ encoded from the target video.
This formulation is built on rectified flow and flow-matching objectives, which learn continuous transport dynamics through vector-field regression~\cite{liu2022rectifiedflow,lipman2022flow,tong2023conditionalflowmatching}.
During training, the refiner threshold $\tau \sim \operatorname{Uniform}(0.85, 0.95)$ defines the maximum noise level for the conditional trajectory.
We perturb $\x_{\mathrm{lr}}$ with Gaussian noise $\epsilon$ to form the noisy starting condition $\x_{\tau} = (1-\tau)\x_{\mathrm{lr}} + \tau\epsilon$.
For a sampled training timestep $t \in [0, \tau]$, the perturbed latent $\x_t$ and the target velocity $v^{\star}_{\mathrm{ref}}$ are formulated as:
\begin{equation}
    \x_t = \left(1-\frac{t}{\tau}\right)\x_0 + \frac{t}{\tau}\x_{\tau}, \quad v^{\star}_{\mathrm{ref}} = \frac{\x_{\tau}-\x_0}{\tau}.
\end{equation}
The model is optimized using the same flow-matching loss formulation as the base generator, restricted to timesteps $t \le \tau$.
This thresholded trajectory limits denoising to noise regimes near the degraded condition.
Consequently, the refiner preserves the base model's global semantics and motion.
Its capacity is dedicated to restoring high-frequency details, sharpening textures, and correcting local artifacts.

\begin{figure}[!t]
\centering
\includegraphics[width=\linewidth]{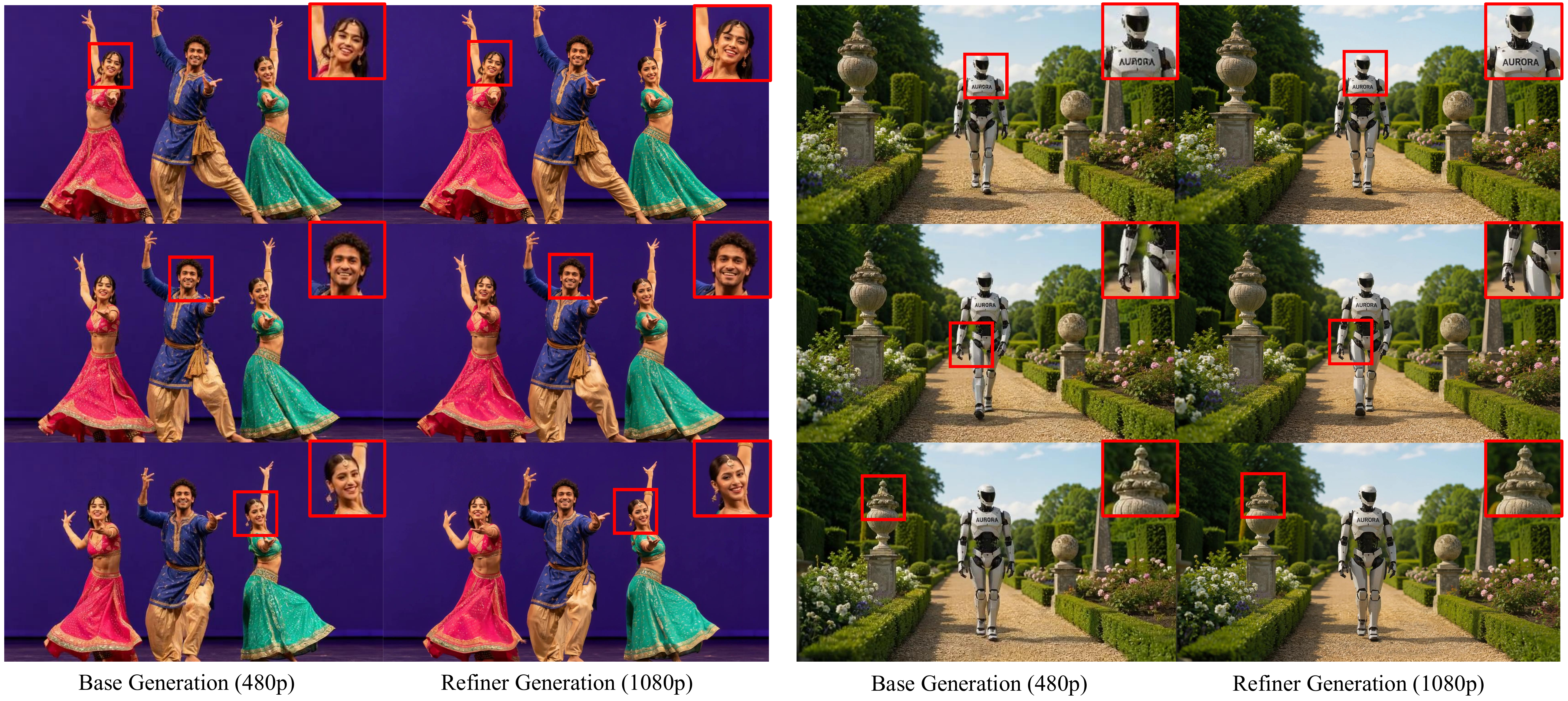}
\caption{\textbf{Refiner Generation.} 
We compare the base video generation against the refined video generation.
The left example's comprehensive prompt is  ``\textit{Three dancers, a man in the center and two women on either side, are performing a traditional Indian dance on a stage...}''.
The right example's comprehensive prompt is ``\textit{A humanoid robot named 'AURORA' walks steadily down a gravel path in a meticulously manicured formal garden...}''.}
\label{fig:refiner_generation}
\end{figure}

During inference, the low-resolution video generated by the base stage is upsampled to the target resolution in pixel space.
We then encode it to obtain the condition latent $\x_{\mathrm{lr}}$.
We perturb $\x_{\mathrm{lr}}$ with Gaussian noise $\epsilon$ at a starting timestep $t = \tau_{\text{inf}}$ (e.g., $\tau_{\text{inf}} = 0.85$).
This yields the noisy starting latent $\x_{\tau_{\text{inf}}} = (1-\tau_{\text{inf}})\x_{\mathrm{lr}} + \tau_{\text{inf}}\epsilon$.
Denoising is performed by integrating the rectified flow ODE trajectory backwards from $t = \tau_{\text{inf}}$ to $t = 0$, producing the clean video latent.
As shown in \cref{fig:refiner_generation}, the refiner significantly enhances face appearance (left), while successfully restoring high-frequency details and producing sharp, legible OCR text (right).

% !TEX root = ../main.tex
\section{Data} \label{sec:data}
The performance of video generation models is inextricably linked to the scale, quality, and diversity of their training data.
However, naively scaling up data collection yields diminishing returns and is often bottlenecked by practical factors such as acquisition costs and computational overhead.
To address these challenges, we have developed an integrated and scalable data infrastructure built through several synergistic modules.
The \textbf{Data Profiling Engine} extracts multi-dimensional attributes for each sample---covering structural, semantic, motion, camera, and quality aspects---providing a unified foundation for downstream processing.
The \textbf{World-Knowledge Topological Graph} organizes visual concepts into a hierarchical semantic structure, enabling distribution-aware sampling and enhancing long-tail coverage.
The \textbf{Dense Structured Captioning} module generates hierarchical textual descriptions to provide fine-grained semantic supervision, while a complementary \textbf{Caption Rewriter} maps brief user prompts into this same structured format during inference, thereby closing the prompt distribution gap between training and deployment.
Building upon this infrastructure, we organize the curated corpus into a stage-wise \textbf{Data Curriculum} (\cref{subsec:data_curriculum}), which governs the volume and composition of data consumed during each phase of progressive pre-training (\cref{sec:training}).

\subsection{Data Profiling Engine}
\begin{figure}[t]
\centering
\includegraphics[width=\linewidth]{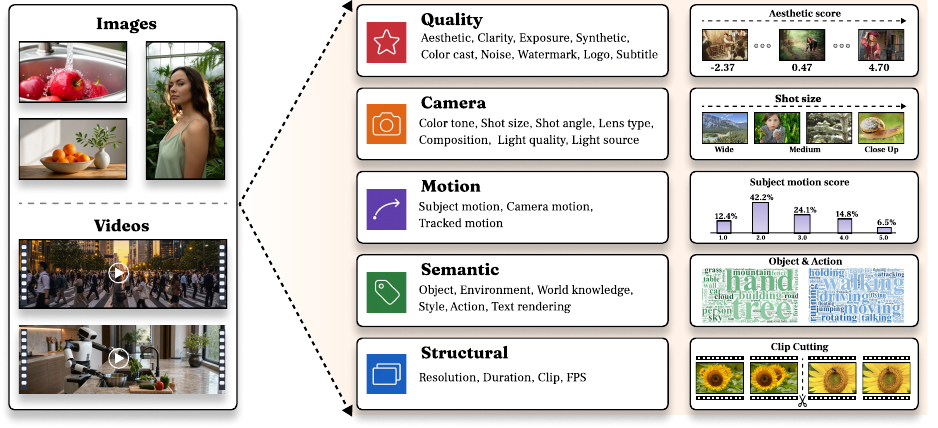}
\caption{\textbf{Overview of the Data Profiling Engine.} Each image or video sample is annotated across five complementary dimensions (structural, semantic, motion, camera, and quality) into a structured profile record, which then drives downstream filtering, balanced sampling, and captioning.}
\label{fig:profiling_engine}
\end{figure}

The Data Profiling Engine converts raw multimodal samples into structured, multi-dimensional records that capture structural, semantic, motion, camera, and quality attributes. Rather than relying on free-form annotations, we project every sample onto a fixed schema, providing a unified and queryable representation for heterogeneous image and video data. These standardized records drive all subsequent processing stages, from filtering and sampling to captioning. The core annotations are generated by powerful vision--language models (VLMs), further augmented by a suite of specialized scoring and detection models detailed below.~\cref{fig:profiling_engine} provides an overview of this pipeline.

\noindent\textbf{Structural Metadata.}
This dimension records fundamental media attributes, including spatial resolution, native frame rate, and duration. For video content, we employ TransNetV2~\cite{soucek2024transnet} to detect shot boundaries and segment the video stream into clip-level units, ensuring that each training clip represents a single, temporally coherent shot.

\noindent\textbf{Semantic Labels.}
Each sample is assigned semantic annotations along with category-level confidence scores, which serve as the foundation for the World-Knowledge Topological Graph used during distribution-aware sampling. Concurrently, the VLM annotator~\cite{Qwen3-VL,qwen3.6-27b} extracts structured tags to decompose the scene into its constituent elements: foreground and background objects, environmental settings, world-knowledge entities, visual styles, actions, and rendered text. This provides both a high-level holistic understanding of the scene and explicit entity anchors for precise data curation.

\noindent\textbf{Motion and Temporal Dynamics.}
For video data, the engine characterizes motion along three complementary signals: camera motion, subject motion, and tracked motion. The VLM annotator decouples camera motion from subject motion, rating the intensity of each. Complementing these VLM-based estimates, a geometry-grounded \emph{tracked-motion} signal derived from LocoTrack~\cite{cho2024local} point tracking screens out near-static or degenerate clips that would otherwise receive spurious motion scores. Together, these cues enable the downstream curation stage to effectively balance static and highly dynamic content.

\noindent\textbf{Camera and Cinematic Attributes.}
Each sample is further tagged with seven distinct cinematic attributes, generated by the VLM annotator distilled on cinematography data. These attributes include color tone, shot size, shot angle, lens type, composition, light quality, and light source (e.g., a warm-toned, medium-wide, high-angle shot lit by hard daylight). Making these attributes explicit facilitates fine-grained control over cinematic styles during generation and supports style-aware data balancing during curation.

\noindent\textbf{Quality and Aesthetic Signals.}
Visual quality is evaluated using both dedicated learned scorers and VLM judgments. We compute an aesthetic score using HPSv3~\cite{ma2025hpsv3}. Additionally, we estimate the likelihood that a sample is synthetic (AI-generated) using OmniAID~\cite{guo2025omniaid} to detect and down-weight generated imagery. Concurrently, the base pass of the VLM annotator provides ordinal ratings for clarity and exposure, along with binary artifact indicators that flag watermarks, logos, subtitles, color casts, and noise. Collectively, these signals inform the automated filtering and reweighting stage, ensuring that low-quality and artifact-laden samples are removed or down-weighted prior to training.

\begin{figure}[t]
\centering
\includegraphics[width=\linewidth]{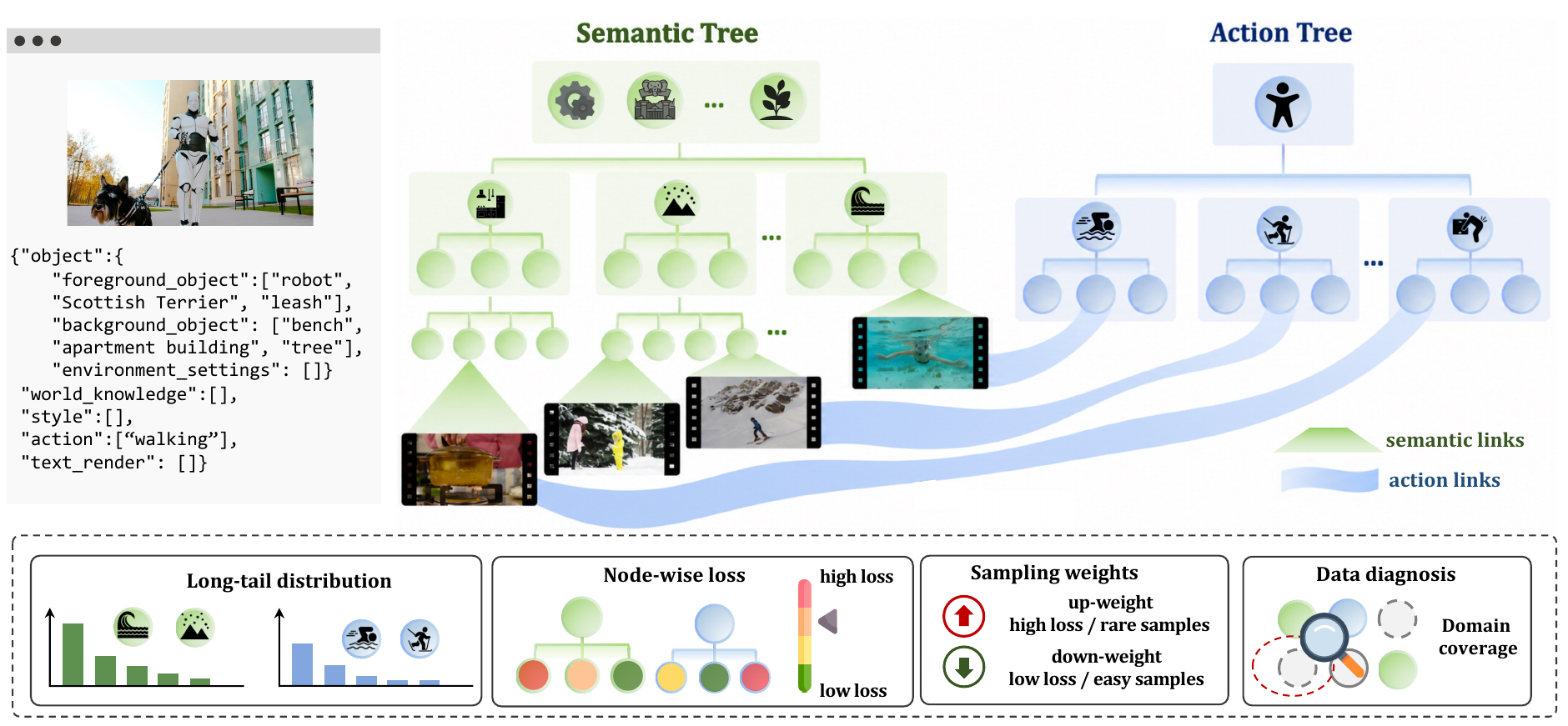}
\caption{\textbf{World-Knowledge Topological Graph.} Structured tags from the Data Profiling Engine link each sample to a semantic tree of visual concepts and, for videos with actions, to an action tree of dynamic behaviors. The graph serves as a control surface for data curation: node statistics and training-feedback signals are used to up-weight rare or difficult concepts, down-weight saturated easy modes, and identify under-covered domains for targeted data expansion.}
\label{fig:world_knowledge_graph}
\end{figure}

\subsection{World-Knowledge Topological Graph}

To make data curation distribution-aware rather than quality-aware only, we construct a World-Knowledge Topological Graph~\cite{cai2025z} that organizes samples along two complementary axes: a semantic concept tree and an action tree.
The semantic tree is shared by images and videos, providing a common vocabulary for objects, scenes, visual styles, and world-knowledge entities.
The action tree is attached only to video samples and captures the dynamic behaviors that are especially important for video pre-training, such as object manipulation, sports, daily activities, and human gestures.
Together, the two views allow us to query not only what appears in a sample, but also what happens in it.

\noindent\textbf{Semantic Concept Tree.}
We first extract semantic tags from the Data Profiling Engine, including foreground and background objects, environment settings, world-knowledge entities, styles, and rendered text (\cref{fig:world_knowledge_graph}).
Directly embedding these raw tags is unreliable because many tags are sparse, ambiguous, or dominated by proper nouns and domain-specific terms.
We therefore expand each tag with a short textual introduction that describes its visually grounded meaning, and then embed the concatenation of the tag and its introduction.
The resulting representation is clustered together with frequently used Wikipedia concepts~\cite{wikidump,Page1999ThePC} to form a large hierarchical inventory.
This hierarchy contains $50{,}000$ fine-grained leaf concepts and $1{,}000$ intermediate visual categories.
At lower levels, embedding-based clustering~\cite{vo2024automatic} provides high coverage and scalability.
At the top level, however, purely automatic clustering tends to disagree with human perceptual granularity; for example, visually similar categories may be separated by ontology names, while visually distinct categories may be merged because their text descriptions are semantically close.
We therefore use an LLM-assisted \textbf{Discover--Classify--Consolidate} procedure to merge the $1{,}000$ intermediate categories into $25$ visually coherent top-level groups.
The discovery stage proposes an initial flat taxonomy from representative categories, the classification stage assigns all intermediate categories while allowing bounded new-bucket proposals, and the consolidation stage merges overlapping buckets, renames unclear buckets, and absorbs under-populated ones.
This produces a human-aligned tree in which each sample is assigned to a fine-to-coarse semantic path.

\noindent\textbf{Action Tree.}
For videos, we additionally construct an action tree from the action tags produced by the profiler.
Because raw action tags are short, noisy, and redundant, we first normalize their surface forms and then expand each tag with a short VLM-generated description grounded in representative video snippets.
We embed the concatenation of the action tag and its description, and cluster the resulting representations into canonical action nodes.
This description-augmented clustering reduces ambiguity in short action phrases while grouping visually similar motion patterns.
After clustering, we apply lightweight review and correction to high-frequency or ambiguous nodes, and discard unreliable long-tail actions.
This produces several hundred canonical action nodes covering manipulation, human gestures, sports, and daily activities.
All samples are indexed by their semantic paths, while videos with reliable action annotations are additionally linked to one or more canonical action nodes, giving us a joint semantic-action profile.

\noindent\textbf{Distribution-Aware Sampling and Data Diagnosis.}
The graph is used as a control surface for the late-stage data curriculum.
During challenge-focused continual training, we up-weight rare or difficult semantic and action nodes, especially those related to manipulation, physical contact, and long-tail human activities, while down-weighting over-represented generic video nodes.
Beyond static graph statistics, we further close the loop with training feedback from the earlier pre-training stages.
After the first two stages, we aggregate the denoising loss by semantic nodes and, for video samples, by action nodes, obtaining a node-aware estimate of data difficulty that informs the sampling weights in subsequent stages.
Because every node maintains sample counts and representative examples, the graph also exposes coverage holes: missing actions and under-represented object categories.
We use these signals to guide targeted data acquisition and to rebalance the training mixture, improving long-tail diversity without blindly increasing corpus size.

\subsection{Dense Structured Captioning} \label{subsec:dense_captioning}
Following FIBO~\cite{gutflaish2025fibo}, which shows that training on long structured
captions substantially improves prompt adherence and controllability, we
annotate all training data with dense structured JSON captions. Our captions
cover four types of data---images, videos, VLA videos, and egocentric
videos. All four types share the same schema: video, VLA, and egocentric
captions simply add a few extra fields on top of the image schema.
\begin{itemize}
  \item \textbf{Image captions.} Each image caption has four parts: (1) a
  comprehensive description of the whole image; (2) a set of camera tags
  (color tone, shot size, shot angle, lens type, composition,
  light quality, and light source),
  each chosen from a fixed set of values; (3) an optional world-knowledge
  list for named entities that appear in the image; and (4) a list of
  prominent elements. For every element, the caption describes its location,
  relative size, shape and color, texture, relationship to other elements,
  and orientation. If the element is a person, the caption also describes
  pose, gender, skin tone, expression, and clothing; if it is a group of repeated objects, the
  caption marks it as a cluster and gives a rough count.
  \item \textbf{Video captions.} Video captions add temporal information on
  top of the image schema. The global description is written in two parts:
  what happens in the scene, and how the camera moves. In addition, every
  prominent element carries a list of timestamped actions, e.g., a hand
  enters the frame during $[2.67s, 3.67s]$ to place a food item. The caption
  therefore tells the model not only what is in the video, but also who does
  what, and when.
  \item \textbf{VLA captions.} Robot-manipulation videos use the same video
  schema. Robotic arms and grippers are described as prominent elements like
  any other object, and their timestamped actions split each episode into
  clear phases, e.g., moving down to grasp, opening the gripper to release,
  and retracting. Gender and skin-tone fields are dropped for this data.
  \item \textbf{Egocentric captions.} First-person videos also use the video
  schema. The camera-movement field describes the wearer's head and body
  motion, and the wearer's hands are annotated as prominent elements, with
  timestamped actions describing how they interact with objects. Gender and
  skin-tone fields are likewise dropped.
\end{itemize}
One example caption for each data type is provided in~\cref{app:caption}.

\subsection{Caption Rewriter} \label{subsec:caption_rewriter}
The generator is trained conditionally on the dense structured captions described in~\cref{subsec:dense_captioning}---which are lengthy, attribute-rich, JSON-formatted descriptions. During inference, however, users typically provide brief, free-form prompts. To bridge this train--inference distribution gap, we introduce a \emph{Caption Rewriter} that decomposes prompt expansion and formatting into a two-stage pipeline, avoiding the unreliability of a single-step mapping model.

In the first stage, \textbf{Expand}, a zero-shot large language model (e.g., Qwen3.6-27B~\cite{qwen3.6-27b}) processes the user prompt into a compact, action-centric natural-language description (under a thousand characters). This stage intentionally generates a concise, non-enumerative description that omits overly specific fine attributes---such as texture, exact colors, relative sizes, and precise camera parameters---which are highly prone to hallucination. The second stage, \textbf{Map}, utilizes a Qwen3.6-27B fine-tuned with LoRA~\cite{hu2021lora} to transform this intermediate prose into the final structured JSON caption. Although the Map stage populates the complete schema, it operates under a strictly \emph{bounded completion} constraint: it must preserve all explicit information from the prose verbatim (including scene layout, subjects, timestamped actions, and named entities) and is only permitted to plausibly impute low-risk missing fields. This strategy confines potential hallucinations and allows the two-stage rewriter to produce significantly more reliable and better-structured captions than a single model tasked with expanding and structuring a prompt simultaneously.

Both stages seamlessly support text-to-video (T2V) and text-image-to-video (TI2V) generation. For TI2V, a conditioning first frame is provided as the $t{=}0$ ground truth, ensuring the visual appearance adheres strictly to the image while motion dynamics follow the text. Additionally, a target duration is appended to every prompt to constrain all predicted action timestamps to fall within the video's duration.

\subsection{Data Curriculum} \label{subsec:data_curriculum}
\begin{figure}[t]
\centering
\includegraphics[width=\linewidth]{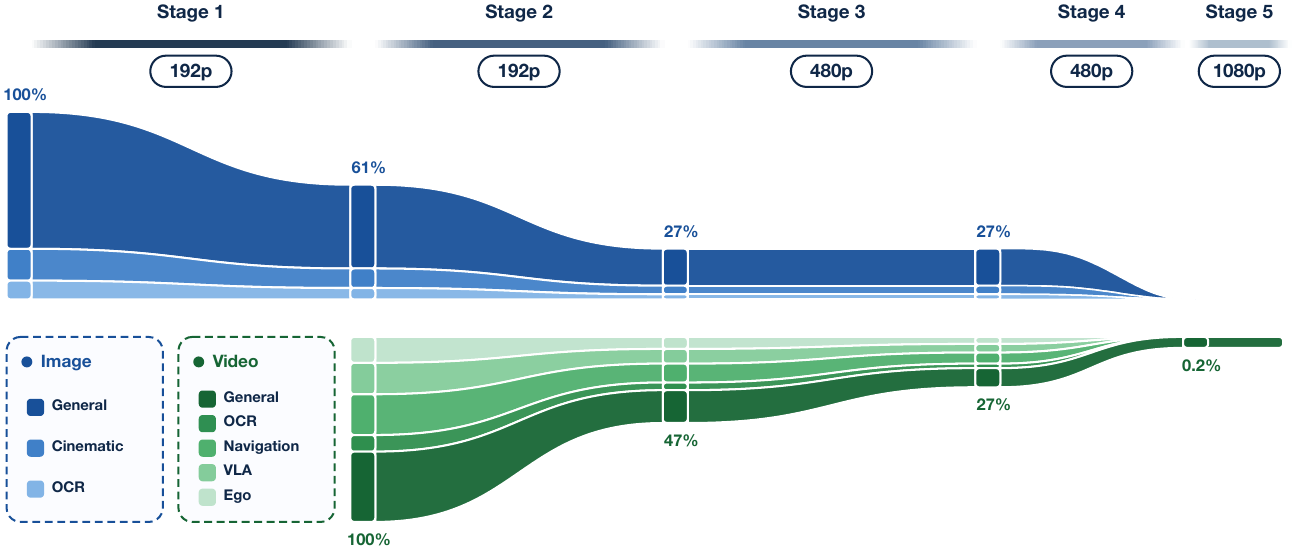}
\caption{\textbf{Data curriculum across the five progressive pre-training stages.} The image stream (blue) and video stream (green) are decomposed into their constituent sources, with band width indicating each source's relative proportion; percentages denote the fraction of data retained at each stage relative to the modality's initial pool.}
\label{fig:data_curriculum}
\end{figure}

Building upon the curated corpus, we organize the training data into a five-stage curriculum aligned with the progressive pre-training schedule (\cref{sec:training}); here we focus on how the data itself evolves across these stages. The modality mix and per-source composition shift systematically---from low-resolution (192p) image-only data, through joint image--video data with embodiment-rich footage at increasing resolution, to a small high-quality video refinement set. The cascaded refiner (\cref{sec:method}) is trained separately on a high-quality 1080p subset of this corpus.~\cref{fig:data_curriculum} summarizes how the modality mix and per-source proportions evolve across the five stages.

\noindent\textbf{Stage 1: Curated low-resolution image pool.}
The initial stage trains exclusively on 192p images, retaining the samples that pass aesthetic-quality and minimum-resolution filters while rigorously discarding the lowest-tier content.

\noindent\textbf{Stage 2: Video introduction and source expansion.}
Stage 2 introduces video content alongside images at a matching 192p resolution: a large corpus of video clips joins a quality-tightened, smaller pool of images. Video clips are admitted through a combination of resolution, aesthetic, and motion filters; notably, the motion criterion integrates a geometry-grounded tracking signal with a VLM-based motion assessment to robustly filter out near-static clips. Crucially, this stage marks the injection of over 70{,}000 hours of embodiment-oriented footage into the corpus---robot manipulation (VLA) spanning real-robot, simulated, open-source, and third-person perspectives across both humanoid and quadruped platforms, together with navigation and egocentric video---alongside text-rich video. Concurrently, the image filters are tightened to enforce higher aesthetic standards, a strict minimum resolution, and freedom from color cast, noise, and blur.

\noindent\textbf{Stage 3: Higher-resolution re-filtering.}
Stage 3 scales both the video and image streams to 480p, tightening the corresponding aesthetic and motion criteria proportionally; the higher-resolution bar substantially narrows the admitted data in both modalities. We purposefully retain high-motion video to strengthen coverage of dynamic content, while holding the image stream to the rigorous quality gates established in Stage~2.

\noindent\textbf{Stage 4: Challenge-focused rebalancing.}
At the same 480p resolution, this stage employs source-aware curation to align the corpus with the demands of embodied intelligence. Abundant general and curated videos are subjected to stringent quality standards covering resolution, aesthetics, clarity, motion, and freedom from artifacts such as watermarks, whereas scarce yet high-value embodiment sources, including manipulation, navigation, egocentric footage, and benchmark datasets, undergo only minimal filtering to maximize their coverage. This deliberate asymmetry trades redundant general footage for the long-tail, action-centric data that is paramount for embodied intelligence, yielding a mixture far more embodiment-heavy than the raw source counts would suggest.

\noindent\textbf{Stage 5: High-quality refinement subset.}
The final stage is a small, high-quality video refinement set at 1080p, used to train the cascaded refiner (\cref{sec:method}): a curated subset drawn from the general source, with samples held to the strictest aesthetic, resolution, and technical-quality bars, and only a tiny fraction (well under 1\% of the initial video pool) retained.

% !TEX root = ../main.tex
\section{Infrastructure} \label{sec:infrastructure}

\subsection{Pre-Training Infrastructure}
Large-scale video pre-training imposes distinct system bottlenecks compared to image-only or language-only workloads~\cite{ho2022imagen,blattmann2023stable,wan2025wan,chowdhery2022palmscalinglanguagemodeling}.
The training pipeline must ingest heterogeneous image and video streams, manage variable sequence lengths dictated by resolution and duration.
To achieve high training throughput and memory efficiency, our pre-training infrastructure is co-designed to address six core challenges: token-budgeted data loading, composable multi-dimensional parallelism, activation-memory management, compile-first graph capture, explicit communication prefetching, and non-blocking runtime I/O~\cite{shoeybi2019megatronlm,narayanan2021efficient,rajbhandari2020zero,zheng2022alpa,liang2025torchtitan,ma2025veomni}.

\subsubsection{Heterogeneous Data Pipeline and Token-Budgeted Packing}
Rather than batching by a fixed number of samples, the data pipeline constructs each mini-batch based on a target token budget. The dataset estimates both visual and conditioning token lengths under the sampled configuration, allowing the sampler to dynamically decide whether to append the sample based on the remaining token capacity of the batch. This online length-aware scheduling is critical because both sides of the packed input are highly variable: visual tokens come from images and videos with diverse native resolutions, aspect ratios, and durations, while conditioning tokens vary across multimodal conditions; together, these factors produce orders-of-magnitude differences in token counts.

\noindent\textbf{Packed one-dimensional batch.}
Following VAE and condition encoding, each sample yields visual tokens $x_i$ and conditioning tokens $y_i$.
Rather than stacking heterogeneous samples into a rectangular tensor, we concatenate all valid segments into a single one-dimensional sequence:
$[x_1, y_1, x_2, y_2, \ldots, x_N, y_N]$.
The data loader constructs packed sequence metadata, such as cumulative sequence lengths and attention masks, enabling variable-length attention kernels (e.g., FlashAttention~\cite{dao2022flashattention}) to process the packed batch in a single forward pass while preventing cross-sample attention leakage.
This unified representation allows images, videos, and different conditioning modalities to be processed jointly without partitioning dataloaders by modality or resolution.
To maximize token utilization and eliminate padding waste, a smart-fill sampler dynamically searches rank-local candidate pools to fill any residual token budget of each packed batch when the next video clip exceeds the leftover limit~\cite{krell2021efficientsequencepacking}.

\subsubsection{Composable Parallel Training}
We organize the distributed training stack into four composable parallel dimensions: data parallelism (DP)~\cite{li2020pytorchdistributed,shoeybi2019megatronlm}, fully sharded data parallelism (FSDP)~\cite{rajbhandari2020zero,zhao2023pytorchfsdp}, sequence parallelism (SP)~\cite{jacobs2023deepspeed}, and expert parallelism (EP)~\cite{shazeer2017moe,lepikhin2020gshard,rajbhandari2022deepspeedmoe,hwang2022tutel}.
The training system represents these dimensions via named multi-dimensional device meshes, allowing each training run to configure and compositionally select the required parallel modes without altering the core training loop.
This unified abstraction integrates throughput scaling, model-state sharding, long-sequence context partition, and MoE token routing under a single parallelization plan~\cite{shazeer2018mesh,xu2021gspmd,zheng2022alpa,narayanan2021efficient,liang2025torchtitan,ma2025veomni}.

\noindent\textbf{Data parallelism (DP).}
DP serves as the primary outer dimension for throughput scaling.
Each DP process group consumes a distinct shard of the token-budgeted data stream, while gradients, loss metrics, and optimizer states are synchronized globally.
For hybrid data sharding, the data-parallel mesh is decomposed into replication and sharding dimensions, allowing cross-node replication and node-local parameter sharding to be configured independently~\cite{li2020pytorchdistributed,shoeybi2019megatronlm,rajbhandari2020zero}.

\noindent\textbf{Fully sharded data parallelism (FSDP).}
We utilize FSDP to shard parameters, gradients, and optimizer states across the sharding dimension, extending to Hybrid Sharded Data Parallelism (HSDP) for multi-node scalability.
FSDP serves as the primary model-state scaling mechanism for dense transformer blocks and non-expert modules~\cite{rajbhandari2020zero,zhao2023pytorchfsdp}.
FSDP sharding is applied downstream of activation checkpointing and compilation decisions, ensuring that sharding hooks, mixed-precision policies, and high-precision ignored parameters are resolved in a deterministic sequence.
This ordering is critical to keeping numerically sensitive parameters (e.g., normalizations, routing scales) replicated in FP32 while compiling the surrounding tensor computations in lower precision~\cite{dehghani2023vit22b}.

\noindent\textbf{Sequence parallelism (SP).}
To support long-context video sequences, we integrate Ulysses sequence parallelism (SP)~\cite{jacobs2023deepspeed}.
Before the transformer blocks, the packed sequence is padded to align with the Ulysses group size and sliced along the token dimension.
Within the attention block, all-to-all collectives transpose token shards into head shards for multi-head attention computation, and a subsequent all-to-all collective restores the sequence-sharded layout.
Finally, the sequence is gathered and unpadded to reconstruct the original one-dimensional packed layout.
This makes sequence length a distributed resource across the cluster rather than a single-accelerator memory limit.

\noindent\textbf{Expert parallelism (EP).}
EP scales sparse MoE layers by distributing the expert pool across ranks and routing each token to the devices that host its selected experts~\cite{shazeer2017moe,lepikhin2020gshard,switchtransformer,rajbhandari2022deepspeedmoe,hwang2022tutel,deepseek2024v3}.
During the MoE forward pass, routed tokens are dispatched through all-to-all communication, processed by local experts, and then combined back to their original sequence positions, preserving global routing semantics without requiring every GPU to host every expert.
In our implementation, the expert dispatch and combine path is accelerated with DeepEP~\cite{deepep2025}.

\subsubsection{Activation Checkpointing}
Activation checkpointing manages the memory footprint of long packed video sequences by recomputing selected forward activations during the backward pass~\cite{chen2016training}.
The system supports multiple checkpointing granularities: full block-level recomputation, layer-level selective checkpointing, operation-level selective checkpointing, and memory-budget-driven checkpointing~\cite{korthikanti2022reducingactivationrecomputationlarge}.
This granularity is configurable, enabling the training recipe to dynamically balance recomputation overhead and memory constraints across different model scales and parallel configurations.
Crucially, the checkpointing path is implemented as non-reentrant to prevent lifetime conflicts between activation recomputation and expert communication collectives.

\subsubsection{Compile-First Graph Capture}
Rather than serving as an optional compatibility feature, graph compilation is a key throughput optimization.
The training pipeline applies \texttt{torch.compile} with the Inductor backend before FSDP2 sharding, allowing local block computations to be fused before sharding hooks are attached~\cite{ansel2024pytorch2}.
In our internal training benchmark, combining full-block activation checkpointing with compilation improves Model Flops Utilization (MFU) by approximately $1.9\times$, indicating that compile-first graph capture recovers a substantial fraction of accelerator utilization otherwise lost to unfused block-level execution.

\subsubsection{Asynchronous Monitoring and Distributed Checkpoint I/O}
Runtime I/O operations are co-designed with the training loop to prevent metric logging and state serialization from stalling the accelerators.
The monitoring pathway is fully asynchronous: training ranks enqueue logging events (e.g., scalars, histograms, plots) onto a bounded queue, which are processed and written to disk by a background thread.
GPU tensors are transferred to host memory before queueing, and the bounded queue prevents logging backpressure from stalling computation.
This non-blocking design allows the system to monitor comprehensive diagnostic metrics—such as step-time breakdowns, MFU, MoE load balance, and routing statistics—without impacting training throughput~\cite{moritz2018ray}.

\subsection{Post-Training Infrastructure}
Existing reinforcement-learning training frameworks are largely designed around language or vision-language models.
In these settings, trajectories are naturally represented as token sequences, and policy log-likelihoods can usually be obtained from per-token probabilities.
Around this assumption, prior work has built mature RLHF/RLVR objectives and systems, including PPO, DPO, and GRPO, as well as infrastructure that decouples large-model training, generation, and reward computation~\cite{
    schulman2017ppo,
    rafailov2023dpo,
    shao2024deepseekmath,
    sheng2024hybridflow,
    hu2024openrlhf,
    mei2024realhf,
    slime2025,
    zhang2026relax}.
Video diffusion model post-training violates these token-centric assumptions: its optimization target is tightly coupled with the denoising process or latent trajectory, its intermediate states are high-dimensional video latents rather than lightweight text tokens, and log-probability computation, credit assignment, and reward evaluation differ substantially from language modeling.

\noindent Recent work such as Flow-GRPO, DanceGRPO, DiffusionNFT, and AWM has explored reinforcement-learning post-training for diffusion models~\cite{
    liu2025flowgrpo,
    xue2025dancegrpo,
    zheng2026diffusionnft,
    xue2025awm}.
Similar RL-based post-training ideas have also started to appear in recent frontier generative-model technical reports~\cite{
    hunyuanvideo15,
    longcatvideo}, suggesting that RL is becoming an increasingly important component of diffusion model post-training.

\subsubsection{Diffusion-Native RL System Design}
RL post-training for large MoE video generation models places exceptionally heavy demands on infrastructure.
A single training sample corresponds to a sequence on the order of 100K tokens; the intermediate states of video RL---latent trajectories and per-step sampling statistics---reach the gigabyte scale, far larger than text tokens; and the sheer parameter count of MoE models puts enormous pressure on both weight synchronization and memory management.

\noindent We therefore design a diffusion-native RL infrastructure for video diffusion models.
The system organizes conditioning, latent trajectories, rewards, and transition-level training data under a unified data abstraction; supports GRPO-style reverse-process optimization and forward-process objectives; and is compatible with both LoRA-style parameter-efficient finetuning and full-model finetuning~\cite{hu2021lora}.
To handle the memory and communication pressure introduced by large intermediate video states, the infrastructure decouples rollout, reward evaluation, and training into separate execution roles.
It uses a communication-aware data abstraction for large latent objects and a service-oriented reward layer for heterogeneous reward models, server-side decoding, and request batching.
This design keeps the RL stack compatible with existing diffusion pipelines while making large-scale video diffusion model post-training more efficient~\cite{
    von-platen-etal-2022-diffusers,
    moritz2018ray}.

\subsubsection{RL System Performance}

\noindent The infrastructure is also heavily optimized for speed: full-parameter weight synchronization of the 30B model completes in 20 seconds per step; gigabyte-scale intermediate states are exchanged between rollout and training within 50 milliseconds across multiple GPU nodes; and the system sustains an end-to-end MFU of 43.9\% over the full RL step.

\subsection{Serving Infrastructure}
\method is served through a Diffusers-compatible model package and an SGLang Diffusion runtime~\cite{von-platen-etal-2022-diffusers,zheng2024sglang}.
The serving stack is designed to satisfy three practical requirements: a portable model package for open deployment, a numerically aligned reference path for tolerance-based regression testing, and an optimized multi-GPU path for long-video generation.
The same runtime supports text-to-image (T2I), base text-to-video (T2V), and text-image-to-video (TI2V) generation and can optionally invoke a second-stage refiner after base video generation.

\subsubsection{Diffusers-Compatible Model Package}
To maximize accessibility and lower the deployment barrier for the open-source community, we organize our release artifact as a Diffusers-compatible model package~\cite{von-platen-etal-2022-diffusers}. Prioritizing compatibility with standard Diffusers APIs ensures that developers and researchers can deploy the model out of the box using generic PyTorch environments, without requiring complex software configurations or custom binary compilation.
Under this packaging scheme, the model is organized as a unified root directory where the base DiT and the refiner DiT are stored as independent components.
Both stages share the same conditioning and autoencoder weights.
A lightweight overlay builder constructs the runtime view expected by Diffusers and SGLang by dynamically exposing the selected DiT component as the active \texttt{transformer} via configuration files.
This design allows the base and refiner serving to reuse the package without duplicating shared weights or copying weight tensors, keeping the release artifact compact and clean.

\subsubsection{SGLang-Native Serving Backend}
To optimize serving efficiency and maximize inference throughput, we build our deployment runtime on top of SGLang~\cite{zheng2024sglang}, a widely adopted acceleration engine renowned for its performance in serving diffusion-based pipelines. Leveraging SGLang allows our deployment stack to tap into highly optimized CUDA kernels and distributed scheduling policies.
Within this framework, the serving runtime provides three distinct execution paths: a direct Diffusers path, a generic SGLang Diffusion backend, and a specialized \method-native adapter.
The direct Diffusers path serves as a readable, eager-mode reference and debugging baseline for checking scheduler modifications and model behavior under a controlled numerical setting.
The \method-native adapter keeps the Diffusers-style denoising loop closely aligned with this reference path while registering our custom pipeline directly into SGLang's serving surface and distributed execution hooks.
This architectural separation allows the core denoising logic to remain stable and easy to audit, while enabling SGLang's optimized compute kernels and distributed execution policies to be applied transparently behind the same model interface.

\subsubsection{Distributed Video Serving}
To accommodate diverse production requirements, our SGLang deployment framework provides two recommended inference configurations: sharding long-video token sequences across multiple GPUs via context parallelism and utilizing batched classifier-free guidance (CFG) to evaluate conditional and unconditional branches efficiently.
Depending on the deployment objective, users can select between the following two specialized execution modes:
\begin{itemize}
    \item \textbf{Fidelity-First Version:} Designed for regression testing and numerical-consistency checks, this baseline configuration follows the master training codebase as closely as possible in scheduler logic, routed-expert execution, and precision policy. It preserves standard grouped matrix multiplication (grouped GEMM) for routed experts, employs vectorized token padding and restoration, and maintains a hybrid precision layout. Specifically, it executes standard transformer layers and text encoders in BF16, while keeping numerically sensitive parameters—such as normalization layers, routing gates, and scale-shift modulation tables—in FP32. In practice, this mode serves as a conservative reference gate by keeping implementation-induced deviations within expected numerical tolerance, rather than prioritizing maximum throughput.
    \item \textbf{Speed-First Version:} Tailored for high-throughput serving, rapid visual screening, and low-latency interactive applications, this profile optimizes compute-heavy bottlenecks while maintaining the same scheduler and model interfaces. It replaces the standard routed expert execution path with highly optimized FP8 SGLang Triton kernels, significantly reducing memory footprint and kernel execution times. Combined with sequence sharding and parallelized guidance evaluation, this accelerated version provides substantial throughput improvements with limited observed impact on video generation quality.
\end{itemize}

% !TEX root = ../main.tex
\section{Training} \label{sec:training}

\subsection{Progressive Pre-Training}
Large-scale video pre-training is highly challenging to optimize when high-resolution videos, multiple conditioning modalities, and heterogeneous data sources are introduced simultaneously from the beginning of training~\cite{ho2022imagen,blattmann2023stable,wan2025wan}. For a sparse Mixture-of-Experts (MoE) diffusion model, this challenge is compounded by the need for the router to establish specialized expert pathways across different tasks and modalities~\cite{shazeer2017moe,switchtransformer,dai2024deepseekmoe}. Exposing the model to full-scale training complexity from the outset may therefore increase the risk of optimization instability, routing collapse, and suboptimal sample quality~\cite{switchtransformer,park2024switch}.

To address these challenges, we design a progressive pre-training curriculum that introduces learning objectives in an optimization-friendly order~\cite{ho2022imagen,blattmann2023stable}. The curriculum first builds stable frame-level visual priors, then adds temporal modeling, expands task conditioning, harmonizes the data distribution, and finally refines high-resolution details. This staged design separates static appearance learning from dynamic evolution and allows the sparse router to specialize gradually rather than being exposed to all sources of heterogeneity at once~\cite{singer2022make,guo2023animatediff}. We organize the curriculum into five stages, each adding one source of training complexity to the previous stage.

\noindent\textbf{Stage 1: Image-Only Prior Acquisition.}
The first stage establishes fundamental text-visual semantic alignment and robust single-frame visual priors~\cite{chen2023pixart,esser2024scaling}. By formulating images as single-frame ($T=1$) video sequences within our unified framework, we enable the model to learn object shapes, textures, and scene aesthetics without the added complexity of temporal dynamics~\cite{singer2022make,guo2023animatediff}. Crucially, this image-first warmup provides a stable optimization environment for the sparse router, allowing it to initialize effectively before encountering heterogeneous video tasks~\cite{switchtransformer,dai2024deepseekmoe}.

\noindent\textbf{Stage 2: Low-Resolution Temporal Learning.}
Stage 2 introduces video data into the image-video training mixture while deliberately maintaining a simplified task formulation. Image samples continue to reinforce frame-level visual priors, while video samples are trained with text-to-video (T2V) generation so that the model can focus on temporal dynamics, camera motion, and frame-to-frame consistency before addressing multi-condition formatting~\cite{ho2022video,bar2024lumiere}. This phase adapts static visual priors into coherent temporal representations without abruptly escalating the optimization difficulty~\cite{singer2022make,guo2023animatediff}.

\noindent\textbf{Stage 3: Multi-Task Conditioning.}
Stage 3 keeps image samples in the training mixture and broadens the video-side task definition from pure T2V to a joint mixture of T2V and text-image-to-video (TI2V) generation~\cite{blattmann2023stable,wan2025wan2}. The primary objective is to teach the model to effectively leverage visual conditions: in the TI2V task, the model must faithfully preserve the provided initial frame while predicting temporally coherent future frames~\cite{esser2023structure,wan2025wan2}. By introducing this requirement only after foundational temporal learning has stabilized, we avoid conflating early motion acquisition with the more challenging constraint of visual condition preservation.

\noindent\textbf{Stage 4: Weighted Distribution Harmonization.} 
Stage 4 focuses on optimizing the data distribution. Large-scale web data exhibits severe quality variance and source imbalance, which can degrade training stability and expert specialization in late-stage pre-training. We therefore switch to a weighted sampler, which samples from high-value and high-quality data sources according to predefined sampling weights via alias tables. This stage lets the model stabilize under a cleaner and more balanced mixture before entering the final refinement phase.

\noindent\textbf{Stage 5: High-Resolution Refinement.}
The final stage functions as a high-resolution refiner, trained on video data. Its primary purpose is to dedicate training capacity to the generation of high-frequency spatial details and the correction of local artifacts in high-resolution samples, all while maintaining temporal sharpness in videos (e.g., minimizing texture flickering)~\cite{saharia2021sr3,ho2022imagen,flashvideo}. Following the cascaded refiner configuration described in~\cref{sec:method}, this stage optimizes a conditional refinement trajectory. It maps upsampled low-resolution base-stage outputs to clean high-resolution targets, ensuring the model retains the base-stage semantics and motion dynamics while successfully restoring fine-grained spatial details~\cite{blattmann2023stable,flashvideo}.

\subsection{Post-Training}
% \todo{jiaqi,chaoran,zijing,zichen}
\subsubsection{Reinforcement Learning with Multi-Aspect Rewards}
\label{sec:training:reward}
\noindent \textbf{Reward Modeling.} 
Existing reinforcement learning (RL) methods in post-training for visual generation typically rely on holistic reward models that output a single scalar score to represent overall human preference or general text-alignment~\cite{xu2023imagereward,kirstain2023pickscore,tang2026enhancing}.
However, video generation is inherently multi-dimensional and prone to complex failure modes, such as static mode collapse, temporal hallucinations, and physical implausibility, which global, coarse-grained rewards fail to penalize effectively.
To address this and provide fine-grained optimization signals, we decouple the evaluation into a comprehensive suite of six specialized reward models to optimize visual quality and physical dynamics during the training process: vision quality, text-video alignment, dynamic degree, motion coherence, human motion consistency, and physical plausibility.
\begin{itemize}
    \item \textbf{Vision Quality.} Following LongCat-Video~\cite{longcatvideo}, we employ HPSv3~\cite{ma2025hpsv3} as the foundation for VQ evaluation to jointly assess visual quality and text-video alignment. We leverage different statistics over all frames in the video to design complementary reward signals that separately capture overall visual fidelity and robust caption alignment. The resulting composite reward penalizes blurriness, artifacts, and low-resolution outputs while remaining sensitive to caption-level alignment.
    
    \item \textbf{Text-Video Alignment.} 
    We propose a fine-grained, action-centric text-video alignment reward based on temporal Visual Question Answering (VQA) to evaluate text-video alignment. 
    We first parse the training caption into structured entities, associating each with specific actions and precise temporal windows. To ensure evaluation accuracy while reducing computational overhead, we introduce an adaptive temporal slicing strategy with a bifurcated windowing mechanism: dynamic actions receive padded windows, whereas static-temporal criteria enforce strict boundaries. We then employ Qwen3.6-27B~\cite{qwen3.6-27b} as a zero-shot evaluator to answer batched verification queries on the sliced frames. 
    The text-video alignment reward is computed as a weighted satisfaction rate normalized to $[0, 1]$, where weights are dynamically assigned based on action complexity. 
    Unlike traditional global semantic matching metrics, this fine-grained approach ensures that accurately generating multi-stage actions yields a significantly higher reward, effectively penalizing temporal hallucinations and missing actions.

    \item \textbf{Dynamic Degree.} 
    We introduce a Dynamic Degree reward evaluated by Qwen3.6-27B~\cite{qwen3.6-27b} to measure and encourage appropriate motion intensity. 
    Specifically, the generated video is spatially downsampled to a maximum height of 480p for efficiency and evaluated using a specialized motion-assessment prompt. 
    The VLM outputs a structured information containing a discrete motion score from 1 to 5, which we linearly map to a continuous reward scalar in $[0, 1]$. 
    By relying on a VLM rather than traditional optical flow, this reward captures semantically meaningful subject dynamics. This counteracts the tendency of text-to-video models to generate static, image-like sequences, effectively breaking the static mode collapse without compromising temporal consistency.

    \item \textbf{Motion Coherence.}
    Text-to-video models generate videos at a fixed playback frame rate (e.g., 24\,fps), yet the generated motion often suffers from a slow-motion effect: it looks slower than it would in the real world, as if captured by a high-speed camera but played back at 24\,fps.
    Following Pulse-of-Motion~\cite{gao2026pulse-of-motion}, we estimate from the generated frames alone how fast the motion actually unfolds.
    The reward mechanism is designed to guide the model toward generating motion that appears natural at a standard 24 fps playback rate. It fully rewards content exhibiting realistic physical speed while penalizing outputs that display an artificial slow-motion effect, thereby steering the model to produce videos with naturally paced dynamics.
    
    \item \textbf{Human-Motion Consistency.} 
    We train a generative human-motion consistency model using a comprehensive distillation approach to evaluate human motion. 
    We build a high-quality dataset of real and synthesized videos to generate five-dimensional artifact scores, specifically targeting impossible topology, facial distortion, hand deformity, limb count errors, and semi-transparent bodies, along with step-by-step reasoning to assess both spatial correctness and temporal fidelity. 
    We subsequently fine-tune a vision-language mixture-of-experts model~\cite{qwen3.6-27b} on this curated corpus. 
    By enforcing supervision over the explicit reasoning traces, the model transcends superficial scalar regression, thereby internalizing the underlying structural and semantic constraints. During the reinforcement learning stage, it outputs structured rationales and discrete scores, which are mapped to a continuous reward scalar in $[0, 1]$. 
    This mechanism explicitly addresses the spatial distortions and temporal-semantic misalignments, guiding the policy toward physically plausible and text-aligned human motion.

    \item \textbf{Physical Plausibility.}
    We introduce a Physical Plausibility reward to assess whether generated video trajectories unfold within a coherent physical scene, where task-relevant entities remain present, spatially grounded, and consistent with the intended task evolution. 
    Rather than measuring only perceptual quality, this evaluator is caption-conditioned and frame-evidence-based: it establishes the expected actors, target objects, and final-state cues from the caption, but assigns scores solely based on visible phenomena in the sampled frames.
    It first assesses foundational physical constraints along three complementary axes: motion causality, which verifies whether objects remain stationary or continue moving unless affected by a plausible external force; object permanence and non-penetration, which checks for boundary loss, penetration, impossible overlap, and unsupported appearance or disappearance; and material-kinematic realism, which examines whether entities follow plausible material behavior and rigid-body motion.
    Building upon these physical foundations, the evaluator also incorporates an assessment of task completion. It verifies that the intended physical state changes actually materialize by checking for correct action occurrence, accurate object manipulation, and the successful achievement of the specified final spatial state.
    The resulting scores are aggregated into a unified physical reward. By jointly evaluating adherence to fundamental physical laws and the realization of task-driven state changes, this design ensures that generated videos maintain a coherent, physics-based world where dynamic manipulations and state changes are both visually grounded and physically plausible.
\end{itemize}

\noindent \textbf{Reward Aggregation.}
The six reward signals are aggregated into a single advantage through decoupled per-reward normalization, described in~\cref{sec:training:grpo}.

\subsubsection{On-Policy GRPO Training}
\label{sec:training:grpo}

We post-train {\method} with Group Relative Policy Optimization (GRPO)~\cite{shao2024deepseekmath,liu2025flowgrpo,xue2025dancegrpo} to maximize the multi-aspect rewards. Our setup follows the single-step exploration paradigm of Flash-GRPO~\cite{he2026flashgrpo}: each rollout is stochastic at exactly one denoising step shared within a group, the noise is injected via Coefficients-Preserving Sampling (CPS)~\cite{wang2025cps}, the policy gradient is reweighted to balance timesteps, and training is strictly on-policy without KL regularization. We detail each component below.

\noindent\textbf{Group-Shared Single-Step Exploration.}
GRPO estimates advantages by comparing a group of $G$ rollouts of the same prompt, which requires stochastic sampling. In Flow-GRPO and DanceGRPO~\cite{liu2025flowgrpo,xue2025dancegrpo}, every denoising step is stochastic and all steps share the same trajectory-level advantage, which causes a severe credit-assignment problem~\cite{he2025tempflow,li2025mixgrpo,longcatvideo}. Following Flash-GRPO~\cite{he2026flashgrpo}, we instead make exactly one denoising step stochastic. For each group, a step index $k$ is drawn uniformly from the first half of the inference schedule and shared by all $G$ rollouts. The transition $t_k \to t_{k+1}$ is sampled stochastically, while all other steps remain deterministic ODE steps. Each trajectory then contains exactly one stochastic transition, advantages within a group are compared at the same timestep, and the policy update is applied exactly to the transition that produced the variation, giving precise credit assignment.

\noindent\textbf{Coefficients-Preserving Sampling.}
For the stochastic step we adopt CPS~\cite{wang2025cps}, a DDIM-style transition~\cite{song2020ddim}, instead of the common SDE conversion~\cite{liu2025flowgrpo}. The Euler--Maruyama step of the converted SDE injects more noise than the schedule prescribes~\cite{wang2025cps}, leaving visible artifacts that corrupt reward scores, and its diffusion coefficient explodes at high noise levels, requiring an extra clipping patch~\cite{longcatvideo}; CPS avoids both by construction. Let $1 = t_0 > t_1 > \cdots > t_N = 0$ be the inference schedule. At step $t_i$, the velocity prediction $\hat{v}_\theta$ gives the clean and noise estimates:
\begin{equation}
\hat\x_0 = \x_{t_i} - t_i \hat{v}_\theta, \qquad \hat{\epsilon} = \x_{t_i} + (1 - t_i) \hat{v}_\theta.
\end{equation}
The CPS transition to $t_{i+1}$ is:
\begin{equation}
\x_{t_{i+1}} = \mu_\theta + s_i \epsilon, \qquad
\mu_\theta = (1 - t_{i+1}) \hat\x_0 + \sqrt{t_{i+1}^2 - s_i^2}\, \hat{\epsilon},
\label{eq:cps}
\end{equation}
where $\epsilon \sim \mathcal{N}(0, \mathbf{I})$ is fresh Gaussian noise, $s_i = t_{i+1} \sin(\eta\pi/2)$ is the injected noise scale, and $\eta \in [0, 1]$ controls the exploration strength ($\eta = 0.7$ in our experiments). Since $(t_{i+1}^2 - s_i^2) + s_i^2 = t_{i+1}^2$, the total noise after the transition matches the schedule coefficient exactly: samples stay on the marginal path of the deterministic sampler, and rewards are always evaluated on clean videos. Following~\cite{wang2025cps}, the transition log-likelihood takes the simplified form $\log \pi_\theta(\x_{t_{i+1}} \mid \x_{t_i}) \propto -\lVert \x_{t_{i+1}} - \mu_\theta \rVert^2$.

\noindent\textbf{Timestep-Balanced Gradient Reweighting.}
The strength of the policy gradient at step $k$ scales with the transition gain:
\begin{equation}
\kappa_k = 2 s_k \left| \frac{\partial \mu_\theta}{\partial \hat{v}_\theta} \right|, \qquad
\frac{\partial \mu_\theta}{\partial \hat{v}_\theta} = (1 - t_k)\, t_{k+1} \cos\!\left(\frac{\eta\pi}{2}\right) - t_k (1 - t_{k+1}),
\end{equation}
which varies strongly across the schedule, so a uniformly sampled critical step would let a few timesteps dominate training~\cite{he2026flashgrpo,he2025tempflow,longcatvideo}. We reweight each transition by the inverse gain, normalized to unit mean over the schedule:
\begin{equation}
\lambda_k = \frac{\kappa_k^{-1}}{\frac{1}{N} \sum_{j=0}^{N-1} \kappa_j^{-1}}.
\end{equation}
Here, $N$ is the number of inference steps, and the normalizer is computed analytically over the schedule grid rather than from batch statistics, since all samples in an update batch share one timestep. This equalizes update magnitudes across timesteps without changing the effective learning rate.

\noindent\textbf{Multi-Reward Advantage Normalization.}
The rewards of~\cref{sec:training:reward} have different scales and variances, so we normalize each reward independently before fusing them with weights~\cite{liu2026gdpo}:
\begin{equation}
\hat{A}^{(i)} = \sum_{r} w_r \frac{R_r\big(\x_0^{(i)}, c\big) - \mu_r}{\sigma_r + \delta},
\end{equation}
where $R_r(\x_0^{(i)}, c)$ is the $r$-th reward of sample $i$ under prompt $c$, $w_r$ is its weight, $\mu_r$ and $\sigma_r$ are the mean and standard deviation of $R_r$ within the group, and $\delta$ is a small constant.

\noindent\textbf{Objective.}
The final loss is the reweighted policy gradient on the sampled critical transitions:
\begin{equation}
\mathcal{L}_{\mathrm{GRPO}}(\theta) = - \E\left[ \lambda_k\, \hat{A}^{(i)} \log \pi_\theta\big(\x_{t_{k+1}}^{(i)} \mid \x_{t_k}^{(i)}\big) \right].
\end{equation}
Each rollout batch is consumed by a single gradient update, so training is strictly on-policy and importance ratios stay at one. Following~\cite{xue2025dancegrpo,he2026flashgrpo,xue2026videoposttrain}, we use no KL penalty and no reference model, and we fine-tune all model parameters.

\subsubsection{Negative-Aware Finetuning with Real-World Videos}
\label{sec:training:NFT}

Most existing RL post-training methods for diffusion models rely on reward models to provide optimization signals~\cite{liu2025flowgrpo,xue2025dancegrpo}, which introduces the risk of reward hacking in the video domain. 
To mitigate this, we leverage real-world videos as direct preference signals: a real video clip serves as the positive (chosen) sample, while the model-generated video under the same prompt serves as the negative (rejected) sample. 
This data configuration shares a similar motivation with RealDPO~\cite{guo2025realdpo}, which also pairs real videos with generated ones. 
For the optimization we adopt the forward-process optimization framework of DiffusionNFT~\cite{zheng2026diffusionnft}, bypassing the need to backpropagate through denoising trajectories. 

\noindent\textbf{Preference Pair Construction.}
At each training step, we sample a batch of real video clips from a curated dataset and encode them through the pretrained VAE into clean latents $\x_0^w$ (chosen). 
The active policy generates $N$ videos per prompt via the inference pipeline, and their VAE latents form the rejected pool $\{\x_{0,i}^l\}_{i=1}^N$.
Each rejected latent $\x_0^l$ is paired with the corresponding chosen latent $\x_0^w$ to form a preference pair.
Both latents are perturbed to a shared random timestep $t \sim \mathcal{U}(0, 1)$ with a shared noise vector $\boldsymbol{\varepsilon} \sim \mathcal{N}(\mathbf{0}, \mathbf{I})$:
\begin{equation}
\x_t^w = (1 - t) \x_0^w + t \boldsymbol{\varepsilon},
\qquad
\x_t^l = (1 - t) \x_0^l + t \boldsymbol{\varepsilon}.
\end{equation}
The clean latent is recovered from the noised state via $\hat{\x}_0 = \x_t - t \cdot \hat{v}$, where $\hat{v}$ denotes the predicted velocity.
Operating on single-step noised states avoids trajectory-level log-probability computation, keeping each update computationally lightweight.

\begin{figure}[!ht]
    \centering
    \includegraphics[width=0.95\linewidth]{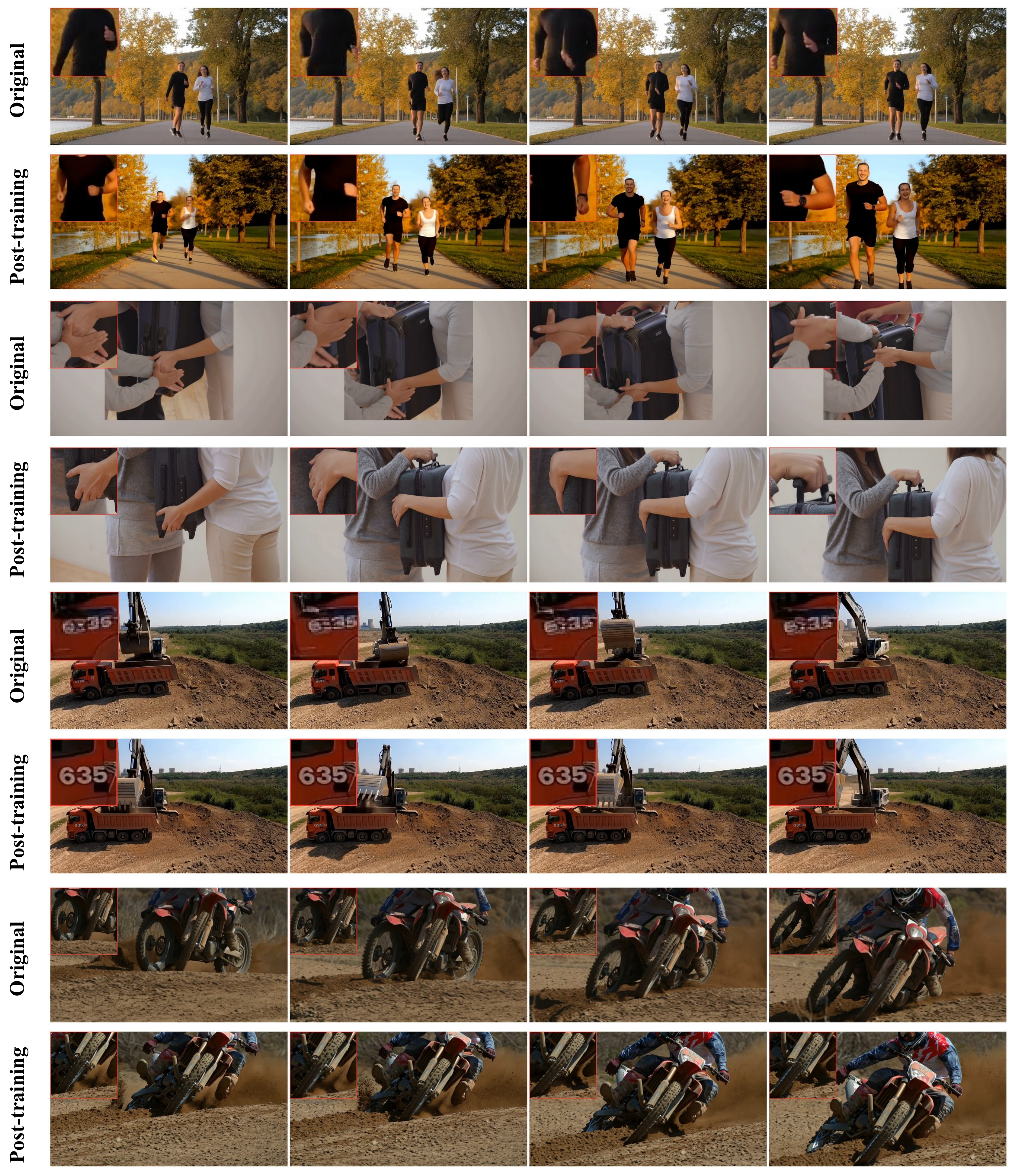}
    \caption{Qualitative comparison on general video quality before and after post-training, demonstrating marked improvements in several fundamental video generation domains. Post-training effectively resolves critical artifacts including inconsistent hand and limb synthesis, blurred or incorrect text rendering, and structural object deformation.}
    \label{fig:rl_general}
\end{figure}

\begin{figure}[!ht]
    \centering
    \includegraphics[width=0.95\linewidth]{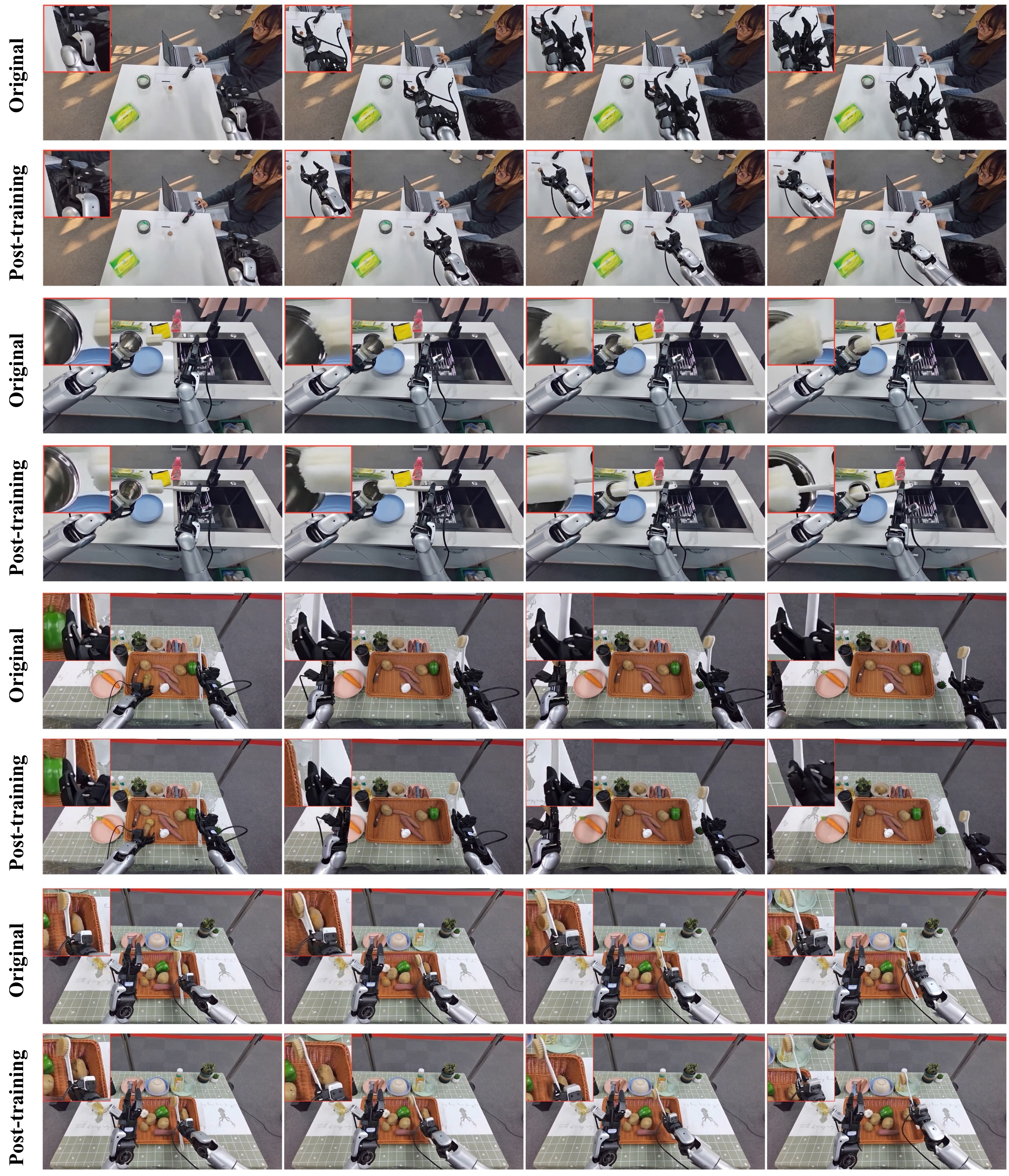}
    \caption{Qualitative comparison on embodied scenarios before and after post-training. The post-training phase significantly enhances physical plausibility by resolving baseline artifacts such as structural distortion of the arm and grasped objects, non-physical penetration, premature object release, and object duplication.}
    \label{fig:rl_embodied}
\end{figure}

\noindent\textbf{Negative-Aware Optimization.}
We maintain two prediction pathways sharing the same base model: (i) the \emph{active policy} $\theta$, which receives gradients; and (ii) the \emph{old policy}, an exponential moving average (EMA) copy of $\theta$ that stabilizes the optimization by preventing the active policy from changing too rapidly.
For any noised state $\x_t^s$ ($s \in \{w, l\}$), the two pathways produce velocity predictions $\hat{v}_{\theta}^s$ and $\hat{v}_{\text{old}}^s$.

Following DiffusionNFT~\cite{zheng2026diffusionnft}, we construct an implicit positive policy that blends the active and old velocity predictions to imitate the target, and an implicit negative policy that extrapolates past the old policy to suppress it:
\begin{align}
\hat{v}_{\text{pos}} = \beta \hat{v}_{\theta} + (1-\beta) \hat{v}_{\text{old}}, \qquad
\hat{v}_{\text{neg}} = (1+\beta) \hat{v}_{\text{old}} - \beta \hat{v}_{\theta},
\end{align}
where $\beta \in (0,1]$ controls how strongly the active policy is allowed to deviate from the old policy.
DiffusionNFT defines the per-sample loss as a reward-weighted mixture of positive and negative branches:
\begin{equation}
\Loss_{\text{NFT}}(r) = r \big\| \hat{v}_{\text{pos}} - v \big\|^2 + (1 - r) \big\| \hat{v}_{\text{neg}} - v \big\|^2,
\end{equation}
where $v = \boldsymbol{\varepsilon} - \x_0$ is the ground-truth flow matching velocity and $r \in [0,1]$ is a normalized reward.

\noindent\textbf{Pairwise Preference Adaptation.}
We adapt this formulation to our pairwise setting by treating the real video as the optimal sample ($r = 1$) and the generated video as the negative sample.
This yields:
\begin{equation}
\Loss_{\text{chosen}} = \Loss_{\text{NFT}}(1)^w = \big\| \hat{v}_{\text{pos}}^w - v^w \big\|^2,
\qquad
\Loss_{\text{reject}} = \Loss_{\text{NFT}}(r)^l = r \big\| \hat{v}_{\text{pos}}^l - v^l \big\|^2 + (1 - r) \big\| \hat{v}_{\text{neg}}^l - v^l \big\|^2.
\end{equation}
In principle, $r$ for the rejected sample could be assigned by a reward model according to generation quality. 
For simplicity and to avoid introducing reward model overhead, we set $r = 0$. 
We also regularize the active policy against a frozen copy of the base model to prevent the policy from drifting too far during fine-tuning:
\begin{equation}
\Loss_{\text{KL}} = \tfrac{1}{2} \bigl( \|\hat{v}_{\theta}^w - \hat{v}_{\text{ref}}^w\|^2 + \|\hat{v}_{\theta}^l - \hat{v}_{\text{ref}}^l\|^2 \bigr).
\end{equation}
The total training objective is:
\begin{equation}
\Loss_{\text{RealNFT}} = \Loss_{\text{chosen}} + \Loss_{\text{reject}} + \lambda_{\text{KL}} \Loss_{\text{KL}}.
\end{equation}

\subsection{Action-to-Video Post-Training}
\label{sec:training:acwm}

\begin{figure}[htbp]
    \centering
    \includegraphics[width=0.7\linewidth]{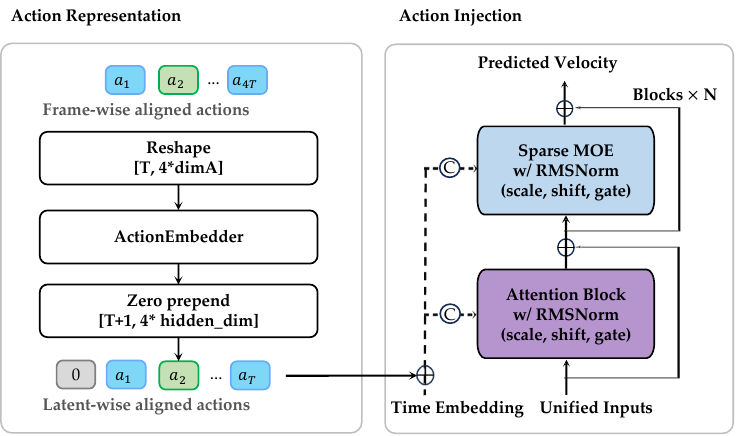}
    \caption{Architecture of \acwm. Given frame-wise actions for the future $4T$ frames, \acwm first converts the raw commands into relative actions, flattens the action sequence, and maps it with a learnable \texttt{ActionEmbedder} to action latents. A zero action is prepended for the initial state to temporally align action sequence with the $T+1$ visual latents. The action embeddings are injected into the pre-trained transformer blocks with time embeddings.}
    \label{fig:acwm_architecture}
\end{figure}

Embodied world simulation requires a model to roll out plausible future visual trajectories from the current state and a planned action sequence \cite{veorobotics2025,zhu2025unified}.
Beyond visual generation, such simulated futures are central to embodied AI and robot planning, supporting policy evaluation \cite{veorobotics2025,tseng2026sc3}, test-time planning through imagined rollouts \cite{wan2026worldagen}, robot-data scaling via synthetic trajectories \cite{guo2025ctrl}, and reinforcement-learning environments for interactive policy improvement \cite{guo2026vlaw,jiang2026wovr}.
To examine whether \method can transfer its pre-trained understanding of physical rationality, spatial relations, and temporal evolution to this downstream setting, we further post-train it as an action-conditioned world model, denoted \acwm.
Conditioned only on the initial world state, represented by the first image and a textual state description, and on the robot action parameters, \acwm generates action-conditioned visual rollouts of future world states. This setting poses a challenging test of physical-law modeling and action following. We organize the post-training design around three components:
\begin{itemize}
    \item \textit{Data recaptioning.} Pre-trained {\methodname} excels at prompt following, where detailed prompts describe the evolution of the whole video, including physical dynamics and action descriptions. To adapt this simulator into a predictor, we rewrite the data captions so that each prompt describes only the initial state. This encourages the model to drive video evolution using only the given robot actions. We further apply a strict future-leakage check to ensure that the prompts do not reveal future observations or dynamics.
    \item \textit{Action representation and injection.} Given an initial frame and future robot action chunks, we first convert the raw actions into relative actions, so that each step encodes the incremental change from the preceding state. \acwm then flattens the full action chunk into a single sequence and projects it through the \texttt{ActionEmbedder}. As shown in \cref{fig:acwm_architecture}, we prepend a zero action for the initial observation and keep the action latents temporally aligned with the visual latents. The encoded latent-wise aligned actions are injected as residual signals to modulate each transformer block, together with the time embedding. To stabilize post-training, we zero-initialize the last layer of the \texttt{ActionEmbedder}, allowing the newly introduced action branch to be integrated gradually into the pre-trained backbone.
    \item \textit{Training setup.} We adopt the Fourier GR-1 post-training datasets for experiments~\cite{gao2026dreamdojo}. Each training sample follows the same formulation: the model observes the initial world state, receives a sequence of robot actions, and is supervised to generate the corresponding future visual rollout. Starting from the unified pre-trained {\methodname}, we optimize the \texttt{ActionEmbedder} together with the full transformer backbone, rather than training the action branch in isolation, to adapt the original video simulator to predict future frames conditioned on actions. We run post-training for $8k$ steps with a global batch size of 64 and a learning rate of $1e^{-5}$.
\end{itemize}

This adaptation benefits from the synergy between strong pre-trained world priors and targeted action-rollout data. {\methodname} provides action-aware priors over physical-world causality, spatial relations, and temporal evolution, while our recaptioned, leakage-filtered GR-1 trajectories provide clean supervision for action-conditioned rollouts. As a result, \acwm achieves high-quality action following after post-training and can be readily adapted to embodied world simulation. We include additional experimental results in~\cref{sec:eval:acwm}.

\subsection{Distillation}
To improve inference efficiency, we distill \method into a few-step generator following the improved Distribution Matching Distillation (DMD2) framework~\cite{yin2024improved}. Let $G_{\theta}$ denote the student generator conditioned on the unified condition $c$, and let $x_0 = G_{\theta}(z,c)$, where $z \sim \mathcal{N}(0,I)$. For a sampled timestep $t$, we perturb the generated latent video as $x_t = \alpha_t x_0 + \sigma_t \epsilon$. The method matches the student distribution to the teacher distribution by minimizing a reverse-KL-style distribution-matching objective over diffusion noise levels:
\begin{equation}
\nabla_{\theta}\mathcal{L}_{\mathrm{DMD}}
=
\mathbb{E}_{z,t,\epsilon}
\left[
-w_t
\left(
s_{\mathrm{real}}(x_t,t,c)-s_{\mathrm{fake}}(x_t,t,c)
\right)
\frac{\partial G_{\theta}(z,c)}{\partial \theta}
\right],
\end{equation}
where $s_{\mathrm{real}}$ is provided by the teacher video diffusion model, and $s_{\mathrm{fake}}$ is estimated by an auxiliary score model trained online on samples from the current student. We also retain a lightweight GAN objective to provide real-data supervision and improve visual quality.
% !TEX root = ../main.tex
\section{Evaluation} \label{sec:eval}

\subsection{Internal Benchmark}

\begin{figure}[t]
    \centering
    \begin{subfigure}{0.49\linewidth}
        \centering
        \includegraphics[width=\linewidth]{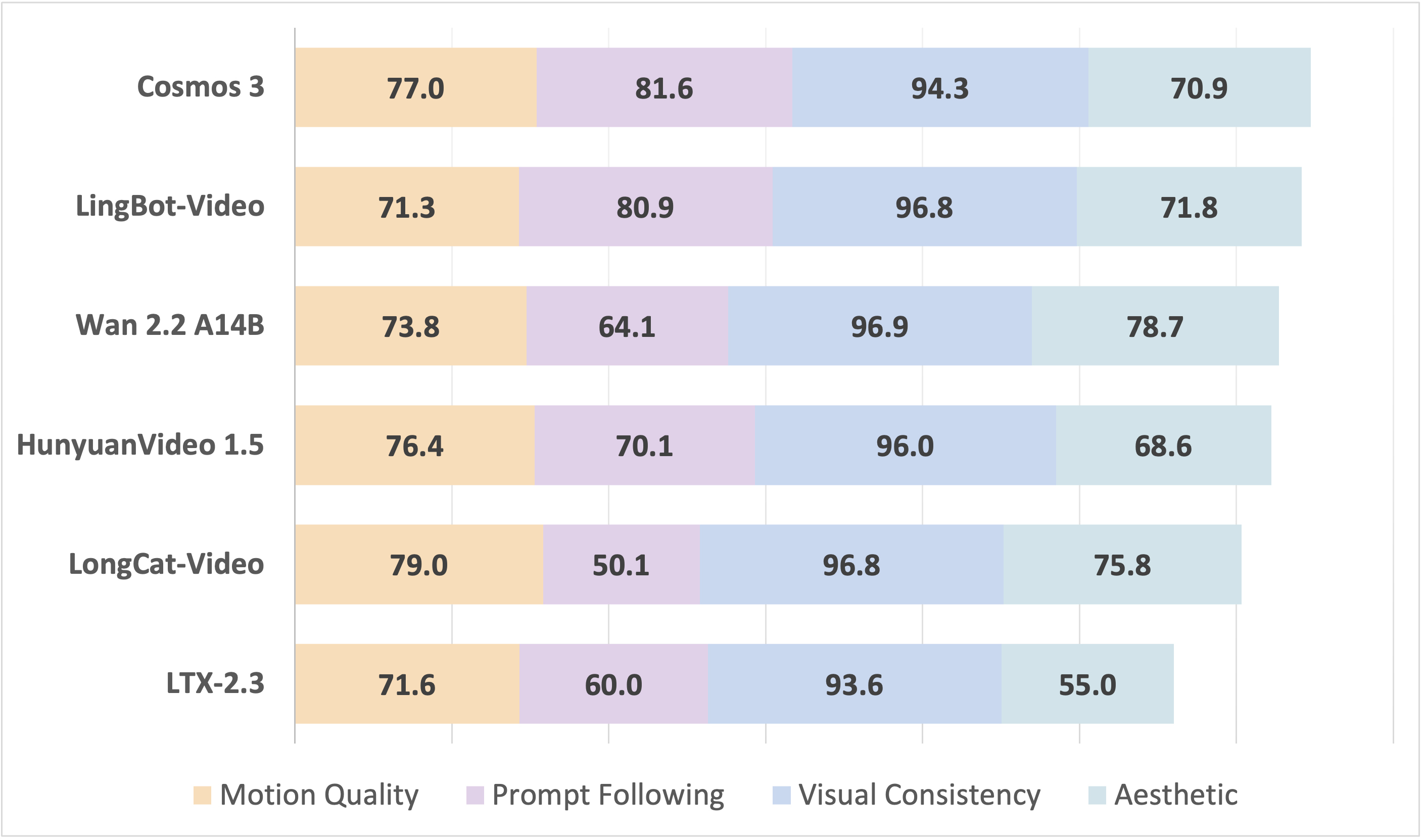}
        \caption{T2V quality score.}
        \label{fig:bench_internal_t2v_quality}
    \end{subfigure}
    \hfill
    \begin{subfigure}{0.49\linewidth}
        \centering
        \includegraphics[width=\linewidth]{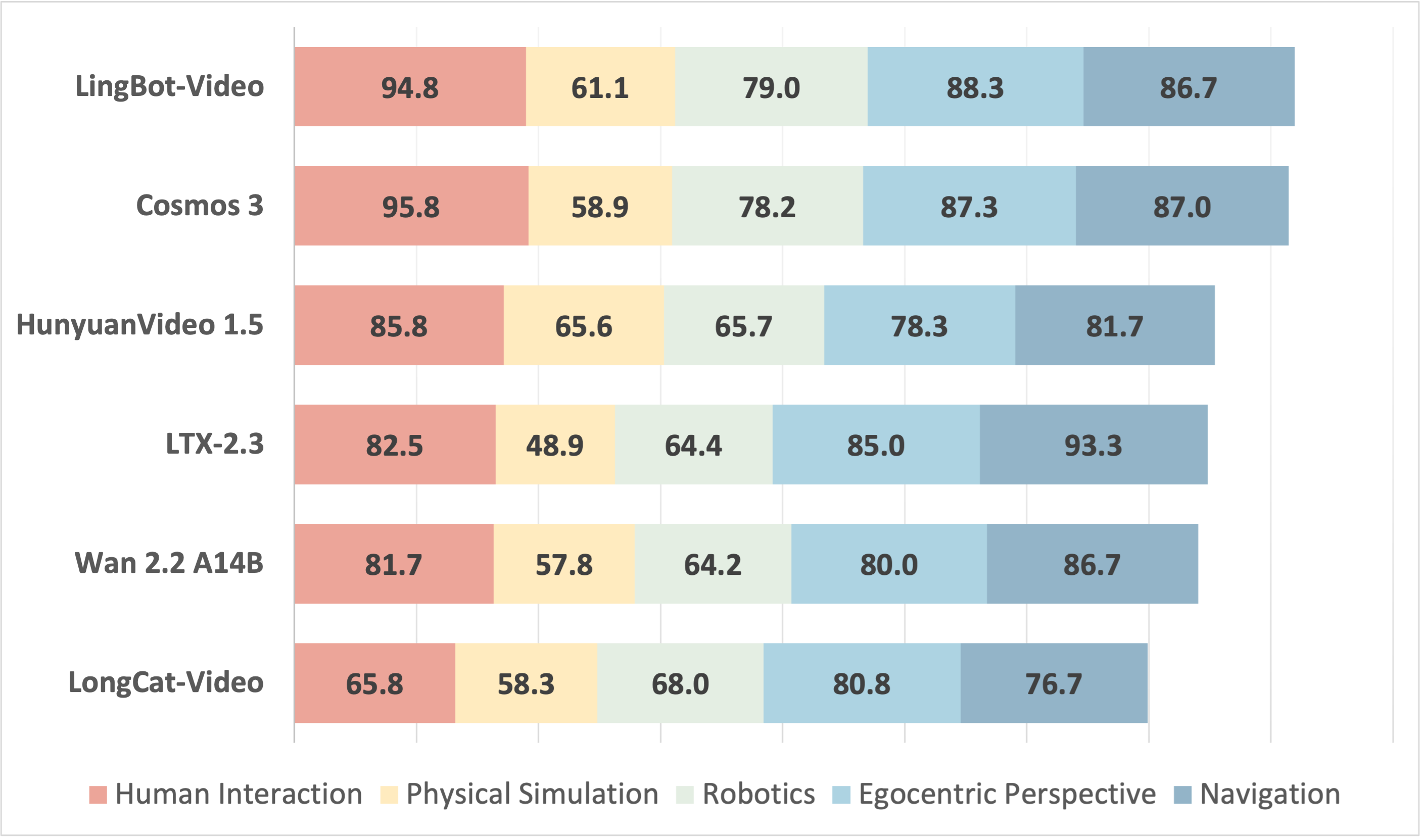}
        \caption{T2V domain score.}
        \label{fig:bench_internal_t2v_domain}
    \end{subfigure}

    \vspace{1em}
    
    \begin{subfigure}{0.49\linewidth}
        \centering
        \includegraphics[width=\linewidth]{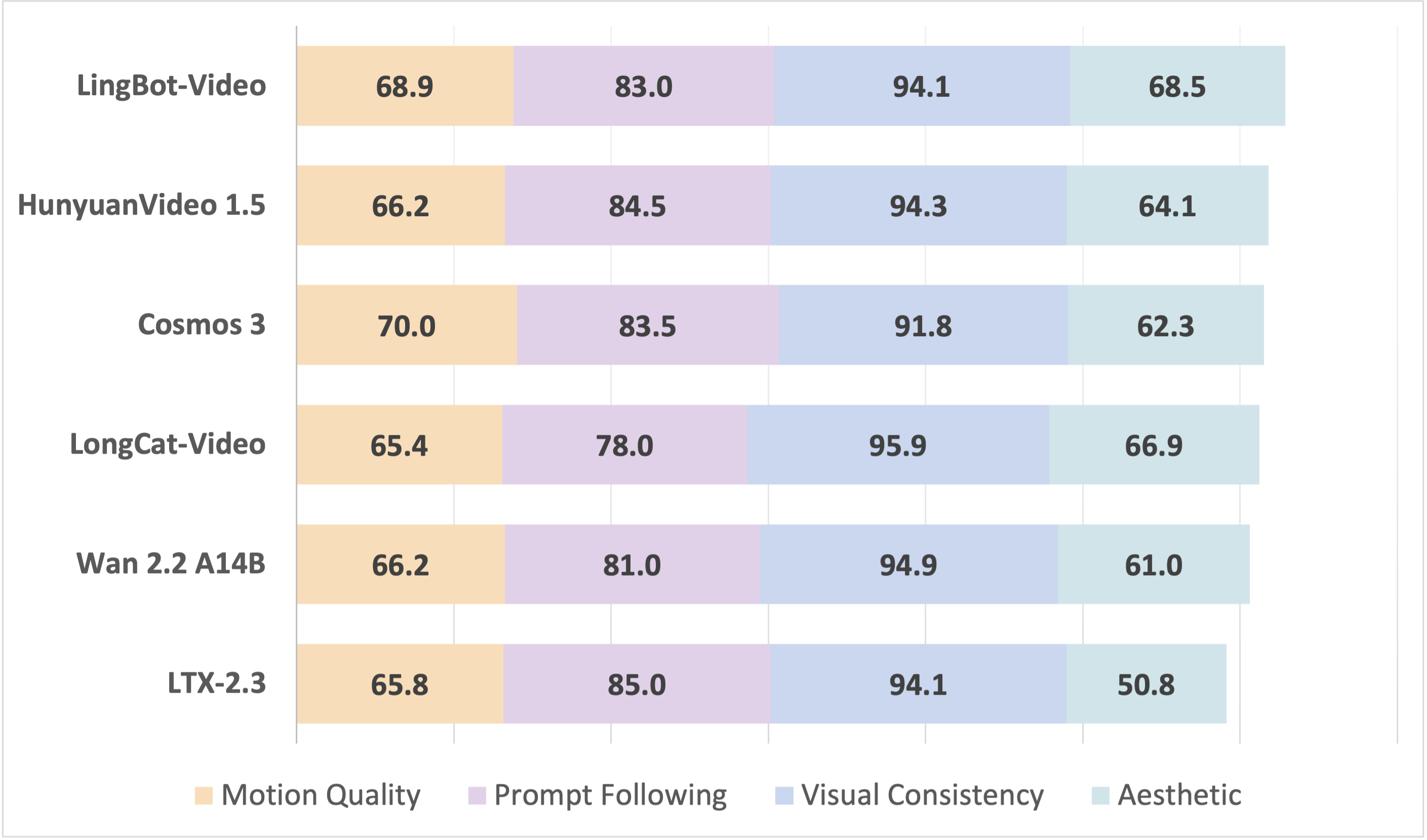}
        \caption{TI2V quality score.}
        \label{fig:bench_internal_ti2v_quality}
    \end{subfigure}
    \hfill
    \begin{subfigure}{0.49\linewidth}
        \centering
        \includegraphics[width=\linewidth]{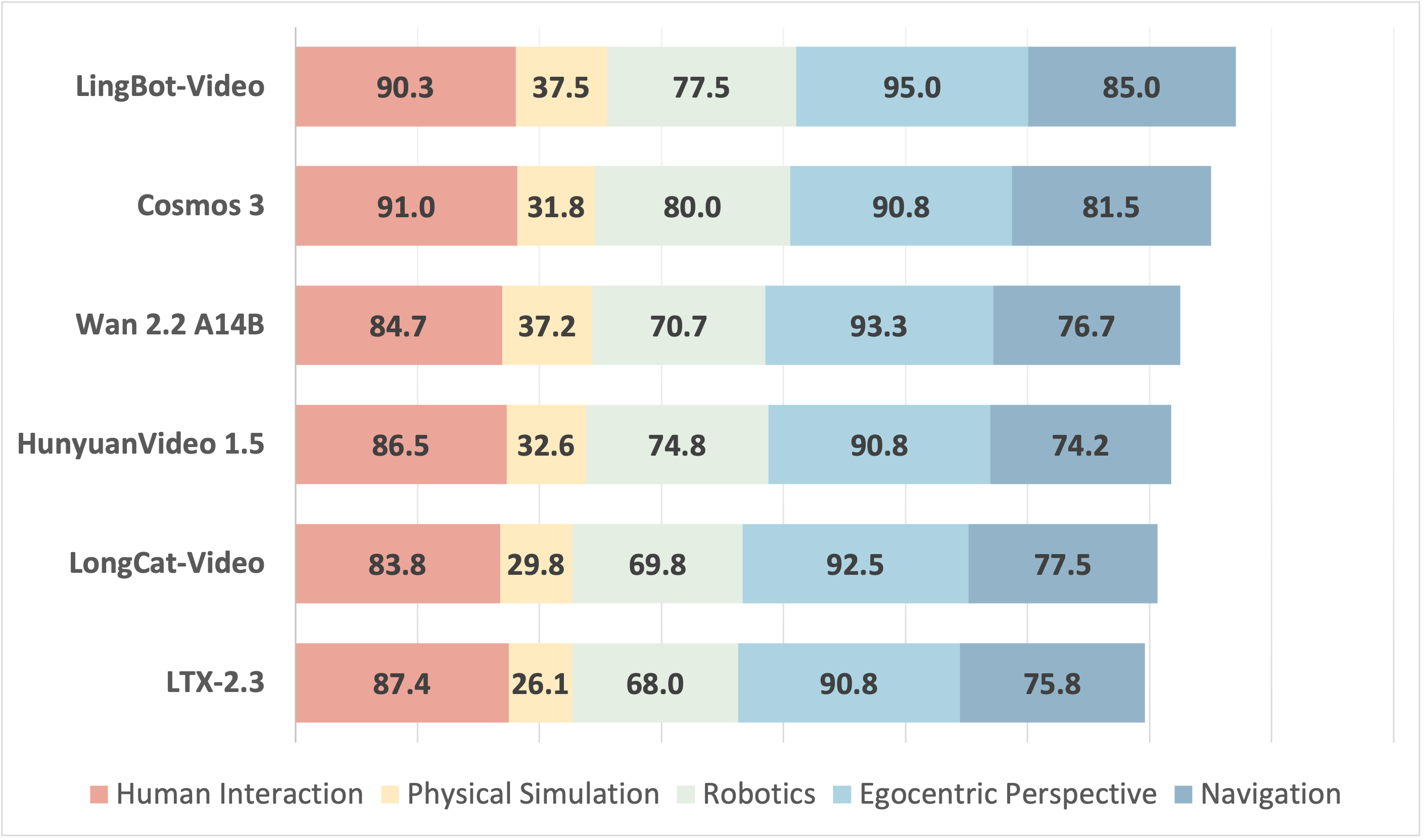}
        \caption{TI2V domain score.}
        \label{fig:bench_internal_ti2v_domain}
    \end{subfigure}
    
    \caption{\textbf{Quantitative evaluation on our internal benchmark.} We evaluate the performance of \method and other state-of-the-art open-source competitors across two dimensions: \textit{general quality} (for overall visual appeal and coherence) and \textit{embodied domain} (for specific category distributions). Top row shows results under the Text-to-Video (T2V) setting, while the bottom row illustrates results under the Text-and-Image-to-Video (TI2V) setting.}
    \label{fig:bench_internal}
\end{figure}

To verify the capability of \method as a physical world model, we conduct a comprehensive evaluation on our internal benchmark across two distinct dimensions: \textit{General Quality}, which assesses fundamental generative capabilities, and \textit{Embodied Domain}, which probes specialized, high-difficulty scenarios relevant to embodied AI and real-world interactions. 
To comprehensively evaluate the generation capabilities, the internal benchmark cover two core generation settings: \textit{Text-to-Video (T2V)} and \textit{Text-and-Image-to-Video (TI2V)}.

\noindent\textbf{General Quality.}
The general domain focuses on the foundational video generation quality, ensuring that the generated videos are visually pleasing, temporally consistent, and semantically accurate.
\begin{itemize}
  \item \textbf{Motion Quality} measures the naturalness, continuity, and physical plausibility of movements.
  It evaluates whether the video exhibits smooth motion trajectories and remains free from severe temporal artifacts such as flickering, structural deformation, or unnatural human actions (e.g., identity drift or clothing deformation).

  \item \textbf{Prompt Following} assesses how faithfully the model adheres to the input text instructions.
  It measures semantic alignment across multiple entities, complex counting scenarios, sequential action order, specified camera movements, \textit{etc}.

  \item \textbf{Visual Consistency} quantifies the model's ability to maintain identity and scene context over time.
  This includes preserving the consistency of background layouts, main subject details, and specific instances across frames.
  For the TI2V setting, it additionally incorporates first-frame image preservation to ensure the generated video faithfully inherits the appearance and style of the input reference image.

  \item \textbf{Aesthetic Quality} evaluates the general visual appeal, artistic value, and stylistic execution of the generated video, emphasizing overall cinematic composition, lighting, and texture.
\end{itemize}

\noindent\textbf{Embodied Domain.}
Beyond general quality, \method is targeted at complex, interactive, and embodied scenarios. 
For a real-world robot equipped with cameras, the core objective is not to generate a scene from scratch, but to predict future physical interactions conditioned on the current observation (the initial frame $I_0$) and a control command ($T$). 
The embodied domain evaluation specifically probes the model's performance on highly specialized tasks that demand physical understanding, spatial reasoning, and interaction. 
It covers the following categories:
\begin{itemize}
  \item \textbf{Human Interaction} evaluates the model's capacity to synthesize intricate physical interactions and expressive behaviors.
  This covers fine-grained categories such as human interaction, object manipulation, animal interaction and sports dynamics. 
  It also extends to human-related evaluation, such as detailed hand and finger motions, and subtle facial emotions and expressions.

  \item \textbf{Physical Simulation} focuses on how well the model adheres to intuitive physical laws, acting as a ``world simulator'' for embodied generation.
  It covers scenarios governed by classical mechanics, optics (e.g., reflections, shadows), thermodynamics, fluid dynamics, material properties, and magnetism.

  \item \textbf{Robotics} targets embodied agent scenarios, testing the model's capability to generate coherent movements for diverse robotic platforms, including humanoid robots, robotic arms, and quadrupedal robots.

  \item \textbf{Egocentric Perspective} assesses the model's proficiency in rendering first-person perspective videos, which are crucial for ego-agent.

  \item \textbf{Navigation} tests the model's understanding of spatial layouts and motion planning across diverse environments, including outdoor streets, autonomous driving scenes, interactive game environments, and complex indoor layouts.
\end{itemize}

\begin{figure}[t]
    \centering
    \begin{subfigure}{0.49\linewidth}
        \centering
        \includegraphics[width=\linewidth]{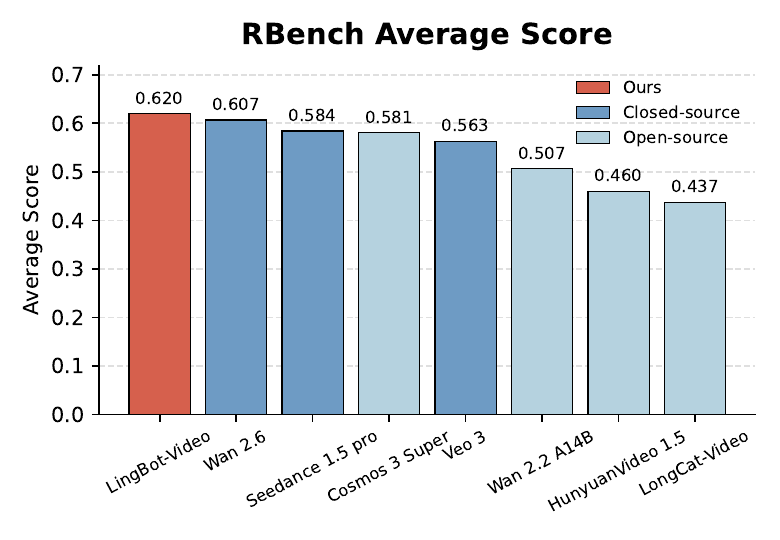}
        \caption{RBench score.}
        \label{fig:rbench_bar}
    \end{subfigure}
    \hfill
    \begin{subfigure}{0.49\linewidth}
        \centering
        \includegraphics[width=\linewidth]{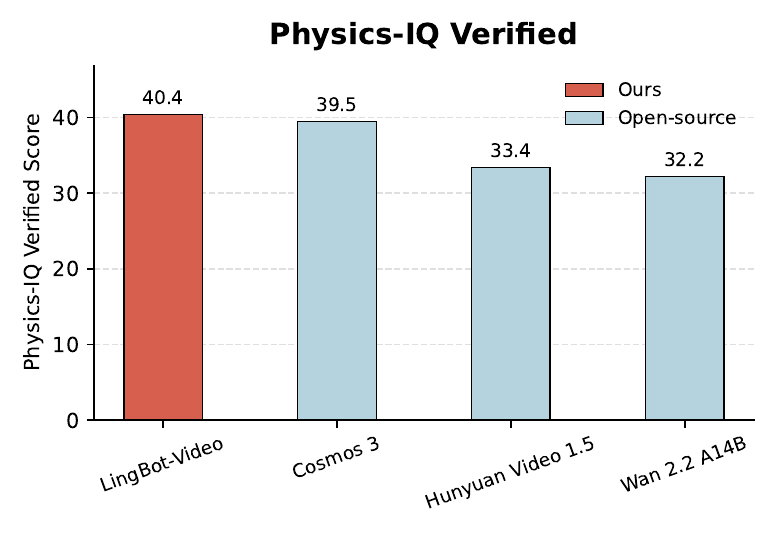}
        \caption{Physics-IQ Verified I2V score.}
        \label{fig:physics_iq_bar}
    \end{subfigure}
    \caption{\textbf{Public benchmark score comparison.} We visualize the average scores from \cref{tab:rbench} and the Physics-IQ verified scores, with \method highlighted against open-source and closed-source baselines.}
    \label{fig:public_benchmark_bars}
\end{figure}

\noindent\textbf{Results.}
We compare \method with five open-source models, including NVIDIA Cosmos~3 Super-Image-to-Video, Wan~2.2~A14B, LongCat-Video, Hunyuan Video~1.5, and LTX-2.3. 
As shown in~\cref{fig:bench_internal_ti2v_quality} and~\cref{fig:bench_internal_ti2v_domain}, \method achieves state-of-the-art performance among all open-source competitors on the TI2V task, securing the top spot in both general quality and embodied domain scores. 
This demonstrates our model's superior capability in simulating precise physical trajectories, such as robotic arm manipulation and obstacle avoidance. 
For the T2V task (~\cref{fig:bench_internal_t2v_quality} and~\cref{fig:bench_internal_t2v_domain}), while our model ranks second in general quality, we still consistently outperform competitive baselines such as Cosmos on the embodied domain score. 
This edge, even in the absence of initial image conditioning, highlights that \method possesses robust and intrinsic physical priors crucial for embodied AI applications.

\begin{table}[t]
    \centering
    \footnotesize
    \setlength{\tabcolsep}{3pt}
    \begin{tabular}{l l c c c c c c c c c c}
        \toprule
        \multirow{2}{*}{Models} & \multirow{2}{*}{Type} & \multirow{2}{*}{Avg.} & \multicolumn{5}{c}{Tasks} & \multicolumn{4}{c}{Embodiments} \\
        \cmidrule(lr){4-8} \cmidrule(lr){9-12}
        & & & Manip. & Spatial & Multi-entity & Long-hor. & Reasoning & Single arm & Dual arm & Quadruped & Humanoid \\
        \midrule
        \rowcolor{gray!15} % 设置整行底色为 15% 的浅灰色
        \method & open-source & \textbf{0.620} & \textbf{0.578} & \underline{0.643} & 0.444 & \textbf{0.634} & \underline{0.505} & 0.636 & 0.639 & \textbf{0.758} & 0.689 \\
        Cosmos3 Super & open-source & 0.581 & 0.487 & 0.642 & 0.444 & \underline{0.591} & 0.395 & 0.615 & 0.623 & \underline{0.739} & \underline{0.691} \\
        LongCat-Video & open-source & 0.437 & 0.372 & 0.310 & 0.220 & 0.384 & 0.186 & 0.586 & 0.576 & 0.681 & 0.621 \\
        Wan~2.2~A14B & open-source & 0.507 & 0.381 & 0.454 & 0.373 & 0.501 & 0.330 & 0.608 & 0.582 & 0.690 & 0.648 \\
        HunyuanVideo~1.5 & open-source & 0.460 & 0.442 & 0.316 & 0.312 & 0.438 & 0.364 & 0.513 & 0.526 & 0.634 & 0.595 \\
        \midrule
        Wan~2.6 & closed-source & \underline{0.607} & 0.546 & \textbf{0.656} & \underline{0.479} & 0.514 & \textbf{0.531} & \textbf{0.666} & \textbf{0.681} & 0.723 & 0.667 \\
        Seedance~1.5 pro & closed-source & 0.584 & \underline{0.577} & 0.495 & \textbf{0.484} & 0.570 & 0.470 & \underline{0.648} & \underline{0.641} & 0.680 & \textbf{0.692} \\
        Veo~3 & closed-source & 0.563 & 0.521 & 0.508 & 0.430 & 0.530 & 0.504 & 0.634 & 0.610 & 0.689 & 0.637 \\
        % \method & \multicolumn{11}{c}{\textit{Results to be completed (Ongoing)}} \\
        \bottomrule
    \end{tabular}
    \caption{\textbf{RBench evaluation results.} We report the average score and sub-dimension scores across five task-oriented and four embodiment-specific categories for open-source and closed-source models.
    Scores of some models are sourced from RBench~\cite{deng2026rethinking}.}
    \label{tab:rbench}
\end{table}

\subsection{Public Benchmark}

To complement our internal evaluations, we further evaluate \method on public automated evaluation benchmarks on RBench~\cite{deng2026rethinking} and Physics-IQ Verified~\cite{radsch2026physics} designed for video generation in embodied and physical domains. 

\noindent\textbf{RBench.} RBench~\cite{deng2026rethinking} specifically targets the correctness of robot-centric interactions, making it a focused complement for the robotics category in our evaluation.
The benchmark includes 650 text-image prompts partitioned into two primary tracks: 250 task-oriented scenarios covering five interaction types (Manipulation, Spatial Relationship, Multi-entity Collaboration, Long-horizon Planning, and Visual Reasoning) and 400 embodiment-specific scenarios spanning four robot morphologies (Single-arm, Dual-arm, Humanoid, and Quadruped).
The compelling performance on RBench further validates \method's embodied generation capability, particularly in robotics-centric scenarios that demand physical coherence and precise instruction following.
This aligns with the robotics-domain advantage observed in our internal benchmark (\cref{fig:bench_internal_ti2v_domain}) and demonstrates that \method's physical world modeling generalizes beyond our internal evaluation suite.

% \noindent\textbf{Physics-IQ Verified} 
\noindent\textbf{Physics-IQ Verified.}
Physics-IQ Verified~\cite{radsch2026physics} is a refined version of the Physics-IQ~\cite{Motamed_2026_WACV} benchmark for evaluating whether video generation models can predict real-world physical phenomena rather than merely producing visually plausible motion.
The benchmark is built from 66 controlled physical experiments spanning solid dynamics, fluid dynamics, thermodynamics, optics, and magnetism.
Each experiment is recorded from three viewpoints and two takes, yielding 396 real-world videos in total.
Two evaluation modes are supported.
In image-to-video (I2V), the model is conditioned on a single switch frame plus an optional text prompt and predicts the subsequent motion.
In video-to-video (V2V) continuation, the model is conditioned on a 3-second conditioning video plus an optional text prompt and predicts the next 5 seconds of motion.
Generated videos are compared against ground-truth physical continuations along four complementary axes: spatial overlap, temporal alignment, magnitude-weighted spatial agreement, and pixel-level error.
We evaluate \method under the I2V setting, which matches our image-conditioned generation interface.
As shown in \cref{fig:public_benchmark_bars}(b), \method obtains a Physics-IQ Verified score of 40.4, ranking first among the evaluated open-source models and narrowly surpassing Cosmos~3 (39.5).
The margin over Hunyuan Video~1.5 (33.4) and Wan~2.2~A14B (32.2) is more pronounced, indicating stronger predictive consistency on real physical processes.
Together with the RBench results, this suggests that \method's world-modeling capability extends from robot-centric interaction scenarios to broader physical dynamics.

\subsection{User Study}

\begin{figure}[htbp]
    \centering
    \begin{subfigure}{0.49\linewidth}
        \centering
        \includegraphics[width=\linewidth]{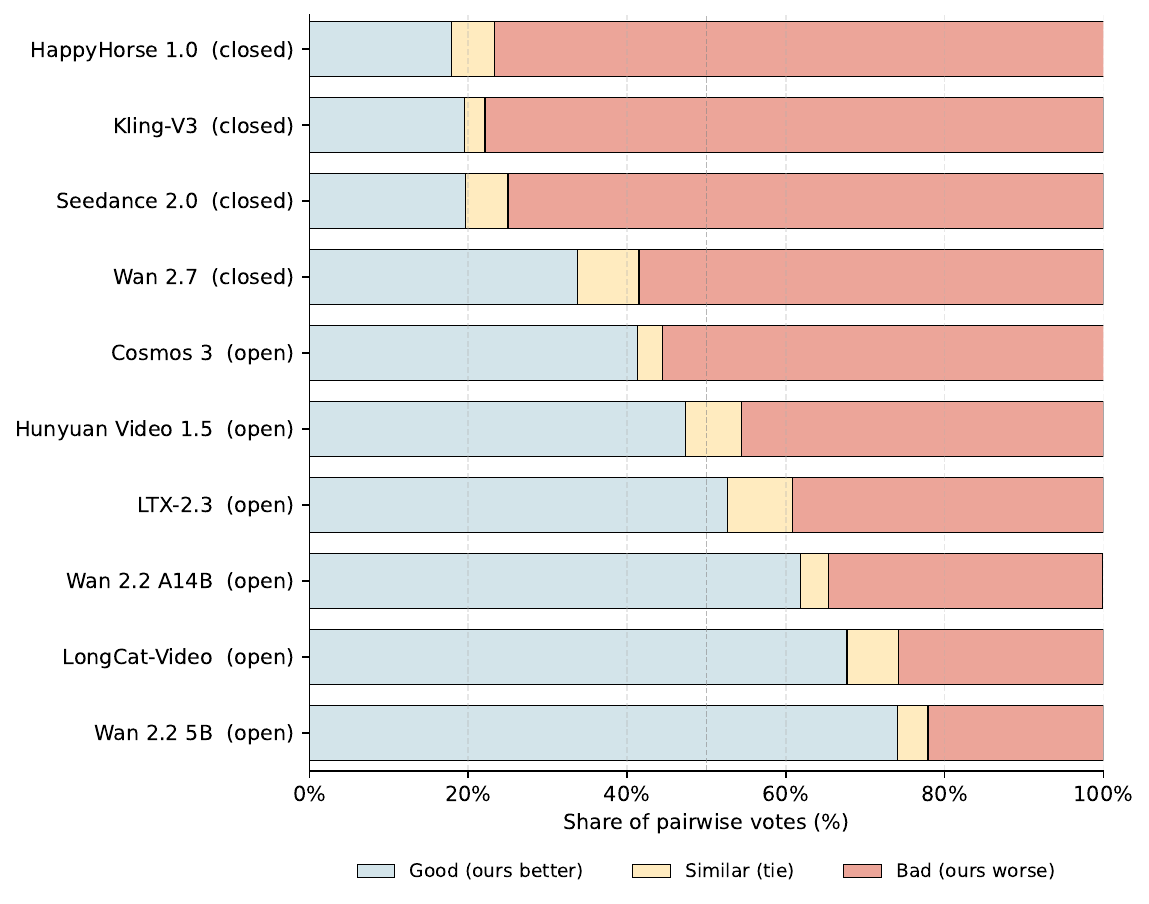}
        \caption{T2V.}
        \label{fig:user_study_t2v}
    \end{subfigure}
    \hfill
    \begin{subfigure}{0.49\linewidth}
        \centering
        \includegraphics[width=\linewidth]{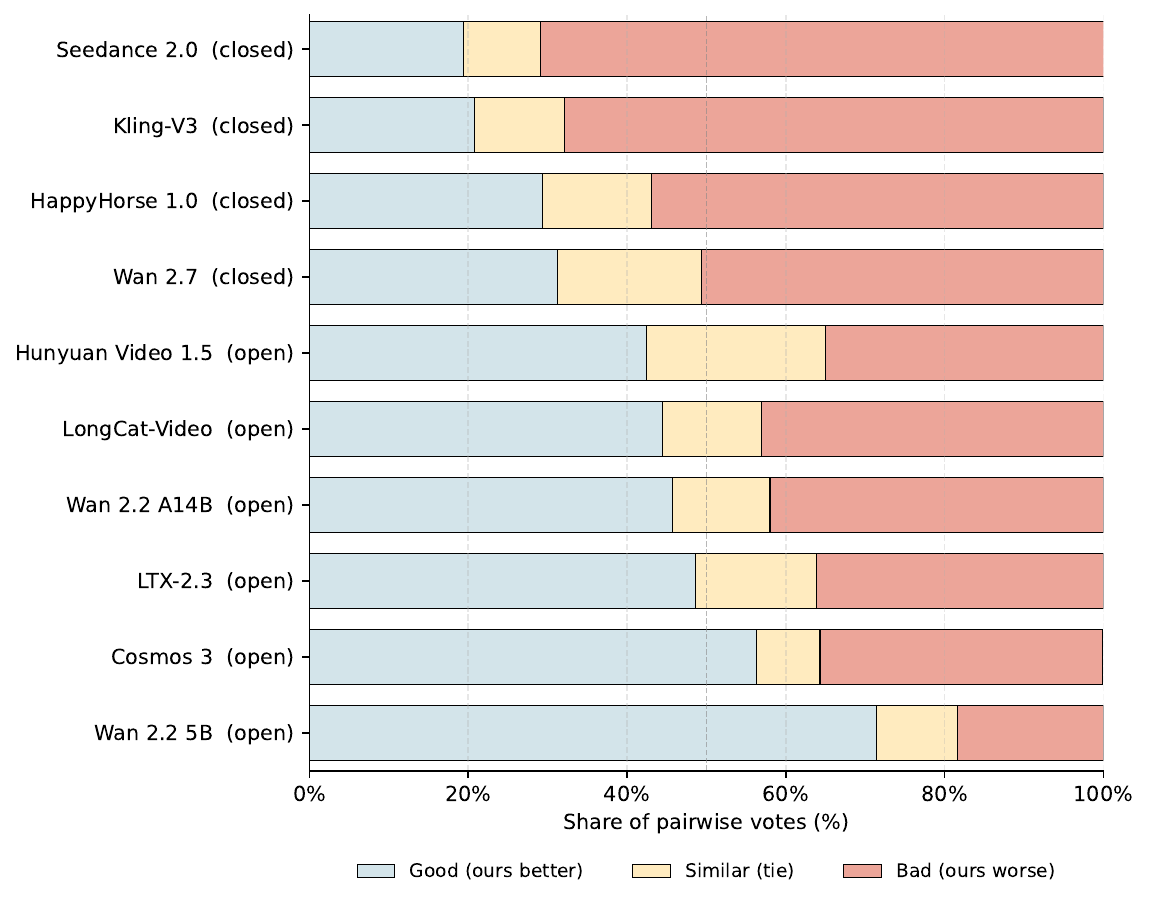}
        \caption{TI2V.}
        \label{fig:user_study_ti2v}
    \end{subfigure}
    \caption{\textbf{User study results.} Good-Same-Bad human evaluation results for T2V and TI2V generation.}
    \label{fig:user_study}
\end{figure}

We conduct a Good-Same-Bad (GSB) human evaluation to benchmark \method against leading open-source and commercial video generation models.
The evaluation adopts the same domain as our internal benchmark.
For each prompt, human raters are presented with a pair of videos in a randomized and anonymized side-by-side format, and are asked to judge whether the first video is \textit{Good} (better), \textit{Same} (comparable), or \textit{Bad} (worse) than the second. 
This blind pairwise setup forces a direct quality comparison, mitigating subjective scoring bias.
In this study, we compare \method against six open-source models (NVIDIA Cosmos~3, Wan~2.2~5B, Wan~2.2~A14B, LongCat-Video, HunyuanVideo~1.5, and LTX-2.3) and four commercial models (Kling-V3, Wan~2.7, Seedance~2.0, and HappyHorse~1.0).
In total, we collect human ratings across 400 prompts for each comparison pair.
The full GSB breakdown for text-to-video (T2V) and text-and-image-to-video (TI2V) generation is reported in \cref{fig:user_study}.
For T2V generation, \method clearly outperforms several open-source baselines, with the \textit{Good} rate exceeding the \textit{Bad} rate against Wan~2.2~5B, LongCat-Video, Wan~2.2~A14B, and LTX-2.3.
The comparison is closer against stronger open-source models such as HunyuanVideo~1.5 and Cosmos~3, indicating that \method remains a competitive setting.
For TI2V generation, the advantage becomes more consistent: \method obtains higher \textit{Good} than \textit{Bad} rates against all evaluated open-source baselines, with especially large margins over Wan~2.2~5B and clear gains over Cosmos~3, LTX-2.3, Wan~2.2~A14B, LongCat-Video, and HunyuanVideo~1.5.
This stronger TI2V result suggests \method as a strong open-source model while  trailing the stronger commercial models.

\subsection{Action-to-Video Post-Training}
\label{sec:eval:acwm}

\begin{figure}[t]
    \centering
    \includegraphics[width=1\linewidth]{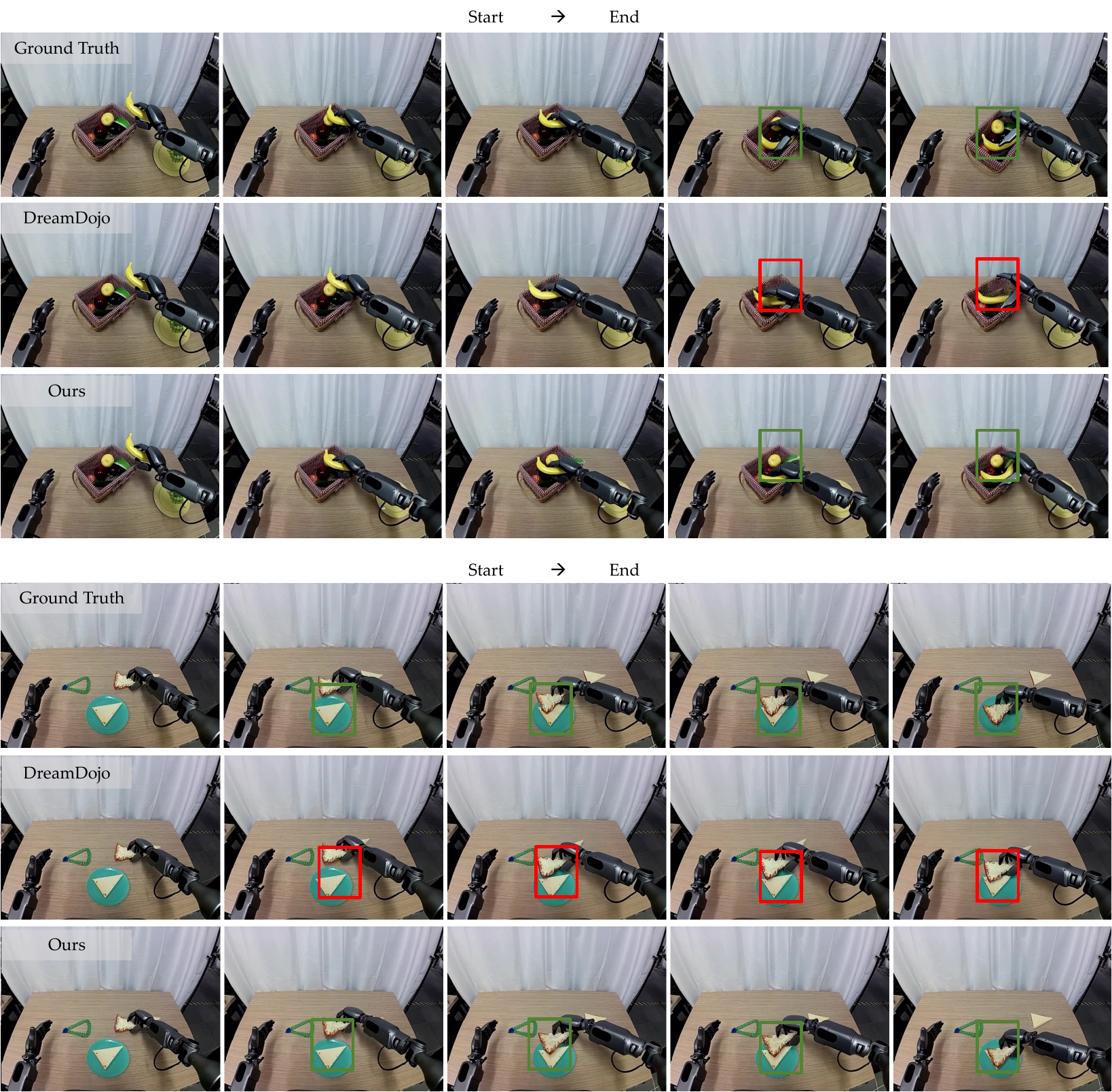}
    \caption{Compared with DreamDojo~\cite{gao2026dreamdojo}, our model demonstrates better adherence to physical laws, such as preserving the yellow apple in the first example, and stronger action following, as shown by the hand pose relative to the sandwich.} 
    \label{fig:res-acwm}
\end{figure}
We include the results of \acwm on EgoDex Eval and DreamDojo-HV Eval in \cref{fig:res-acwm} to examine whether the post-trained model generalizes beyond the GR-1 trajectories used for training. Both evaluation datasets contain novel objects and actions that are absent from the GR-1 post-training dataset~\cite{gao2026dreamdojo}, making them suitable for testing out-of-distribution action following rather than memorization of training rollouts. The results shown in \cref{fig:res-acwm} demonstrate that our model adheres better to physical laws and follows actions more accurately.

\clearpage
\begin{figure}[p]
  \centering
  \includegraphics[width=\linewidth]{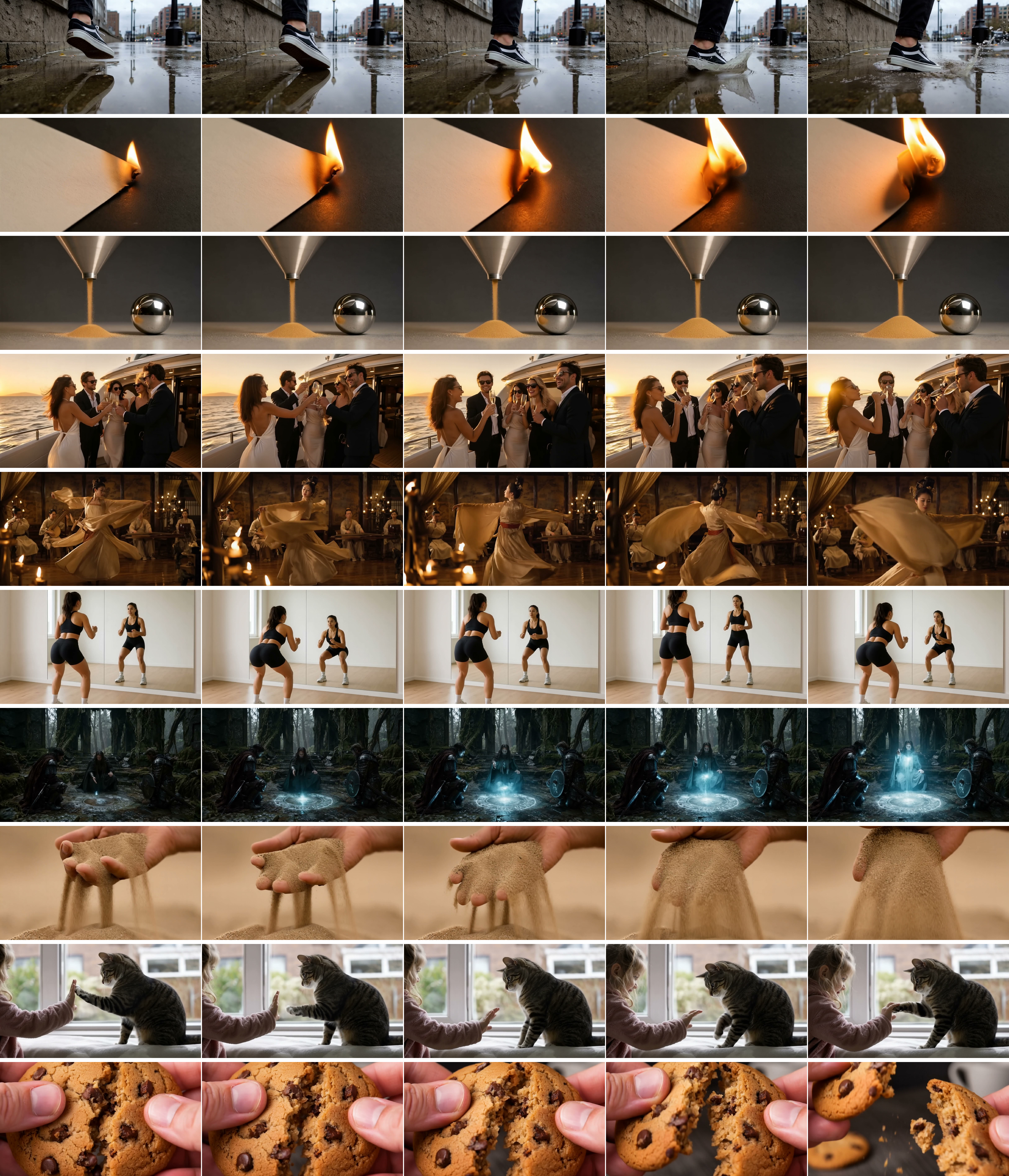}
  \caption{Qualitative results of \method on text-and-image-to-video generation. Each row shows five keyframes uniformly sampled from one generated video; time flows from left to right.}
  \label{fig:cases_ti2v}
\end{figure}
\clearpage
\begin{figure}[p]
  \centering
  \includegraphics[width=\linewidth]{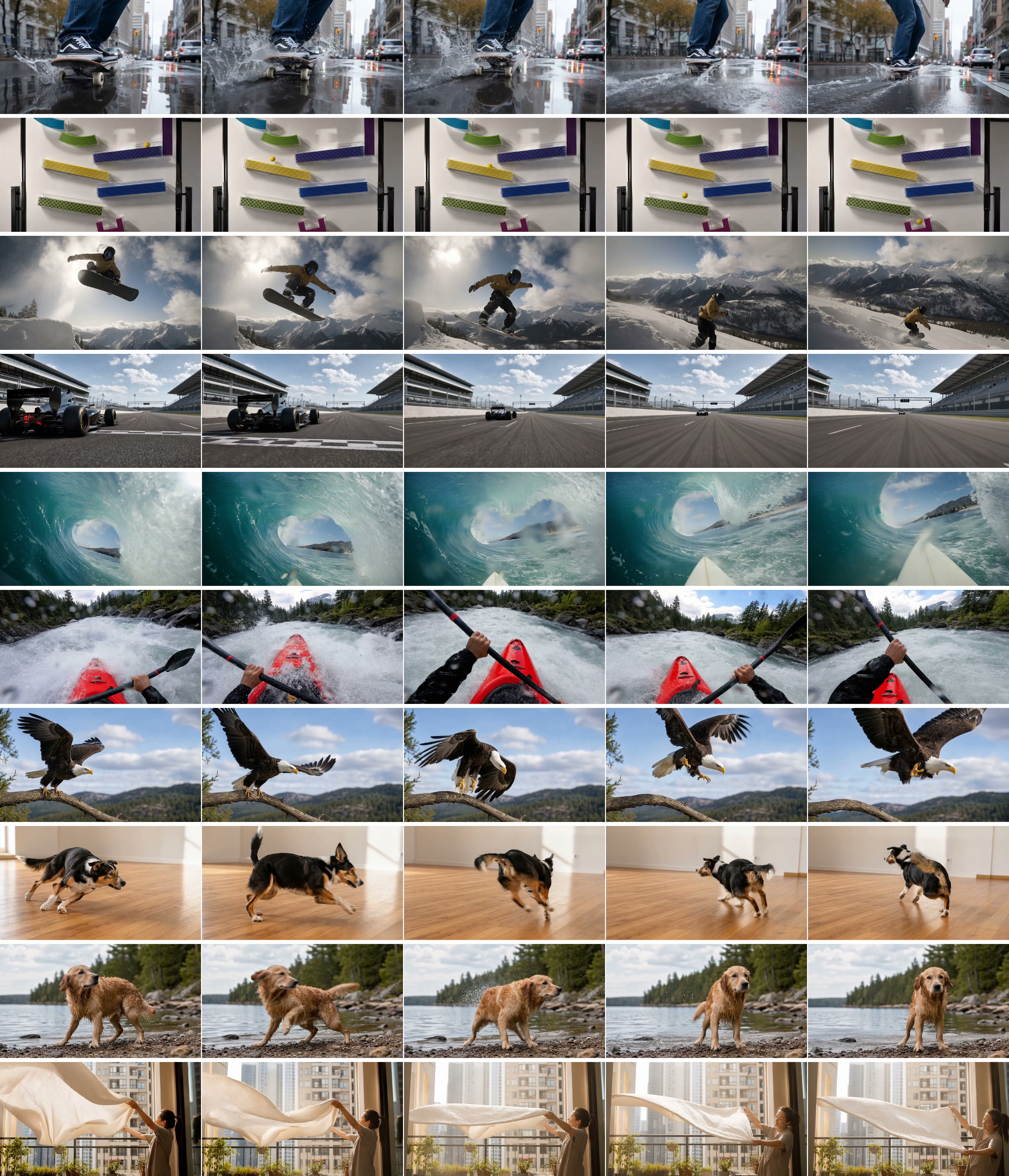}
  \caption{Qualitative results of \method on text-and-image-to-video generation.}
\end{figure}
\clearpage
\begin{figure}[p]
  \centering
  \includegraphics[width=\linewidth]{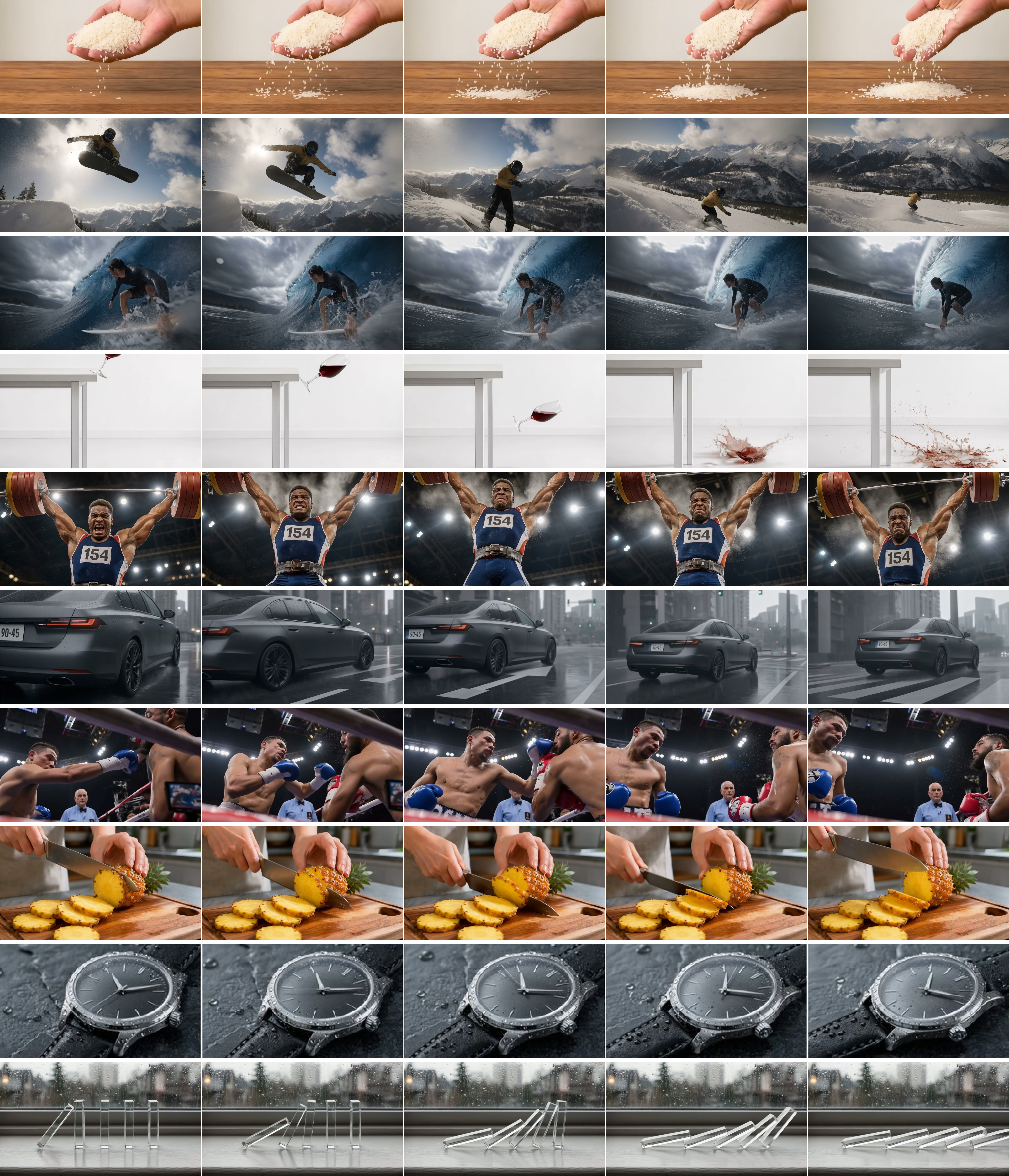}
  \caption{Qualitative results of \method on text-and-image-to-video generation.}
\end{figure}
\clearpage
\begin{figure}[p]
  \centering
  \includegraphics[width=\linewidth]{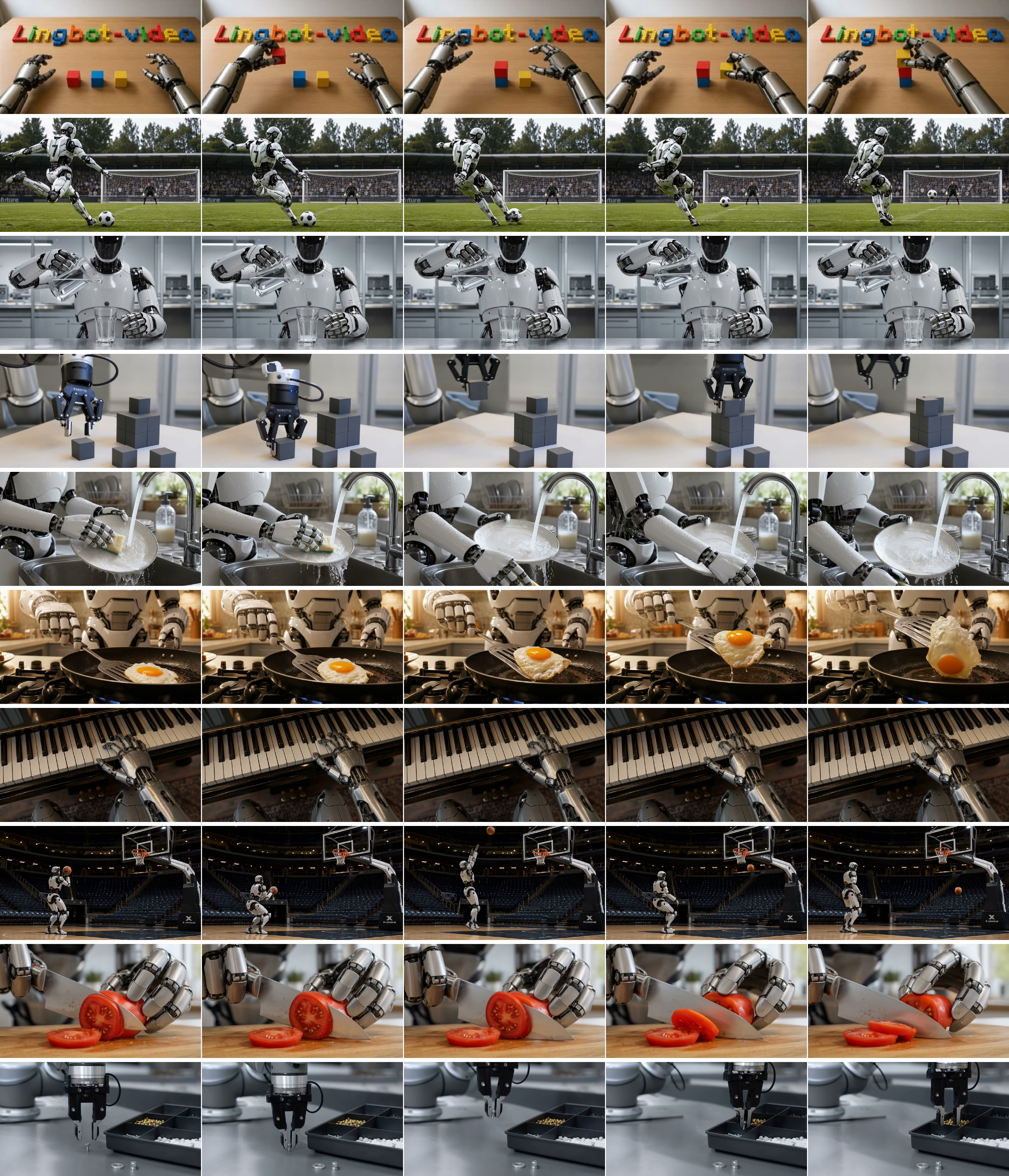}
  \caption{Qualitative results of \method on text-and-image-to-video generation.}
\end{figure}
\clearpage
\begin{figure}[p]
  \centering
  \includegraphics[width=\linewidth]{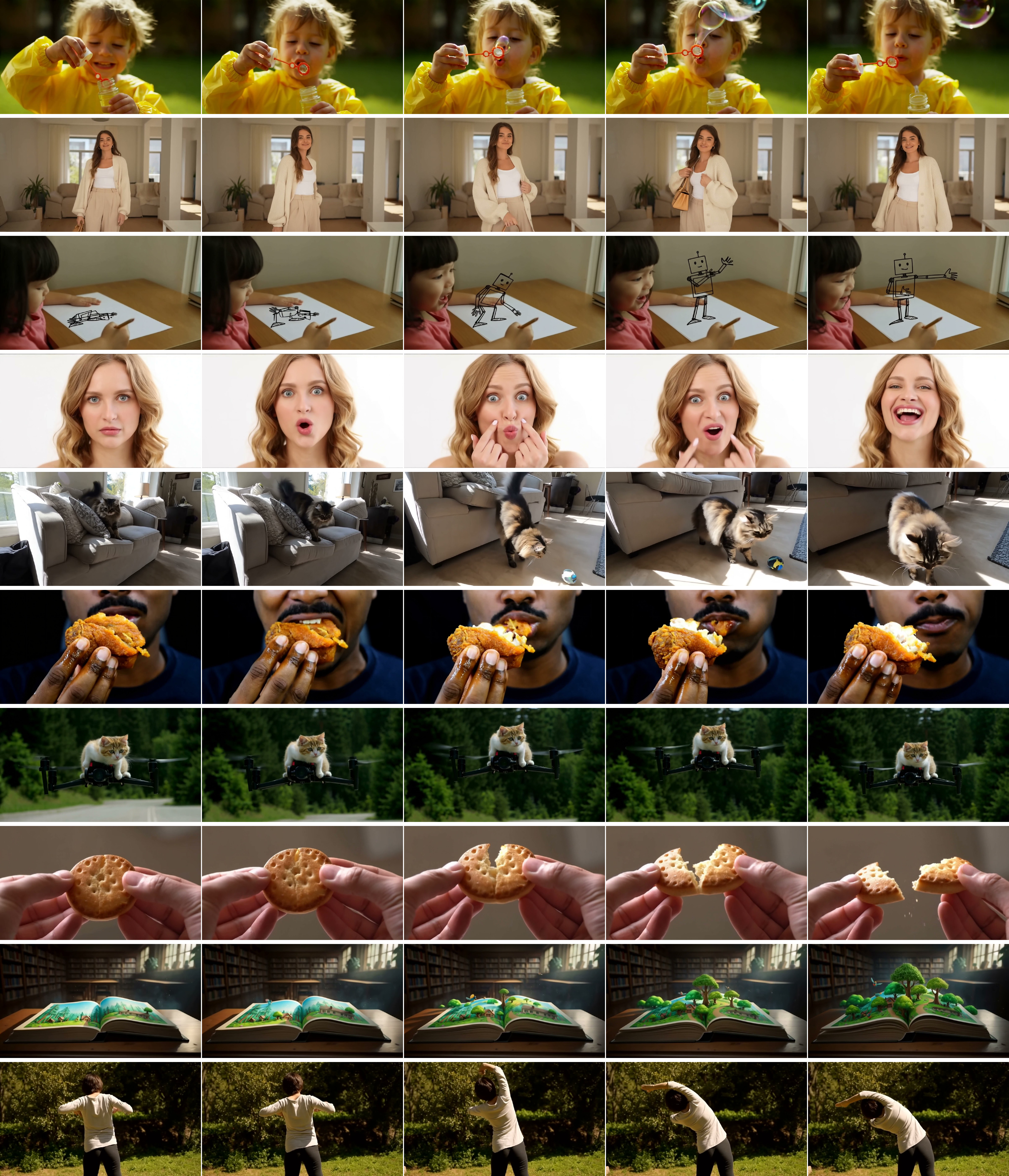}
  \caption{Qualitative results of \method on text-to-video generation. Each row shows five keyframes uniformly sampled from one generated video; time flows from left to right.}
  \label{fig:cases_t2v}
\end{figure}
\clearpage

% !TEX root = ../main.tex

% !TEX root = ../main.tex
\section*{Conclusion and Discussion}
In this work, we present \method, a pioneering MoE-based video foundation model tailored specifically for embodied intelligence, successfully bridging the gap between digital creativity and physical actuation. 
By scaling up the MoE-based Diffusion Transformer from scratch, we achieve an optimal balance between modeling capacity and inference efficiency, paving the way for the development of the next-generation robot brain.
We aim for it to serve as an embodied video simulator that plays critical roles for the robotics community:
\begin{itemize}
    \item \textbf{Data Engine:} Synthesizes high-fidelity, low-cost training data at scale to mitigate data scarcity in robotics.
    \item \textbf{Policy Evaluator:} Serves as a visual simulator to evaluate robot policies safety-critically without real-world risks.
    \item \textbf{Action Planner:} Predicts ``what happens next'' to assist the robot's real-time decision-making and planning.
\end{itemize}
By open-sourcing \method, we hope to inspire and collaborate with the community to collectively push the boundaries of embodied physical engines and next-generation robot brains.
% !TEX root = ../main.tex
\section*{Contributors}

\noindent\textbf{Pre-training:} Shuailei Ma, Jingjing Wang, Kecheng Zheng, Yinghao Xu

\noindent\textbf{Data Infra:} Xinyang Wang, Jingjing Wang, Jiaqi Liao, Shuailei Ma, Yuqi Gan, Weisen Wang, Wei Wu, Jiahao Shao, Hao Ouyang, Qiuyu Wang, Yipengjing Sun, Liangxiao Hu

\noindent\textbf{Post-training:} Jiaqi Liao, Chaoran Feng, Zijing Hu, Zichen Xi, Yanhong Zeng, Qin Zhao, Zifan Shi, Shangzhan Zhang, Nan Xue

\noindent\textbf{Refiner:} Chong Bao

\noindent\textbf{Serving Infra:} Shuailei Ma

\noindent\textbf{Evaluation:} Jingjing Wang*, Zijing Hu*, Chaoran Feng, Lunke Pan

\noindent\textbf{Project Sponsor:} Xing Zhu, Yujun Shen

\noindent\textbf{Project Lead:} Ka Leong Cheng

\noindent $^*$Equal Contribution

% !TEX root = ../main.tex
\section*{Acknowledgments}
We thank 
Qingyan Bai,
Jingye Chen,
Jingyue Chen,
Ka Yu Cheng,
Xiaoyue Duan,
Xiaoqian Ma,
Yihao Meng,
Fan Fan,
Biao Gong,
Bo Jiang,
Yangyan Li,
Yixuan Li,
Zichen Liu,
Fan Lu,
Yichong Lu,
Fangqing Teng,
Hanlin Wang,
Jiahao Wang,
Junke Wang,
Ruonan Wang,
Wenfei Xie,
Jingmei Zhao
Shuai Zhou,
Jiapeng Zhu,
Jiayi Zhu
(\textbf{\textit{listed alphabetically by last name}}) for their valuable discussions and assistance.

% !TEX root = ../main.tex
{
\small
\bibliographystyle{plain}
\IfFileExists{./ref.bib}{
    \bibliography{ref}
}{
    \bibliography{./ref}
}
}

% !TEX root = ../main.tex
\clearpage
\appendix
\section{Structured Caption Details}
\label{app:caption}

\subsection{Example Structured Captions}
\label{app:caption:examples}

We show one representative caption per data category, reproduced verbatim
from the training data.

\begin{captionexample}{Example structured caption for image data.}
{
  "comprehensive_description": "This image captures the opulent interior of a grand palace hall, likely the Grand Kremlin Palace. The room is characterized by a lavish Baroque or Neoclassical style, featuring a high vaulted ceiling adorned with intricate gold patterns and a central medallion. The walls are covered in deep red fabric with gold embroidery, punctuated by tall, fluted white columns with ornate gold capitals. At the far end of the hall, a raised dais holds a large, framed portrait of a historical figure and a single red upholstered throne. The floor is a masterpiece of polished wood parquet, featuring complex geometric and circular inlaid patterns in various shades of brown. A massive, multi-tiered crystal chandelier hangs from the upper left, casting a warm, golden glow throughout the space. The overall atmosphere is one of immense wealth, power, and historical significance, with a symmetrical composition that emphasizes the room's scale and architectural detail.",
  "camera_info": {
    "color": "Warm",
    "frame_size": "Extreme Wide",
    "shot_type_angle": "Low angle",
    "lens_size": "Wide",
    "composition": "Balanced",
    "lighting": "Hard light",
    "lighting_type": "Artificial light"
  },
  "world_knowledge": [],
  "prominent_elements": [
    {
      "name": "vaulted ceiling",
      "description": "A high, arched ceiling featuring a complex pattern of gold medallions and intricate carvings.",
      "location": "spanning the top half of the frame",
      "relative_size": "dominant",
      "shape_and_color": "arched; gold and cream",
      "texture": "carved and metallic",
      "appearance_details": "Features a central large medallion and radiating patterns of smaller circular motifs.",
      "relationship": "Forms the upper boundary of the entire scene.",
      "orientation": "arched",
      "pose": "",
      "expression": "",
      "clothing": "",
      "gender": "",
      "skin_tone_and_texture": ""
    },
    {
      "name": "red wall panels",
      "description": "Large sections of wall covered in deep red fabric with gold embroidery.",
      "location": "spanning the middle and background walls",
      "relative_size": "dominant",
      "shape_and_color": "rectangular; deep red and gold",
      "texture": "fabric and metallic embroidery",
      "appearance_details": "The fabric is decorated with repeating gold floral and heraldic patterns.",
      "relationship": "Provides the primary background for the columns and the throne area.",
      "orientation": "vertical",
      "pose": "",
      "expression": "",
      "clothing": "",
      "gender": "",
      "skin_tone_and_texture": "",
      "is_cluster": true,
      "number_of_objects": "several"
    },
    {
      "name": "fluted columns",
      "description": "Tall, white columns with vertical grooves and ornate gold capitals.",
      "location": "distributed along the walls and at the far end",
      "relative_size": "large",
      "shape_and_color": "cylindrical; white and gold",
      "texture": "smooth and metallic",
      "appearance_details": "The capitals are highly decorative with gold leaf and classical motifs.",
      "relationship": "They support the upper architectural elements and frame the throne area.",
      "orientation": "upright",
      "pose": "",
      "expression": "",
      "clothing": "",
      "gender": "",
      "skin_tone_and_texture": "",
      "is_cluster": true,
      "number_of_objects": "many"
    },
    {
      "name": "parquet floor",
      "description": "A highly polished wooden floor with intricate geometric and circular inlaid patterns.",
      "location": "spanning the bottom half of the frame",
      "relative_size": "dominant",
      "shape_and_color": "flat; various shades of brown and tan",
      "texture": "smooth and glossy",
      "appearance_details": "Features complex interlocking geometric shapes and circular medallions.",
      "relationship": "Reflects the light from the chandelier and the colors of the walls.",
      "orientation": "horizontal",
      "pose": "",
      "expression": "",
      "clothing": "",
      "gender": "",
      "skin_tone_and_texture": ""
    },
    {
      "name": "large portrait",
      "description": "A framed painting depicting a historical figure, likely a monarch.",
      "location": "center background, above the throne",
      "relative_size": "medium",
      "shape_and_color": "rectangular; dark tones with gold frame",
      "texture": "matte",
      "appearance_details": "The figure is dressed in elaborate historical attire and is seated.",
      "relationship": "Positioned as the focal point of the room's back wall.",
      "orientation": "upright",
      "pose": "",
      "expression": "",
      "clothing": "",
      "gender": "",
      "skin_tone_and_texture": ""
    },
    {
      "name": "ornate throne",
      "description": "A single, high-backed chair upholstered in red fabric with gold trim.",
      "location": "center background, on a raised platform",
      "relative_size": "small",
      "shape_and_color": "rectangular; red and gold",
      "texture": "fabric and metallic",
      "appearance_details": "Features a high back and gold-colored armrests and legs.",
      "relationship": "Sits on a small red-carpeted dais in front of the portrait.",
      "orientation": "upright",
      "pose": "",
      "expression": "",
      "clothing": "",
      "gender": "",
      "skin_tone_and_texture": ""
    },
    {
      "name": "crystal chandelier",
      "description": "A large, multi-tiered chandelier with numerous light sources.",
      "location": "top-left corner",
      "relative_size": "medium",
      "shape_and_color": "complex; gold and clear",
      "texture": "metallic and glass",
      "appearance_details": "Features multiple arms and hanging crystal elements.",
      "relationship": "Hangs from the ceiling and provides the main light source for the left side of the room.",
      "orientation": "hanging",
      "pose": "",
      "expression": "",
      "clothing": "",
      "gender": "",
      "skin_tone_and_texture": ""
    }
  ]
}
\end{captionexample}

\begin{captionexample}{Example structured caption for video data.}
{
  "comprehensive_description": {
    "scene_content_description": "The video captures an outdoor cooking scene where several meat rolls are being prepared on a large, black, circular metal griddle. The griddle is positioned over an open fire, with visible flames and glowing embers at the bottom. The meat rolls, made of thin slices of red meat stuffed with green herbs and orange carrot pieces, are held together by wooden toothpicks. Throughout the video, wisps of white smoke rise from the hot surface. A person's hand periodically enters the frame from the top right to place additional meat rolls onto the griddle. The background is softly blurred, showing hints of greenery and a wooden structure, suggesting a garden or backyard setting. The lighting is bright and natural, creating a warm and rustic atmosphere.",
    "camera_movement_description": "The camera is stationary throughout the video, maintaining a steady, eye-level close-up shot of the griddle and the cooking process."
  },
  "camera_info": {
    "color": "Saturated",
    "frame_size": "Wide",
    "shot_type_angle": "High angle",
    "lens_size": "Medium",
    "composition": "Center",
    "lighting": "Hard light",
    "lighting_type": "Daylight"
  },
  "world_knowledge": [],
  "prominent_elements": [
    {
      "name": "meat rolls",
      "description": "Several cylindrical rolls made of thin red meat slices, stuffed with green herbs and orange carrot pieces, secured with wooden toothpicks.",
      "actions": [
        {
          "timestamp": "[0.0s - 7.1s]",
          "action": "The rolls are placed onto the griddle and sizzle as they cook, with smoke rising from them."
        }
      ],
      "location": "Center of the frame on the griddle",
      "relative_size": "large",
      "shape_and_color": "Cylindrical shapes with red, green, and orange colors",
      "texture": "Fleshy and moist",
      "appearance_details": "Visible toothpicks and stuffing ingredients like parsley and carrots.",
      "relationship": "Placed on the black griddle to be cooked.",
      "orientation": "Horizontal",
      "pose": "",
      "expression": "",
      "clothing": "",
      "gender": "",
      "skin_tone_and_texture": "",
      "is_cluster": true,
      "number_of_objects": "several"
    },
    {
      "name": "black griddle",
      "description": "A large, circular, flat metal cooking surface with a slightly textured, matte black finish.",
      "actions": [
        {
          "timestamp": "[0.0s - 7.1s]",
          "action": ""
        }
      ],
      "location": "Occupies most of the lower and middle frame",
      "relative_size": "dominant",
      "shape_and_color": "Circular and black",
      "texture": "Matte and slightly rough",
      "appearance_details": "Shows signs of heat and oil from cooking.",
      "relationship": "Serves as the cooking surface for the meat rolls.",
      "orientation": "Horizontal",
      "pose": "",
      "expression": "",
      "clothing": "",
      "gender": "",
      "skin_tone_and_texture": ""
    },
    {
      "name": "person's hand",
      "description": "A human hand that appears periodically to place meat rolls on the griddle.",
      "actions": [
        {
          "timestamp": "[0.0s - 0.67s]",
          "action": "Enters from the top right and places a roll."
        },
        {
          "timestamp": "[2.67s - 3.67s]",
          "action": "Enters from the top right and places a roll."
        },
        {
          "timestamp": "[5.33s - 6.0s]",
          "action": "Enters from the top right and places a roll."
        }
      ],
      "location": "Top right corner of the frame",
      "relative_size": "medium",
      "shape_and_color": "Flesh-toned hand",
      "texture": "Smooth skin",
      "appearance_details": "Only the hand and part of the forearm are visible.",
      "relationship": "Interacts with the meat rolls and the griddle.",
      "orientation": "Reaching downward",
      "pose": "Reaching and grasping",
      "expression": "",
      "clothing": "",
      "gender": "male",
      "skin_tone_and_texture": "Light skin tone"
    },
    {
      "name": "smoke",
      "description": "Wisps of white and grey smoke rising from the hot griddle.",
      "actions": [
        {
          "timestamp": "[0.0s - 7.1s]",
          "action": "Continuously rises and drifts toward the left side of the frame."
        }
      ],
      "location": "Upper left and center of the frame",
      "relative_size": "medium",
      "shape_and_color": "Amorphous and white/grey",
      "texture": "Wispy and translucent",
      "appearance_details": "Thin, rising plumes.",
      "relationship": "Produced by the heat of the griddle and the cooking meat.",
      "orientation": "Rising upward",
      "pose": "",
      "expression": "",
      "clothing": "",
      "gender": "",
      "skin_tone_and_texture": ""
    }
  ]
}
\end{captionexample}

\begin{captionexample}{Example structured caption for VLA data.}
{
  "comprehensive_description": {
    "scene_content_description": "The video presents a first-person perspective of a robotic workspace, likely a simulated grocery store or automated sorting station. In the foreground, a metal wire shopping cart with red handles is positioned, containing a clear plastic bag with red printed text. Behind the cart is a wooden display shelf divided into several compartments, neatly stocked with a variety of colorful artificial produce. The items include yellow bell peppers, purple eggplants, red pumpkins, white mushrooms, brown potatoes, green cucumbers, yellow corn, red tomatoes, and red chili peppers. Two robotic arms are visible in the frame. The left robotic arm, featuring a grey body and a black two-finger gripper, remains completely stationary in the upper left corner throughout the video. The right robotic arm, which has a white body and a black two-finger gripper, is the active subject. It begins by holding a green cucumber, moves downwards to position the cucumber over the plastic bag in the shopping cart, and then opens its gripper to release the cucumber into the bag. After dropping the item, the right robotic arm retracts upwards and slightly to the right, returning to a higher position above the display shelf. The lighting is bright and even, highlighting the vibrant colors of the artificial vegetables.",
    "camera_movement_description": "The camera remains completely stationary throughout the entire video, maintaining a fixed, slightly high-angle perspective looking down at the shopping cart and the display shelf."
  },
  "camera_info": {
    "color": "Saturated",
    "frame_size": "Wide",
    "shot_type_angle": "High angle",
    "lens_size": "Ultra Wide / Fisheye",
    "composition": "Balanced",
    "lighting": "Hard light",
    "lighting_type": "Artificial light"
  },
  "world_knowledge": [],
  "prominent_elements": [
    {
      "name": "right robotic arm",
      "description": "A robotic arm with a white cylindrical body and a black, two-finger gripper mechanism.",
      "actions": [
        {
          "timestamp": "[0.0s - 1.5s]",
          "action": "Moves downwards while grasping a green cucumber."
        },
        {
          "timestamp": "[1.5s - 5.5s]",
          "action": "Opens its gripper to release the cucumber."
        },
        {
          "timestamp": "[5.5s - 7.2s]",
          "action": "Moves upwards and to the right."
        }
      ],
      "location": "Originates in the upper right, moves to the lower center, and returns to the upper right.",
      "relative_size": "large",
      "shape_and_color": "Cylindrical and angular, white and black.",
      "texture": "Smooth, metallic and plastic.",
      "appearance_details": "Features joints, cables, and a distinct two-finger gripper.",
      "relationship": "Interacts with the green cucumber and moves above the shopping cart.",
      "orientation": "Tilted downwards initially, then retracts upwards.",
      "pose": "",
      "expression": "",
      "clothing": "",
      "is_cluster": false,
      "number_of_objects": ""
    },
    {
      "name": "left robotic arm",
      "description": "A robotic arm with a grey rectangular body and a black, two-finger gripper mechanism.",
      "actions": [
        {
          "timestamp": "[0.0s - 7.2s]",
          "action": "Remains stationary."
        }
      ],
      "location": "Upper left corner of the frame.",
      "relative_size": "large",
      "shape_and_color": "Angular, grey and black.",
      "texture": "Smooth, metallic and plastic.",
      "appearance_details": "Features a distinct two-finger gripper and visible joints.",
      "relationship": "Positioned above the display shelf, inactive.",
      "orientation": "Tilted downwards.",
      "pose": "",
      "expression": "",
      "clothing": "",
      "is_cluster": false,
      "number_of_objects": ""
    },
    {
      "name": "green cucumber",
      "description": "A long, green, artificial vegetable.",
      "actions": [
        {
          "timestamp": "[0.0s - 1.5s]",
          "action": "Grasped by the right robotic arm and moves downwards."
        },
        {
          "timestamp": "[1.5s - 5.5s]",
          "action": "Falls downwards into the plastic bag."
        },
        {
          "timestamp": "[5.5s - 7.2s]",
          "action": "Rests inside the plastic bag."
        }
      ],
      "location": "Starts in the right robotic arm's gripper, ends up in the plastic bag in the lower center.",
      "relative_size": "small",
      "shape_and_color": "Elongated, green.",
      "texture": "Smooth.",
      "appearance_details": "Looks like a standard cucumber.",
      "relationship": "Held by the right robotic arm, then dropped into the plastic bag.",
      "orientation": "Tilted downwards while held, horizontal when resting in the bag.",
      "pose": "",
      "expression": "",
      "clothing": "",
      "is_cluster": false,
      "number_of_objects": ""
    },
    {
      "name": "shopping cart",
      "description": "A metal wire shopping cart with red handles.",
      "actions": [
        {
          "timestamp": "[0.0s - 7.2s]",
          "action": "Remains stationary."
        }
      ],
      "location": "Lower half of the frame.",
      "relative_size": "dominant",
      "shape_and_color": "Rectangular wireframe, silver and red.",
      "texture": "Metallic.",
      "appearance_details": "Contains a clear plastic bag with red text.",
      "relationship": "Serves as the receptacle for the dropped cucumber.",
      "orientation": "Horizontal.",
      "pose": "",
      "expression": "",
      "clothing": "",
      "is_cluster": false,
      "number_of_objects": ""
    },
    {
      "name": "display shelf",
      "description": "A wooden shelf divided into compartments, holding various artificial vegetables.",
      "actions": [
        {
          "timestamp": "[0.0s - 7.2s]",
          "action": "Remains stationary."
        }
      ],
      "location": "Upper half of the frame, behind the shopping cart.",
      "relative_size": "dominant",
      "shape_and_color": "Rectangular, light brown wood with colorful items.",
      "texture": "Wood grain.",
      "appearance_details": "Contains neatly arranged yellow peppers, purple eggplants, red pumpkins, mushrooms, potatoes, corn, tomatoes, and chili peppers.",
      "relationship": "Provides the background context and source of the items.",
      "orientation": "Horizontal.",
      "pose": "",
      "expression": "",
      "clothing": "",
      "is_cluster": false,
      "number_of_objects": ""
    }
  ]
}
\end{captionexample}

\begin{captionexample}{Example structured caption for egocentric data.}
{
  "comprehensive_description": {
    "scene_content_description": "The video presents a first-person perspective of a person working on a piece of machinery, likely in a garage or workshop setting. The environment features a concrete floor scattered with small debris. The primary subject is a large, bright orange lawnmower positioned centrally in the frame. The lawnmower has a prominent black rear tire with deep treads on the left side, a black engine cover on the right, and a blue 'Kohler Professional' sticker on its side. The person's left arm, wearing a black short-sleeved shirt, is visible resting flat on the concrete floor on the left side of the frame. The person's legs, clad in blue jeans, are visible at the bottom. The action begins with the person's right arm, also wearing a black short-sleeved shirt, entering the frame from the right side. The right hand reaches towards the center of the lawnmower, specifically targeting a black cable or wire near the engine area. The hand grasps the cable and pulls it upwards and towards the right, manipulating the component. The left hand remains stationary on the floor throughout the sequence, providing stability.",
    "camera_movement_description": "The camera is body-mounted, providing a first-person point of view. It exhibits slight, continuous panning and tilting movements that correspond to the natural head and body motions of the person working. The shot size remains a close-up on the machinery and the person's hands."
  },
  "camera_info": {
    "color": "Saturated",
    "frame_size": "Medium",
    "shot_type_angle": "High angle",
    "lens_size": "Ultra Wide / Fisheye",
    "composition": "Center",
    "lighting": "Hard light",
    "lighting_type": "Daylight"
  },
  "world_knowledge": [],
  "prominent_elements": [
    {
      "name": "right hand",
      "description": "A human right hand and forearm, wearing a black short-sleeved shirt.",
      "actions": [
        {
          "timestamp": "[0.0s - 2.0s]",
          "action": ""
        },
        {
          "timestamp": "[2.0s - 5.0s]",
          "action": "Enters from the right, reaches towards the center, grasps a black cable, and pulls it upwards and to the right."
        }
      ],
      "location": "Moves from the right edge towards the center-right of the frame.",
      "relative_size": "medium",
      "shape_and_color": "Arm shape, skin tone, black sleeve.",
      "texture": "Smooth skin, fabric texture.",
      "appearance_details": "Wearing a black short-sleeved shirt.",
      "relationship": "Interacts with the black cable on the lawnmower.",
      "orientation": "Extended forward and slightly downwards.",
      "pose": "",
      "expression": "",
      "clothing": "Black short-sleeved shirt.",
      "is_cluster": false,
      "number_of_objects": ""
    },
    {
      "name": "left hand",
      "description": "A human left hand and forearm, wearing a black short-sleeved shirt, resting flat on the ground.",
      "actions": [
        {
          "timestamp": "[0.0s - 5.0s]",
          "action": "Remains stationary on the floor."
        }
      ],
      "location": "Bottom left of the frame.",
      "relative_size": "medium",
      "shape_and_color": "Hand shape, skin tone, black sleeve.",
      "texture": "Smooth skin, fabric texture.",
      "appearance_details": "Wearing a black short-sleeved shirt, a ring is visible on one finger.",
      "relationship": "Resting on the concrete floor, providing stability for the person.",
      "orientation": "Flat, horizontal.",
      "pose": "",
      "expression": "",
      "clothing": "Black short-sleeved shirt.",
      "is_cluster": false,
      "number_of_objects": ""
    },
    {
      "name": "lawnmower",
      "description": "A large piece of outdoor machinery, primarily orange with black components.",
      "actions": [
        {
          "timestamp": "[0.0s - 5.0s]",
          "action": "Remains stationary."
        }
      ],
      "location": "Occupies the center and right portions of the frame.",
      "relative_size": "dominant",
      "shape_and_color": "Complex mechanical shape, predominantly bright orange and black.",
      "texture": "Smooth painted metal, rubber tire.",
      "appearance_details": "Features a large black rear tire with deep treads, a black engine cover, and a blue 'Kohler Professional' sticker.",
      "relationship": "The object being worked on by the right hand.",
      "orientation": "Upright.",
      "pose": "",
      "expression": "",
      "clothing": "",
      "is_cluster": false,
      "number_of_objects": ""
    },
    {
      "name": "black cable",
      "description": "A thin, flexible black wire or cable attached to the lawnmower.",
      "actions": [
        {
          "timestamp": "[0.0s - 2.0s]",
          "action": "Remains stationary."
        },
        {
          "timestamp": "[2.0s - 5.0s]",
          "action": "Grasped by the right hand and pulled upwards and to the right."
        }
      ],
      "location": "Center of the frame, near the engine area of the lawnmower.",
      "relative_size": "small",
      "shape_and_color": "Thin, linear, black.",
      "texture": "Smooth, flexible.",
      "appearance_details": "Appears to be a control cable or wire.",
      "relationship": "Attached to the lawnmower, manipulated by the right hand.",
      "orientation": "Curved, extending from the engine area.",
      "pose": "",
      "expression": "",
      "clothing": "",
      "is_cluster": false,
      "number_of_objects": ""
    }
  ]
}
\end{captionexample}

\end{document}